\documentclass{article}

\usepackage{arxiv}

\usepackage[utf8]{inputenc} 
\usepackage[T1]{fontenc}    
\usepackage{hyperref}       
\usepackage{url}            
\usepackage{booktabs}       
\usepackage{amsfonts}       
\usepackage{nicefrac}       
\usepackage{microtype}      
\usepackage{lipsum}		
\usepackage{graphicx}
\usepackage{natbib}
\usepackage{doi}
\usepackage{amsmath}
\usepackage{multirow}
\usepackage{tabularx}
\usepackage{array}
\usepackage{bm}
\usepackage{enumitem}
\usepackage{appendix}

\newcolumntype{L}[1]{>{\raggedright\arraybackslash}p{#1}}
\newcolumntype{Y}{>{\raggedright\arraybackslash}X}

\newcommand{\code}[1]{\texttt{#1}}

\newcommand{\labellist}[1]{%
  \begin{tabular}[t]{@{}l@{}}
  #1
  \end{tabular}%
}

\newcommand{\celltop}[1]{%
  \vtop{\hsize=\linewidth
  \strut #1\strut}%
}

\newcommand{\annlabels}[1]{%
  \renewcommand{\arraystretch}{1.08}%
  \begin{tabular}[t]{@{}l@{}}
  #1
  \end{tabular}%
}

\title{Data and Evaluation Closed-Loop for Model Capability Enhancement}

\author{
Zhixuan Li\thanks{Co-first authors with equal contributions.} \\
Baidu Inc. \\
\texttt{lizhixuan.2017@tsinghua.org.cn}
\And
Jiangan Yuan\makeatletter\footnotemark[1]\makeatother \\
Baidu Inc. \\
\texttt{jianganyuan@link.cuhk.edu.cn}
\And
Han Xu \\
Baidu Inc. \\
\texttt{xhbj66@gmail.com}
}

\renewcommand{\shorttitle}{Data and Evaluation Closed-Loop}

\hypersetup{
pdftitle={A template for the arxiv style},
pdfsubject={q-bio.NC, q-bio.QM},
pdfauthor={David S.~Hippocampus, Elias D.~Striatum},
pdfkeywords={First keyword, Second keyword, More},
}

\begin{document}
\maketitle

\begin{abstract}
Model capability is the central variable in LLM pre-training, yet is never observed directly: data shapes it prospectively, while evaluation reveals it only retrospectively, compressing samples, prompts, decoding, and scoring rules into one noisy score. Practical optimization runs this backward: a failure is observed first, and the engineer must infer the corpus fix. The two sides speak incompatible vocabularies---benchmark names and per-sample correctness versus data sources, domains, and quality labels---so this inference is usually intuition, not method. We close this gap with the \emph{capability slice}: a group of evaluation samples sharing background condition, task type, solving operation, and output constraint---precise enough to localize a single weakness yet stable enough to survive aggregation, unlike a benchmark name, too coarse, or a single sample, too noisy. Built around this unit, an evaluation taxonomy, a non-instruction data taxonomy, and mapping rules form a closed loop turning a benchmark-level failure into a targeted, testable data intervention. We test this loop on two case studies pulling in opposite directions. First, the loop rules the data out: continued pre-training drives BBH down by $-46.82\%$, but diagnosis traces this to a single masked \texttt{\textless EOS\textgreater} loss rather than weakened reasoning; restoring it recovers BBH to $66.44$, above the original checkpoint, without changing the data. Second, the loop rules the data in: a persistent math-reasoning weakness is decomposed by solving operation into specific failing combinations, and a weakness-targeted sampling procedure built from it lifts AIME2025/AIME2026 Pass@128 from $6.67$/$0.00$ to $26.67$ each. The same unmodified loop reaches opposite, correct verdicts in both cases, showing the evaluation-to-data inference can be routine, auditable, and experimentally validated rather than intuitive.
\end{abstract}

\section{Introduction}
\label{sec:introduction}

In large language model (LLM) pre-training, the quantity that ultimately matters is model capability, but capability itself is never observed directly: it is shaped, prospectively, by the training data the model consumes, and it surfaces only after the fact, retrospectively, as a benchmark score. That score is itself an aggregate, compressing heterogeneous evaluation samples, prompt templates, decoding configurations, answer-extraction rules, and scoring protocols into a single number attached to a benchmark name. State-of-the-art systems are now evaluated on suites that aggregate dozens of such benchmarks, spanning language understanding, factual knowledge, mathematical reasoning, and code generation~\citep{hendrycks2021mmlu,zellers2019hellaswag,bisk2020piqa,suzgun2023bbh,cobbe2021gsm8k,hendrycks2021math,chen2021humaneval,austin2021mbpp,joshi2017triviaqa,lai2017race,dua2019drop,wang2024mmlupro,huang2023ceval,li2024cmmlu,zhong2024agieval}, with headline performance reported as a single score per benchmark name. This compression obscures more than it reveals: two models with the same MMLU score may fail on disjoint kinds of problems; a gain on a broad benchmark may stem from improved closed-book factual recall rather than from stronger reasoning; a regression on BBH may originate from output-format violations rather than from any change in task-solving ability. As long as a benchmark name is treated as a direct readout of model capability, the question that matters for the next training run---what data signal, if added, reweighted, or filtered, would address the failure---remains under-determined.

Practical optimization, however, must traverse this relationship in reverse: the failure shows up first, in an evaluation result, and only afterward can an engineer work backward to the change in the training corpus that might address it. This is consistent with a simple operational claim that motivates the rest of the paper: under a fixed model architecture, training objective, and optimization recipe, training data is the principal controllable factor shaping model capability, while evaluation supplies only a noisy observation of that capability. The data-side problem has accordingly received considerable attention, with corpus-composition methods that search for or optimize over data mixtures~\citep{xie2023doremi,liu2024regmix,ye2024datamixinglaws}, large-scale quality filtering~\citep{penedo2024fineweb,li2024datacomplm}, and synthetic instruction generation~\citep{wang2023selfinstruct,xu2024wizardlm,taori2023alpaca,wang2023tulu}. On the evaluation side, an equally large body of work has refined what is measured, through finer-grained benchmarks~\citep{rein2024gpqa,glazer2024frontiermath} and sample-level analyses of failure modes. Both literatures are mature, and we draw on both; what remains comparatively under-developed is the \emph{interface} between them. The obstacle is that evaluation and training data are described in different languages: evaluation results arrive as benchmark scores or per-sample correctness, whereas training data is catalogued by source, domain, and quality label, and bridging the two has so far been a matter of intuition more than of method. Given an evaluation failure, what kind of training-data signal is the failure actually evidence of, and how should one decide among, for instance, adding more mathematical content, reweighting an existing source, or repairing supervision density?

This paper develops that interface. Mapping between the two sides at the level of benchmark names is too coarse to isolate a single weakness, while mapping at the level of individual samples is too noisy to be trusted; neither granularity supports a reliable inference from failure to fix. We therefore introduce an intermediate representation, the \emph{capability slice}: a group of evaluation samples that share structured conditions along four dimensions---background condition, task type, solving operation, and output constraint. A capability slice is more specific than a benchmark and more stable than a single sample, and it is the unit on which our diagnosis, data-side mapping, and intervention validation all operate.

Around this representation, we build an analysis toolkit with three components (Section~\ref{sec:analysis_toolkit}), instantiating the conceptual framework developed in Section~\ref{sec:conceptual_framework}. First, an \textbf{evaluation sample taxonomy} (Section~\ref{subsec:evaluation_sample_taxonomy}) annotates each sample along the four dimensions above, applied to sixteen benchmarks commonly used during pre-training; this taxonomy supports benchmark-composition analysis, the localization of weak capability slices below the benchmark level, and a structured attribution of post-intervention score changes. Second, a \textbf{non-instruction data taxonomy} (Section~\ref{subsec:non_instruction_data_taxonomy}) characterizes raw corpus segments along four data-side dimensions---content world, discourse structure, operation opportunity, and supervision density---chosen so that, where meaningful, they correspond to the evaluation-side dimensions. Third, a set of \textbf{evaluation-to-data mapping rules} (Section~\ref{subsec:evaluation_to_data_mapping_rules}) connects the two taxonomies: a small set of taxonomy-aligned correspondences covers the dimensions that were jointly designed across the two sides, and a complementary LLM-assisted mechanism handles slices that fall outside them. Together, the three components define a closed loop,
\[
\text{Weak Capability Slice} \rightarrow \text{Data Affordance Profile} \rightarrow \text{Data Action} \rightarrow \text{Experimental Validation} \rightarrow \text{Result Analysis},
\]
that converts a benchmark-level observation into an auditable, iterable optimization procedure. We treat the mapping rules as structured heuristics for generating data-intervention hypotheses rather than as causal claims; any resulting intervention remains subject to downstream experimental validation.

We put this loop under stress with two case studies built on a continued pre-training pipeline, deliberately chosen so that they push the diagnosis in opposite directions. In the first (Section~\ref{subsec:case_study_output_constraint}), the loop is asked to explain a regression, and the right answer turns out to rule the data out entirely: continued pre-training improves nearly every benchmark but produces a large regression on BBH ($-46.82\%$ relative). Reading the regression through the output-constraint dimension shows that the model frequently emits the correct gold answer and then continues generating beyond it, that the resulting filtered responses become abnormally long, and---tracing the failure into the training pipeline itself---that the loss on the \(\texttt{\textless EOS\textgreater}\) token has been masked during continued pre-training. Restoring this single piece of supervision raises BBH from $25.14$ to $66.44$, above the warm-start checkpoint, without degrading, and in several cases modestly improving, the other benchmarks; the training data itself is never touched. The diagnostic loop thereby converts what initially appeared to be a reasoning regression into a localized, fully correctable termination-supervision bug.

The second case study (Section~\ref{subsec:case_study_solving_operation}) addresses a more characteristic failure mode of pre-training, weak mathematical problem solving, and here the loop runs the other way: the diagnosis rules the data in. The warm-start checkpoint attains only $0.05$ at AIME2025 Pass@1 and $0.00$ at AIME2026 Pass@1; on AIME2026 the model fails to surface a correct trajectory even at Pass@128, indicating that test-time scaling alone cannot resolve the gap. Decomposing the evaluation set by solving operation shows that the weakness is not isolated arithmetic but specific operation combinations involving constraint tracking, boundary-case reasoning, symbolic transformation, counting, and comparison; a pairwise- and composition-effect analysis further shows that several of these combinations are non-additive, performing worse than the sum of their pairwise effects would predict. From this diagnosis we derive an importance-sampling procedure (Section~\ref{subsubsec:sampling_from_synthetic_instruction_data}) that ranks operation combinations by a diagnosis-aligned weakness score, matches synthetic instruction examples to the highest-scoring combinations, and draws from the resulting pool using Efraimidis--Spirakis weighted reservoir sampling without replacement~\citep{efraimidis2006weighted} under a bounded duplication factor. Holding the training recipe and instruction-data budget fixed, the resulting checkpoint reaches $26.67$ at AIME2025 Pass@128 and $26.67$ at AIME2026 Pass@128, while leaving general-domain performance essentially unchanged.

We close this introduction with two clarifications about the kind of contribution this paper makes, before turning to the body of the paper. First, the contribution is methodological rather than a claim about any specific model or benchmark: the taxonomies and mapping rules supply a vocabulary and a set of heuristics for converting a benchmark-level observation into a testable data-intervention hypothesis, and every such hypothesis is still required to clear a controlled, single-variable experimental design before being treated as a finding. Second, what unifies the two case studies into a single demonstration, rather than two unrelated debugging anecdotes, is that the same closed loop reaches the correct diagnosis in both directions: in the first, it rules the data out, redirecting what looked like a data or capability problem toward the training objective, where the fix actually belonged; in the second, it rules the data in, localizing a genuine capability gap to specific operation combinations and resolving it through a targeted data intervention. The loop is therefore informative not only when it points toward a data fix, but also when it correctly rules one out. Taken together, the two case studies make the same point from opposite sides: turning an evaluation failure into a data fix is something the same procedure can do reliably, with each inferential step open to inspection and each resulting hypothesis checked experimentally, rather than something that depends on an engineer's intuition. The remainder of this paper follows the structure outlined above: Section~\ref{sec:conceptual_framework} develops the conceptual framework in full, Section~\ref{sec:analysis_toolkit} presents the analysis toolkit and the closed loop it instantiates, Section~\ref{sec:case_study} reports the two case studies, and Section~\ref{sec:conclusion} discusses the scope and limitations of the resulting methodology.

\section{Conceptual Framework}
\label{sec:conceptual_framework}

We organize our framework around three central objects: \emph{evaluation}, \emph{model capability}, and \emph{training data}. Evaluation yields observations of model behavior, but these observations are entangled with implementation choices such as the prompt template, decoding strategy, answer-extraction rule, and scoring protocol. Model capability refers to the model's stable behavioral pattern under structured conditions, rather than to a score attached to a benchmark name. Training data provides the primary capability-forming signal: it exposes the model to knowledge, discourse structures, reasoning opportunities, and supervised input--output patterns from which capabilities emerge and through which they are expressed.

The core premise of this work is that evaluation is a noisy observation of model capability, while model capability is, in turn, shaped by training data under a fixed model architecture, training objective, and optimization recipe. Thus, a benchmark score should not be interpreted as a direct readout of a single model capability; rather, it is an aggregate outcome produced by the interaction between model capability, sample composition, evaluation protocol, and measurement noise. This motivates a more structured form of analysis, in which benchmark results are decomposed into sample-level observation conditions, aggregated into capability-level behavioral patterns, and then mapped back to the training data that may strengthen or weaken these patterns.

This section follows an engineering analysis order:
\[
\text{Evaluation} \rightarrow \text{Model Capability} \rightarrow \text{Training Data}.
\]
We begin with evaluation because practical optimization is typically triggered by an observed failure: a model underperforms on a benchmark, on a subset of samples, or on a specific task family. To make such failures actionable, we first examine what an evaluation sample is actually observing. We then define capability as the model's stable behavioral pattern over groups of structurally similar evaluation samples, which we later formalize as capability slices. Finally, we connect these capability slices to training data, which provides the intervention surface for improving future models.

This analysis order differs from the underlying causal and optimization loop. In the training process, data first shapes model capability, which is then exposed through evaluation:
\[
\text{Training Data} \rightarrow \text{Model Capability} \rightarrow \text{Evaluation}.
\]
Once evaluation reveals a weakness, the diagnosis is mapped back onto a data intervention, closing the engineering loop:
\[
\text{Evaluation Failure} \rightarrow \text{Capability Diagnosis} \rightarrow \text{Data Intervention}.
\]
The purpose of the conceptual framework is therefore not merely to describe evaluation, capability, and training data as separate components, but to define an actionable interface among them: evaluation tells us where the model fails, capability analysis tells us what kind of behavior is unstable or missing, and training-data analysis tells us what signals can be adjusted to remedy the failure.
\subsection{Understanding Evaluation}
\label{subsec:understanding_evaluation}

Evaluation serves as the entry point of the proposed analysis framework, since practical model optimization is typically triggered by an observed failure: a model underperforms on a benchmark, on a subset of samples, or on a specific task family. However, a benchmark score should not be interpreted as a direct readout of a single model capability. Instead, it is a noisy aggregate produced by several interacting factors, including the benchmark's sample distribution, the prompt template, the decoding strategy, the answer-extraction rule, the scoring protocol, sample ambiguity, potential contamination, and the model's underlying capability. We formalize this by writing the observed benchmark score as
\[
s_b = \mathcal{M}\bigl(\mathcal{D}_b, \pi, \delta, \rho, f_\theta\bigr) + \epsilon_b,
\]
where $\mathcal{M}$ denotes the evaluation procedure that aggregates these factors into a score, $\mathcal{D}_b$ denotes the benchmark sample set, $\pi$ denotes the prompting protocol, $\delta$ denotes the decoding strategy, $\rho$ denotes the parsing and scoring rule, $f_\theta$ denotes the evaluated model, and $\epsilon_b$ summarizes measurement noise such as annotation noise, ambiguous samples, unstable generation, and contamination effects. This formulation is not meant to serve as a statistical estimator; rather, it highlights that the observed score conflates model behavior with evaluation-specific implementation choices.

This distinction matters in practice because many widely used benchmarks are internally heterogeneous. A benchmark may contain samples that require factual recall, passage understanding, commonsense plausibility judgment, arithmetic computation, symbolic manipulation, or output-format following. Conversely, samples from different benchmarks may probe similar underlying behaviors. As a result, benchmark-level comparison can obscure the source of improvement or degradation. For example, an average gain on a broad benchmark may stem from improved closed-book factual recall rather than from stronger reasoning, whereas a drop on a math-oriented benchmark may be caused by answer-format errors rather than by failures in symbolic manipulation.

Because benchmark scores are noisy aggregates, benchmark-level analysis alone is insufficient for data-driven optimization. To make evaluation actionable, we move from benchmark-level scores to sample-level observation conditions. The relevant question is therefore not only whether the model answered a sample correctly, but what the sample was designed to observe: what information the model needed to condition on, what task it was asked to perform, what operations were required to transform the input into the answer, and what output form was required by the scorer. We therefore treat each evaluation sample as a structured observation of model behavior rather than as an isolated binary outcome.

We decompose each evaluation sample along four dimensions: \emph{background condition}, \emph{task type}, \emph{solving operation}, and \emph{output constraint}. These dimensions are designed to answer four complementary questions:
\[
\begin{aligned}
\text{background condition} &:\quad \text{What information does the model need to condition on?} \\
\text{task type} &:\quad \text{What task is the model asked to perform?} \\
\text{solving operation} &:\quad \text{What operations are required to transform the input into the answer?} \\
\text{output constraint} &:\quad \text{In what form must the answer be produced and scored?}
\end{aligned}
\]
This decomposition is intentionally independent of benchmark names. Its purpose is to expose the structure of the observation itself, so that heterogeneous samples within the same benchmark can be separated and structurally similar samples across benchmarks can be aggregated.

The first dimension, \emph{background condition}, describes the information environment required to solve the sample. It includes whether the sample is closed-book or context-dependent, whether the relevant evidence is local or distributed across a long passage, and whether the input takes the form of a dialogue, table, code snippet, or expository paragraph. It also includes whether the sample contains ambiguity, noisy formatting, or underspecified references. This dimension is motivated by the observation that many model failures are caused not by the target task alone, but by the information condition under which the task is presented. For instance, factual question answering under a closed-book setting differs from factual question answering where the answer is explicitly stated in a long passage; similarly, arithmetic over a clean word problem differs from arithmetic over a noisy table. By explicitly labeling the background condition, we can separate failures caused by missing knowledge, long-context dependence, discourse complexity, or input noise, etc.

The second dimension, \emph{task type}, identifies the primary task requested by the sample. Examples include factual question answering, passage understanding, logical reasoning, arithmetic word problem solving, code generation, etc. This dimension provides a coarse semantic grouping of evaluation samples, and its main role is to prevent over-reliance on benchmark names: a single benchmark may contain multiple task types, while the same task type may appear across several benchmarks. Task-type labeling therefore supports cross-benchmark aggregation, enabling comparison of model behavior on similar samples across different evaluation suites.

The third dimension, \emph{solving operation}, describes the operations required to derive the answer from the available information. These operations may include fact recall, span extraction, multi-hop composition, arithmetic computation, program synthesis, etc. This dimension is motivated by the observation that task names are often too coarse to explain failures: two samples may both be categorized as mathematical problem solving, yet one may primarily require numerical calculation while another requires symbolic transformation. Similarly, two reading-comprehension samples may differ substantially depending on whether they require local span extraction, cross-paragraph evidence aggregation, or comparison among options. Solving-operation labels thus provide a finer-grained decomposition than task type alone, making it possible to localize a failure to the specific operation the model fails to execute.

The fourth dimension, \emph{output constraint}, captures the required answer form and the scoring sensitivity. This includes whether the output is a multiple-choice option, a short entity, a number or executable code, as well as constraints such as exact-match scoring, numerical tolerance, option-letter formatting, code executability, and instruction-following requirements. This dimension is necessary because a model can fail a sample even when it possesses the relevant knowledge and performs the main reasoning step correctly: in such cases, the failure may instead originate from answer extraction, formatting, verbosity, invalid units, or non-executable code. Separating output constraints from task type and solving operation allows us to distinguish capability failures from expression and scoring failures.

This four-dimensional decomposition provides several practical benefits. First, it reveals heterogeneity inside benchmark-level scores: instead of treating a benchmark as a single capability probe, we can identify which subsets of samples contribute to an observed gain or drop. Second, it enables cross-benchmark aggregation by grouping samples that share similar observation conditions. Third, it helps localize the source of failure, distinguishing whether a model lacks background knowledge, cannot process the required context, fails to perform the necessary operation, or violates the output constraint. Fourth, and most importantly for this work, it creates a structured interface between evaluation and data: once a failure is localized to a particular combination of background condition, task type, solving operation, and output constraint, we can search for or construct training data that supplies the corresponding learning signal.
\subsection{Understanding Model Capability}
\label{subsec:understanding_model_capability}

The sample-level decomposition introduced in Section~\ref{subsec:understanding_evaluation} provides the basis for defining model capability. Since a benchmark score is a noisy aggregate over heterogeneous samples and evaluation-specific implementation choices, we do not treat model capability as a property of a benchmark name. Instead, we operationalize model capability as the model's stable performance pattern across structured evaluation conditions, which we formalize below as capability slices.

Concretely, an evaluation sample defines a single observation point that specifies a particular background condition, task type, solving operation, and output constraint under which the model is observed. A capability slice is a group of evaluation samples that share similar structured conditions along these dimensions, and model capability is defined as the model's consistent behavioral pattern over one or more such slices. Under this view, capability is neither a benchmark label nor the outcome of a single sample; it is a regularity in model behavior that becomes meaningful only after aggregating observations over structurally similar conditions.

Correctly identifying a weak capability slice requires ruling out two sources of false signal. The first is the isolated sample: a single failed item should not be directly interpreted as a capability failure, since a model may fail because of accidental generation instability, ambiguous wording, or an answer-parsing issue, and such a failure is an observation rather than a diagnosis. For example, an isolated error on a formal mathematics problem does not by itself establish a weakness in mathematical reasoning; only repeated failures on samples that require symbolic transformation under strict numerical output constraints indicate an unstable capability slice. A weak capability slice is therefore established by consistency of failure across structurally similar samples, not by any single outcome.

The second source of false signal is the benchmark name. A benchmark-level score should not be directly interpreted as a capability measurement, since a benchmark often mixes samples with different background conditions, task types, solving operations, and output constraints, such that its aggregate score may conceal multiple distinct behavioral patterns. Consequently, capability slices do not align with benchmark boundaries in either direction: samples from different benchmarks may belong to the same capability slice if they require similar conditions and operations, while samples from the same benchmark may belong to different capability slices if they rely on different operations or output constraints. For instance, a broad knowledge benchmark may contain closed-book entity recall, concept discrimination, commonsense plausibility judgment, and multi-step reasoning samples; treating the benchmark as a single capability would collapse these distinctions. Likewise, a model that performs well on local span extraction but fails on cross-paragraph evidence aggregation should not be described simply as weak at reading comprehension—the failure is more specifically associated with long-context evidence integration. A weak capability slice, in other words, is defined by shared structured conditions, not by benchmark identity.

Once a weakness has been correctly localized in this way, the natural next step is data-driven intervention: identifying what kind of training data can supply the missing signal. A weakness in closed-book factual recall points to insufficient or poorly retained knowledge coverage, whereas a weakness in long-context evidence aggregation reflects insufficient exposure to documents with distributed dependencies and recoverable evidence chains. Limited symbolic transformation ability suggests insufficient exposure to mathematical derivation or proof-like supervision, and weak executable code generation indicates insufficient alignment between natural-language specifications, API constraints, and runnable programs. Thus, capability slices serve as the intermediate representation between evaluation failures and training data interventions.
\subsection{Understanding Training Data}
\label{subsec:understanding_training_data}

The preceding subsections describe evaluation as a noisy aggregate of model capability and evaluation-specific implementation choices, and model capability as a stable behavioral pattern over capability slices. This subsection turns to the remaining component of the framework: training data, which we treat as the primary source of intervention. Within a fixed model architecture, training objective, and optimization recipe, training data is the principal controllable factor shaping model capability. Consequently, once an evaluation failure has been localized to a specific capability slice, the natural next step is to identify what kind of training signal may strengthen the corresponding behavior.

The relationship between training data and model capability is indirect, since training data rarely specifies capability labels explicitly. Instead, it exposes the model to knowledge, discourse structures, reasoning opportunities, and supervised input--output patterns, from which capabilities are learned. Some of these signals are implicit: a textbook passage, for example, may contain definitions, derivations, and worked examples without explicitly asking the model to solve a task. Other signals are explicit: an instruction--response pair may directly ask the model to answer a question, generate code, follow a specified format, or produce a reasoning trace. This distinction motivates a first-order separation between instruction data and non-instruction data.

\subsubsection{Instruction vs. Non-instruction Data}
\label{subsubsec:instruction_vs_non_instruction_data}

We define \emph{instruction data} as training examples that pair an explicit, task-oriented input with a target response demonstrating the desired behavior under that input. The input specifies what the model is asked to do, through a natural-language instruction, a question, a tool-use request, a code specification, or another task-oriented prompt, while the target response realizes the corresponding output, such as a short answer, an explanation, a derivation, a formatted result, or a tool-calling trajectory. In this sense, instruction data makes part of the evaluation condition explicit at training time: it typically specifies the task type, the relevant background context, and the required output constraint, echoing the dimensions used to decompose evaluation samples in Section~\ref{subsec:understanding_evaluation}.

We define \emph{non-instruction data} as the complement of instruction data within the training corpus: text that is not primarily organized as paired task input and target response. This category includes naturally occurring or weakly structured text such as web pages, books, encyclopedic articles, academic papers, news articles, forum discussions, code repositories and tables, etc. Such data is typically consumed through the standard language modeling objective, under which the supervision signal is distributed across the sequence rather than concentrated in an explicit answer span.

This distinction matters because instruction and non-instruction data shape model capability through different mechanisms. Non-instruction data primarily supplies the substrate from which capabilities are formed: knowledge coverage, conceptual associations, discourse structure, latent reasoning patterns, and the operational opportunities embedded in raw text. Instruction data, in contrast, more directly shapes how these latent capabilities are activated and expressed under task-like conditions: it aligns inputs with objectives, constrains output formats, demonstrates reasoning or tool-use trajectories, and teaches the model how to respond when an evaluation-like request is presented.

This separation is not intended to establish a strict hierarchy in which one data type is universally more important than the other; rather, it provides an analytical lens for examining each type on its own terms. For non-instruction data, the central question is what learnable signals are latent in the text and how densely they occur. For instruction data, the central question is how raw content, solving operations, and output constraints are converted into explicit, supervised behavior. The following subsections analyze these two data types from their respective perspectives, so that capability weaknesses diagnosed through evaluation can be mapped to concrete training-data interventions, completing the data-intervention surface introduced at the start of this section.

\subsubsection{Understanding Non-instruction Data}
\label{subsubsec:understanding_non_instruction_data}

As noted in Section~\ref{subsubsec:instruction_vs_non_instruction_data}, non-instruction data does not present itself as explicit task supervision: rather than pairing a clearly separated instruction with a target response, it exposes the model to naturally occurring text, code, tables, documents, and discourse patterns. The learning signal is therefore latent, embedded in what the text talks about, how the information is organized, what transformations are implicitly demonstrated, and how recoverable the next-token prediction target is from its surrounding context. To analyze such data in a way that is useful for capability-oriented optimization, we characterize non-instruction data along four complementary perspectives: \emph{content world}, \emph{discourse structure}, \emph{operation opportunity}, and \emph{supervision density}.

The first perspective, \emph{content world}, describes the knowledge, entities, concepts, domains, and symbolic systems covered by the data, i.e., what part of the world the data exposes the model to. For example, encyclopedic text provides entity--attribute knowledge, scientific articles provide domain concepts and causal mechanisms, mathematical textbooks provide formal objects and transformations, and code repositories provide APIs, programming idioms, and implementation patterns. This perspective matters because many capability failures originate from insufficient coverage of the relevant content space: a model cannot reliably answer questions about an entity, concept, theorem, API, or domain if the corresponding content is absent, sparse, low-quality, or poorly connected within the training corpus.

The second perspective, \emph{discourse structure}, describes how information is organized within the data. It includes whether the text is narrative, expository, conversational, tabular, list-like, proof-like, or code-like; whether the relevant information is local or distributed across a long span; and whether the text contains explicit references, temporal structure, contrastive organization, examples, definitions, or derivation steps. This perspective is motivated by the observation that the same content world can support different learnable behaviors depending on how it is structured: a short encyclopedia paragraph mainly supports entity recall, whereas a long report with distributed evidence supports cross-paragraph aggregation; a list of mathematical formulas exposes symbolic objects, whereas a worked proof exposes transformation trajectories. Discourse structure thus determines how content is made available to the model, independently of what that content is.

The third perspective, \emph{operation opportunity}, describes the implicit operations that the text allows the model to learn, such as fact association, definition matching, comparison, counting, temporal ordering, causal attribution, evidence aggregation, symbolic transformation, algorithmic decomposition, code completion, and specification-to-implementation alignment. Operation opportunity is distinct from explicit task supervision: non-instruction data may never ask the model to perform an operation, yet the operation can still be embedded in the text. A financial report containing multiple comparable quantities creates opportunities for comparison and aggregation; a proof demonstrates symbolic transformation; a code file paired with comments exposes specification-to-code alignment. Among the four perspectives, this one provides the most direct bridge from raw text to the solving operations required by evaluation samples.

The fourth perspective, \emph{supervision density}, describes how strong, explicit, and recoverable the learning signal is under the language modeling objective. Some text offers dense, low-noise supervision, such as definitions followed by examples, worked solutions, aligned code comments, tables with clear headers, or derivations in which each step is locally supported. Other text offers weaker supervision, such as fragmented web pages, noisy forum discussions, ambiguous references, duplicated boilerplate, or passages in which the relevant relation is implicit and difficult to recover from context. Supervision density therefore captures not only whether useful content or operations are present, but also how efficiently they can be learned from next-token prediction; two documents may carry the same nominal information yet differ substantially in how learnable that information is.

Together, these four perspectives form a progression from raw content to learnable signal:
\[
\text{Content World}
\rightarrow
\text{Discourse Structure}
\rightarrow
\text{Operation Opportunity}
\rightarrow
\text{Supervision Density}.
\]
The content world determines what knowledge and symbolic objects are available; the discourse structure determines how these objects are arranged and connected; the operation opportunities determine what transformations, comparisons, derivations, or alignments are implicitly demonstrated by that arrangement; and the supervision density determines how strongly these signals are exposed to the model during training. This progression is useful in practice because data optimization should not only ask whether a corpus covers a given domain, but also whether the corpus presents that domain in a structure that supports the operations required by the target capability slice.

This progression also clarifies how the analysis of non-instruction data connects to the sample-level decomposition introduced in Section~\ref{subsec:understanding_evaluation}. The four perspectives used here are not identical to the four dimensions used to decompose evaluation samples; rather, in designing the two schemes, we sought to keep them broadly aligned wherever possible, though the correspondence is approximate rather than exact. The \emph{content world} of training data corresponds to the knowledge and domain component of the \emph{background condition} of evaluation samples, while the \emph{discourse structure} of training data corresponds to its structural component, namely the context scope, input format, and information organization that the same dimension also captures. The \emph{operation opportunity} in training data maps directly onto the \emph{solving operation} required to answer an evaluation sample. This approximate compatibility allows later sections to map evaluation failures to candidate data interventions: once a model failure is localized to a capability slice, we can search for non-instruction data that covers the relevant content world, presents it in the required discourse structure, embeds the necessary operation opportunities, and provides sufficiently dense supervision.

\subsubsection{Understanding Instruction Data}
\label{subsubsec:understanding_instruction_data}

Section~\ref{subsubsec:instruction_vs_non_instruction_data} characterizes instruction data as the component of the training corpus that converts raw content, solving operations, and output constraints into explicit, supervised behavior. We make this characterization concrete by analyzing instruction data along two complementary perspectives: an \emph{evaluation-side} perspective, which treats an instruction example as a training-time analogue of an evaluation sample, and a \emph{non-instruction-data-side} perspective, which treats it as a supervised reorganization of the latent signals already present in non-instruction data. These two perspectives correspond to the dual role instruction data plays in the framework: it specifies an explicit, task-like condition under which the model is expected to respond, while simultaneously repackaging existing content, structure, and operations into paired input--output supervision.

From the evaluation-side perspective, an instruction example exposes the same four dimensions used in Section~\ref{subsec:understanding_evaluation} to decompose evaluation samples. The input specifies the \emph{background condition}, such as a provided passage, dialogue history, table, code context, or closed-book setting, and the \emph{task type}, such as factual question answering, mathematical problem solving, code generation, information extraction, or tool-use invocation. The target response, in turn, demonstrates the required \emph{solving operation}—evidence aggregation, arithmetic computation, symbolic transformation, program synthesis, or step-by-step reasoning—and instantiates the corresponding \emph{output constraint}, including answer format, numerical normalization, JSON schema conformance, code executability, or citation style.

This perspective is useful because it connects instruction data directly to capability slices. Once an evaluation failure is localized to a specific combination of background condition, task type, solving operation, and output constraint, instruction data can be constructed to provide explicit supervision under the same or a neighboring condition. For instance, a failure on long-context passage QA that requires cross-paragraph evidence aggregation under short-form exact-match scoring motivates instruction data that pairs long documents with questions whose answers require distributed evidence, together with concise, normalized target outputs. In this sense, instruction data directly shapes how capabilities are activated and expressed under evaluation-like conditions.

From the non-instruction-data-side perspective, instruction data can instead be understood as a supervised reorganization of the latent signals characterized in Section~\ref{subsubsec:understanding_non_instruction_data}: \emph{content world}, \emph{discourse structure}, \emph{operation opportunity}, and \emph{supervision density}. Instruction construction selects and reformats these latent signals into explicit task supervision: an encyclopedic passage can be converted into factual or multi-hop question answering, a mathematical derivation into a problem--solution pair, API documentation into a code-generation task and a table into extraction, comparison, or arithmetic questions. Through this process, instruction construction increases the explicitness and density of the learning signal by aligning latent content and operations with an explicit objective and a target response.

This second perspective matters because instruction data is not defined merely by the surface presence of an imperative sentence or a question mark; its value depends on what latent signal it captures and how faithfully that signal is converted into supervision. Low-quality instruction data may specify a task while relying on shallow pattern matching, noisy answers, underspecified context, or weakly grounded reasoning. High-quality instruction data, by contrast, preserves the relevant content world, exposes the necessary discourse structure, renders the target operation observable, and produces a response whose format is aligned with the desired behavior. Instruction-data quality should therefore be assessed not only by task category, but also by the quality of the underlying signal from which the instruction is derived.

The two perspectives are complementary rather than competing. The evaluation-side view specifies what behavior an instruction example is intended to train: under what condition, for what task, using which operation, and under what output constraint. The non-instruction-data-side view specifies where the corresponding learning signal originates: what content is covered, how it is organized, what operation is latent or demonstrated, and how dense and reliable the supervision is. Together, the two views allow instruction data to serve as a bridge between evaluation failures and data construction; later sections use this bridge to map diagnosed capability weaknesses to both non-instruction-data retrieval criteria and instruction-data synthesis criteria.

\subsubsection{How Data Shapes Capability}
\label{subsubsec:how_data_shapes_capability}

The preceding analysis of non-instruction and instruction data (Sections~\ref{subsubsec:understanding_non_instruction_data} and~\ref{subsubsec:understanding_instruction_data}) suggests two complementary paths through which training data shapes model capability. The first is the \emph{non-instruction data path}, in which capability-forming signals are acquired indirectly from raw or weakly structured text under the language modeling objective. The second is the \emph{instruction data path}, in which task-like supervision more explicitly specifies the condition, objective, response pattern, and output constraint. These two paths should be understood as analytical abstractions rather than mutually exclusive mechanisms: in practice, instruction data often repackages signals that originate from non-instruction data, and non-instruction pretraining provides the representational substrate on which instruction tuning operates.

The non-instruction data path is indirect. Non-instruction data primarily shapes the substrate of capability by exposing the model, at scale, to its content world, discourse structure, operation opportunities, and supervision density:
\[
\begin{aligned}
\text{Non-instruction Data} &\rightarrow \text{Content World} \rightarrow \text{Discourse Structure} \\
&\rightarrow \text{Operation Opportunity} \rightarrow \text{Supervision Density} \rightarrow \text{Capability Substrate}.
\end{aligned}
\]
For example, consider a quarterly financial report. Its content world consists of company-specific entities, financial metrics, and accounting concepts; these are organized through a tabular discourse structure that aligns the same quantities across consecutive periods and segments. This structure in turn creates an operation opportunity for comparison and aggregation, and because each table carries clearly labeled headers and explanatory captions, these comparisons are also exposed with high supervision density.

However, even in this example, the report never explicitly instructs the model to compute a growth rate or compare two figures; the underlying operation is demonstrated, not assigned as an explicit task. More broadly, such signals are rarely presented as explicit task supervision: the model acquires them through next-token prediction alone, so the resulting capability may remain latent, incomplete, or difficult to activate under evaluation-like prompts. A model may be repeatedly exposed to comparisons of exactly this kind during pretraining and still fail to compute one reliably when explicitly queried at evaluation time. Non-instruction data is therefore best understood as providing broad capability-forming conditions rather than direct supervision for a specific evaluation slice.

Because this path is indirect, data--capability attribution is inherently noisy. A failure on a given capability slice cannot usually be traced to a single missing document or absent pattern; instead, the relevant question is whether the training corpus provides sufficient coverage, structural exposure, operation opportunities, and learnable signal density for the behavior that slice requires. For instance, improving symbolic reasoning may require not only mathematical content, but derivation-like text in which intermediate transformations are sufficiently dense and consistent. In this sense, non-instruction data shapes capability through cumulative distributional exposure rather than through any single, identifiable intervention.

The instruction data path is more direct. Instruction examples specify task-like conditions and desired responses, and therefore more closely resemble the evaluation samples used to observe model behavior:
\[
\begin{aligned}
\text{Instruction Data} &\rightarrow \text{Background Condition and Task Type} \rightarrow \text{Solving Operation Demonstration} \\
&\rightarrow \text{Output Constraint Alignment} \rightarrow \text{Behavior on Target Capability Slices}.
\end{aligned}
\]
From the evaluation-side perspective developed in Section~\ref{subsubsec:understanding_instruction_data}, an instruction example exposes the model, at training time, to background conditions, task types, solving operations, and output constraints analogous to those used during evaluation. This correspondence makes instruction data particularly effective for shaping how a capability is activated and expressed: whether the model follows the requested task, selects an appropriate reasoning mode, produces the required answer format, and satisfies the relevant scoring constraints.

This relationship is closely related to domain adaptation. If instruction data is closer to the evaluation distribution in task format, input structure, reasoning pattern, and output constraint, it may provide a more direct training signal for the corresponding capability slice. For example, instruction data constructed from long documents paired with short-answer questions may better match long-context QA evaluation than raw documents alone. Yet this similarity should not be interpreted as a sufficient causal explanation: performance may also depend on data quality, diversity, contamination control, optimization stage, loss weighting, and the capability substrate already established through non-instruction pretraining.

This caveat is not incidental: it follows from the dual nature of instruction data established in Section~\ref{subsubsec:understanding_instruction_data}. Instruction data is not an independent source of capability running parallel to non-instruction data, but a transformation applied on top of it. Returning to the financial-report example, an instruction pair built from that report, such as a question asking the model to compute quarter-over-quarter revenue growth, does not introduce a new comparison ability; it converts the comparison opportunity already latent in the table into an explicit, scored task. The instruction format makes the operation harder to ignore and easier to evaluate, but the comparison itself was already there to be learned, or not, from the underlying non-instruction signal. Consequently, instruction data inherits a ceiling from the corpus it draws on: if the source content is shallow, noisy, or weakly grounded, no amount of instruction-format refinement can manufacture a capability that was never available to learn. What such data produces instead is a model that exhibits the surface form of the target behavior without having reliably acquired the underlying operation. This gap often remains invisible until the model is evaluated under conditions slightly outside the instruction distribution.

Taken together, the two paths play complementary but interacting roles in shaping model capability. Non-instruction data primarily builds the broad substrate of knowledge, representations, discourse familiarity, and latent operations; instruction data more directly shapes how this substrate is activated, composed, and expressed under task-like conditions. This division of roles closes the engineering loop introduced in Section~\ref{sec:conceptual_framework}: once an evaluation failure has been localized to a capability slice (Section~\ref{subsec:understanding_model_capability}), diagnosing it requires considering both paths jointly: whether the non-instruction corpus provides sufficient latent signal for the target behavior, and whether the instruction corpus offers explicit supervision that matches the target evaluation slice without overfitting to benchmark-specific artifacts.

\section{Analysis Toolkit}
\label{sec:analysis_toolkit}

\subsection{Overview of the Toolkit}
\label{subsec:analysis_toolkit_overview}

\begin{figure}[t]
    \centering
    \includegraphics[width=0.7\linewidth]{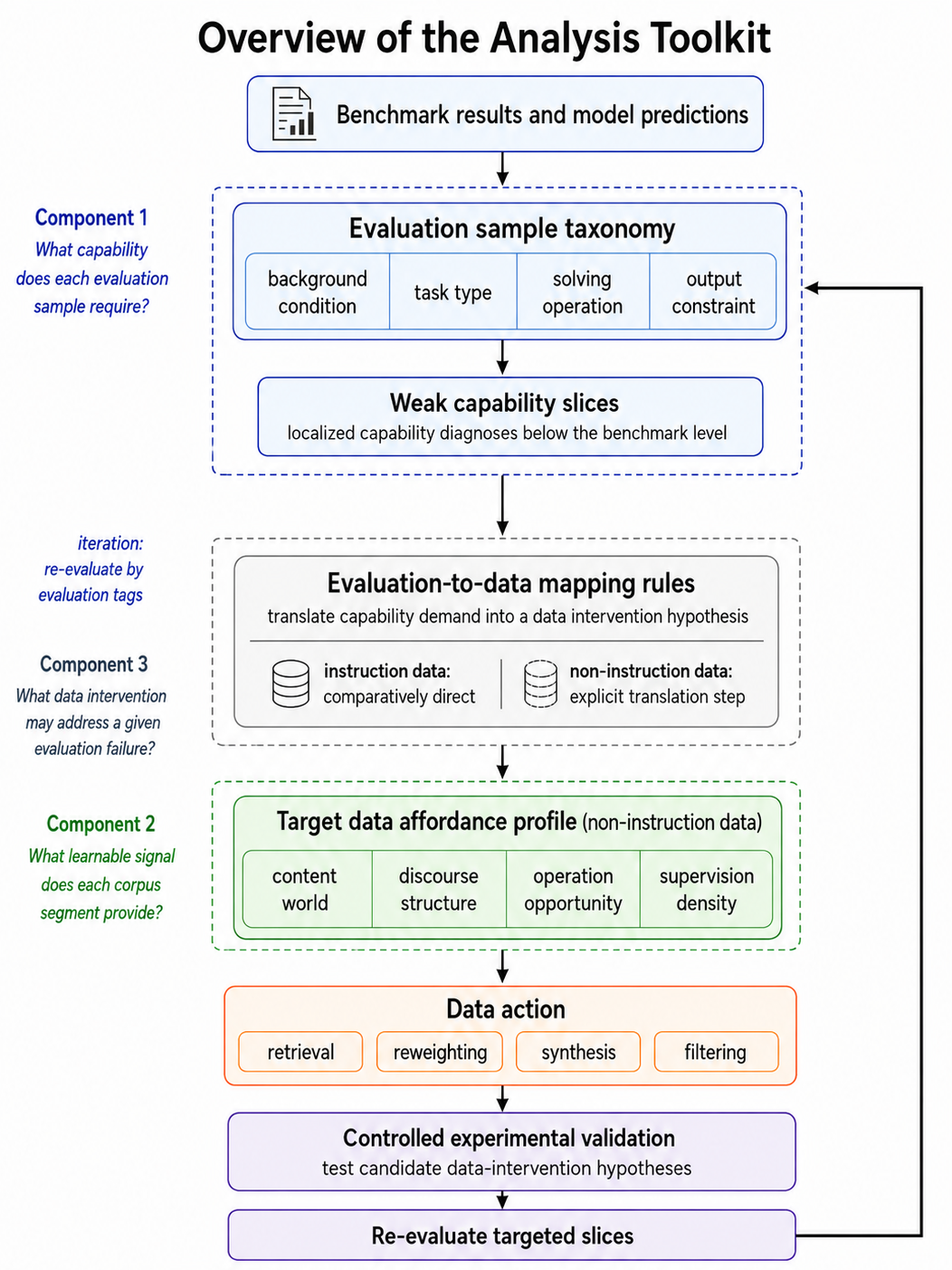}
    \caption{
    Overview of the analysis toolkit. The toolkit translates noisy benchmark-level observations into structured capability diagnoses, and these diagnoses, in turn, into candidate data-intervention hypotheses that can be validated experimentally.
    }
    \label{fig:analysis_toolkit_overview}
\end{figure}

The analysis toolkit operationalizes the conceptual framework introduced in Section~\ref{sec:conceptual_framework}, under which evaluation provides a noisy observation of model capability, while model capability is, in turn, shaped by training data. Its purpose is to convert benchmark-level observations into structured capability diagnoses and, from these diagnoses, into concrete data-intervention hypotheses that can be examined under controlled experimental conditions. As illustrated in Figure~\ref{fig:analysis_toolkit_overview}, the toolkit comprises three components: an evaluation sample taxonomy (Section~\ref{subsec:evaluation_sample_taxonomy}), a non-instruction data taxonomy (Section~\ref{subsec:non_instruction_data_taxonomy}), and a set of evaluation-to-data mapping rules (Section~\ref{subsec:evaluation_to_data_mapping_rules}) that connect the two.

The first component, the \textbf{evaluation sample taxonomy}, decomposes each benchmark sample into a structured description of its capability demand. Rather than treating a benchmark as a homogeneous measurement unit, we annotate each sample along four dimensions: background condition, task type, solving operation, and output constraint. This taxonomy answers the question: \emph{what capability does each evaluation sample require?} It allows benchmark results to be re-aggregated into capability slices below the level of the benchmark name, making it possible to attribute a model weakness to, for example, closed-book factual recall, cross-paragraph evidence aggregation, symbolic transformation, or strict output-formatting requirements, rather than to the benchmark as a whole.

The second component, the \textbf{non-instruction data taxonomy}, characterizes raw corpus segments by the learnable signals they may supply during pre-training. Non-instruction data is described along four complementary dimensions: content world, discourse structure, operation opportunity, and supervision density. This taxonomy answers the question: \emph{what learnable signal does each corpus segment provide?} It yields a data-side affordance profile that supports corpus composition analysis, quality-aware filtering, targeted data retrieval, and reweighting of existing data sources.

The third component, the \textbf{evaluation-to-data mapping rules}, connects the evaluation-side and data-side descriptions. Given a weak capability slice identified from evaluation results, the mapping rules translate its capability demand into a target data affordance profile. Because instruction data shares the same four-dimensional vocabulary used to describe evaluation samples, the corresponding mapping is comparatively direct, amounting to restating the slice specification as a synthesis target; non-instruction data, in contrast, is characterized by a structurally different taxonomy, so translating a weak capability slice into a target content world, discourse structure, operation opportunity, and supervision-density profile requires an explicit translation step. The mapping rules accordingly concentrate on this more indirect, non-instruction-data side of the bridge, combining a small set of taxonomy-aligned dimension correspondences with a complementary LLM-assisted mechanism for slices that fall outside them. This component answers the question: \emph{what data intervention may address a given evaluation failure?} Consistent with their role as hypothesis generators rather than as guarantees of improvement, the rules are treated as structured heuristics that guide data retrieval, reweighting, targeted synthesis, and filtering, with any resulting intervention remaining subject to downstream experimental validation.

Together, these three components instantiate a single closed-loop workflow: a weak capability slice is first localized through the evaluation sample taxonomy; the mapping rules translate it into a target data affordance profile; a corresponding data action is selected by inspecting the non-instruction data taxonomy's annotations over the existing corpus; and the resulting intervention is applied and evaluated under controlled experimental conditions before being mapped back onto the evaluation sample taxonomy to assess whether the targeted slice has improved. This design allows benchmark analysis to move beyond aggregate score comparison and turns the evaluation--data relationship into an actionable, auditable, and iterable optimization loop.
\subsection{Evaluation Sample Taxonomy}
\label{subsec:evaluation_sample_taxonomy}

\subsubsection{Construction and Annotation}
\label{subsubsec:evaluation_sample_taxonomy_construction_annotation}

We construct the taxonomy from samples drawn from sixteen benchmarks commonly used for evaluating pre-training base models: PIQA, HellaSwag, MMLU, TriviaQA, RACE, DROP, MMLU-Pro, BBH, C-Eval, CMMLU, AGIEval-CN, MBPP, HumanEval, GSM8K, MathQA, and MATH (Minerva). Together, these benchmarks cover commonsense plausibility judgment, factual question answering, passage understanding, mathematical problem solving, logical reasoning, and code generation. Sampling from this heterogeneous suite, rather than from any single benchmark, prevents the taxonomy from being tied to a particular benchmark format and allows it to capture capability demands that recur across benchmark boundaries.

The construction process begins with manual inspection of sampled items. For each item, annotators examine the question, the candidate options when available, the reference answer, and the scoring rule, with the goal of identifying the minimal set of conditions and operations required for a correct prediction. We organize the resulting capability requirements into the four dimensions introduced in Section~\ref{subsec:understanding_evaluation}: \emph{background condition}, \emph{task type}, \emph{solving operation}, and \emph{output constraint}.

We refine the taxonomy through multiple rounds of human--LLM co-design. In each round, we draft an annotation prompt, use LLMs to annotate a held-out set of benchmark samples, manually review the resulting annotations, and collect cases of disagreement at category boundaries. The prompt and the label definitions are then revised to reduce the recurrent confusions identified in this review. Typical revisions include clarifying the boundary between task type and solving operation, distinguishing parametric knowledge from contextual evidence, preventing over-labeling of reasoning operators, and aligning output-constraint labels with the benchmark's actual scoring rule rather than an idealized human grading criterion.

For large-scale annotation, each sample is independently labeled by two strong LLMs, DeepSeek-V3.2 and Qwen3.5 Plus, under the same prompt and the same input fields: the question or task instruction, the candidate options when available, the reference answer, and the scoring rule when it can be recovered from the benchmark protocol. We compare the two models' annotations category by category. Annotations on which the two models agree are accepted after a lightweight sanity check, while disagreements are routed to human review. Human reviewers inspect the original sample, the two model annotations, and their accompanying rationales, and then adjudicate the final labels according to the taxonomy's definitions and decision rules. The resulting adjudicated labels support benchmark composition analysis, weak-slice diagnosis, and the downstream mapping from evaluation to training data.

\subsubsection{Background Condition}
\label{subsubsec:evaluation_background_condition}
The background condition category describes the information environment under which the model is evaluated, answering the question: \textit{what information must the model condition on to solve the sample?} This category is necessary because samples that share the same task type may still differ substantially in their information source and presentation. For example, a factual question may be answerable from parametric knowledge alone, whereas a passage-QA item may require evidence drawn from a long context; similarly, a multiple-choice commonsense item may be solvable only by comparing the options themselves when the question stem provides little useful information on its own. To capture these differences, we decompose background condition into six subcategories: language, domain, parametric--contextual need, context scope, discourse form, and ambiguity noise. Each subcategory targets a distinct aspect of the information environment, ranging from the linguistic and domain setting of the sample to the scope, structure, and reliability of the evidence it provides. Table~\ref{tab:background_condition_label_space} summarizes the label space and diagnostic role of each subcategory, and Table~\ref{tab:background_condition_annotation_examples} lists representative annotated examples.

\begin{table}[t]
\centering
\small
\caption{Background condition labels. Domain labels are illustrative rather than exhaustive; the full domain taxonomy follows standard subject categorizations (e.g., the MMLU/C-Eval/CMMLU subject lists)}
\label{tab:background_condition_label_space}
\begin{tabularx}{\linewidth}{@{}L{0.18\linewidth}L{0.38\linewidth}Y@{}}
\toprule
\textbf{Subcategory} & \textbf{Label space} & \textbf{Diagnostic role} \\
\midrule
Language &
\labellist{
\code{zh}, \code{en}, \code{mixed}, \code{code\_mixed}
} &
Identifies the linguistic environment of the sample and supports language-specific slice analysis. \\
\midrule
Domain &
\labellist{
\code{mathematics}, \code{physics}, 
\\
\code{law}, \code{history}, \code{finance}, \code{others}
} &
Separates content coverage from task format and supports domain-conditioned aggregation across benchmarks. \\
\midrule
Parametric--contextual need &
\code{0}, \code{1}, \code{2}, \code{3} &
Distinguishes whether the answer depends primarily on parametric knowledge, on provided context, or on a combination of both, and helps separate knowledge gaps from context-utilization failures. \\
\midrule
Context scope &
\labellist{
\code{standalone}, \code{local\_span}, \code{multi\_sentence}, \\
\code{whole\_passage}, \code{cross\_option}
} &
Measures the span of information that must be integrated to solve the sample, distinguishing local-evidence reliance from long-context dependence. \\
\midrule
Discourse form &
\labellist{
\code{standalone\_question}, \code{dialogue}, \code{narrative}, \\
\code{expository}, \code{report\_news}, \code{list\_or\_table}, \\
\code{code\_spec}
} &
Describes the structural form in which the required information is presented, supporting slice analysis by input format such as dialogue, narrative, or tabular text. \\
\midrule
Ambiguity noise &
\labellist{
\code{low}, \code{medium}, \code{high}
} &
Captures surface ambiguity, unclear references, or irregular formatting that may interfere with solving, helping separate noise-induced errors from genuine capability failures. \\
\bottomrule
\end{tabularx}
\end{table}

\begin{table}[t]
\centering
\footnotesize
\caption{Examples of background condition annotations.}
\label{tab:background_condition_annotation_examples}
\renewcommand{\arraystretch}{1.15}
\begin{tabularx}{\linewidth}{@{}L{0.1\linewidth}L{0.4\linewidth}Y@{}}
\toprule
\textbf{Benchmark} & \textbf{Sample} & \textbf{Labels} \\
\midrule

\celltop{MMLU}
&
\celltop{
\textbf{Question:}
According to the Law of Effect, behaviors followed by negative consequences.

\medskip
\begin{tabular}{@{}ll@{}}
A. & occur more frequently \\
B. & occur less frequently \\
C. & will never be performed again \\
D. & will be performed more forcefully
\end{tabular}
}
&
\annlabels{
\textbf{Language}: \code{en} \\[1.5pt]
\textbf{Domain}: \code{psychology}, \code{education\_and\_training} \\[1.5pt]
\textbf{Parametric need}: \code{3} \\[1.5pt]
\textbf{Contextual need}: \code{0} \\[1.5pt]
\textbf{Context scope}: \code{standalone} \\[1.5pt]
\textbf{Discourse form}: \code{standalone\_question} \\[1.5pt]
\textbf{Ambiguity noise}: \code{low}
} \\

\midrule

\celltop{DROP}
&
\celltop{
\textbf{Passage:}
There were 1,882 households of which 21.4\% had children under the age of 18 living with them, 36.4\% were Marriage living together, 9.7\% had a female householder with no husband present, 3.6\% had a male householder with no wife present, and 50.4\% were non-families. 37.7\% of all households were made up of individuals and 13\% had someone living alone who was 65 years of age or older. The average household size was 2.20 and the average family size was 2.90.

\medskip
\textbf{Question:}
Which household was third most common?
}
&
\annlabels{
\textbf{Language}: \code{en} \\[1.5pt]
\textbf{Domain}: \code{family}, \code{mathematics} \\[1.5pt]
\textbf{Parametric need}: \code{0} \\[1.5pt]
\textbf{Contextual need}: \code{3} \\[1.5pt]
\textbf{Context scope}: \code{whole\_passage} \\[1.5pt]
\textbf{Discourse form}: \code{expository} \\[1.5pt]
\textbf{Ambiguity noise}: \code{low}
} \\

\midrule

\celltop{HumanEval}
&
\celltop{
\textbf{Sample:}

\medskip
{\ttfamily
def make\_a\_pile(n):\par
\hspace*{1em}"""\par
\hspace*{1em}Given a positive integer n, you have to make a pile of n levels of stones.\par
\hspace*{1em}The first level has n stones.\par
\hspace*{1em}The number of stones in the next level is:\par
\hspace*{2em}- the next odd number if n is odd.\par
\hspace*{2em}- the next even number if n is even.\par
\hspace*{1em}Return the number of stones in each level in a list, where element at index\par
\hspace*{1em}i represents the number of stones in the level (i+1).\par
\hspace*{1em}\par
\hspace*{1em}Examples:\par
\hspace*{1em}\textgreater{}\textgreater{}\textgreater{} make\_a\_pile(3)\par
\hspace*{1em}[3, 5, 7]\par
\hspace*{1em}"""
}
}
&
\annlabels{
\textbf{Language}: \code{code\_mixed} \\[1.5pt]
\textbf{Domain}: \code{information\_technology}, \code{mathematics} \\[1.5pt]
\textbf{Parametric need}: \code{3} \\[1.5pt]
\textbf{Contextual need}: \code{3} \\[1.5pt]
\textbf{Context scope}: \code{whole\_passage} \\[1.5pt]
\textbf{Discourse form}: \code{code\_spec}, \code{standalone\_question} \\[1.5pt]
\textbf{Ambiguity noise}: \code{low}
} \\

\bottomrule
\end{tabularx}
\end{table}

\subsubsection{Task Type}
\label{subsubsec:evaluation_task_type}
The task type category describes the primary user-facing objective of the sample, answering the question: \textit{what task is the model asked to perform?} It provides a coarse semantic grouping of evaluation samples, independent of benchmark identity, that captures what the sample is asking for rather than how the answer is to be derived. Task type is intentionally separated from solving operation: task type describes the final objective presented to the model, whereas solving operation describes the intermediate operations required to reach the answer. The same task type may therefore recur across different solving operations, and conversely, samples with superficially different task types may rely on the same underlying operation. Table~\ref{tab:task_type_label_space} lists the task type labels used in our taxonomy, and Table~\ref{tab:task_type_annotation_examples} lists representative annotated examples.

\begin{table}[t]
\centering
\small
\caption{Task type labels.}
\label{tab:task_type_label_space}
\begin{tabularx}{\linewidth}{@{}L{0.4\linewidth}Y@{}}
\toprule
\textbf{Task type} & \textbf{Description} \\
\midrule
\code{factual\_qa} &
Answering closed-book questions about objective facts, entities, relations, or events. \\
\midrule
\code{concept\_discrimination} &
Distinguishing, applying, or reasoning about concepts, definitions, categories, or principles. \\
\midrule
\code{passage\_understanding\_qa} &
Answering questions whose evidence must be located or integrated from an accompanying passage or context. \\
\midrule
\code{commonsense\_plausibility\_judgment} &
Selecting the option or continuation most plausible under commonsense, physical, or social constraints. \\
\midrule
\code{next\_event\_or\_next\_step\_prediction} &
Predicting the most likely next event, action, or continuation given a preceding context. \\
\midrule
\code{logical\_reasoning} &
Reasoning about truth conditions, implications, consistency, necessity, or other abstract relations. \\
\midrule
\code{arithmetic\_word\_problem} &
Solving natural-language problems that require arithmetic or numerical computation. \\
\midrule
\code{formal\_math\_problem\_solving} &
Solving formal mathematical problems involving algebra, geometry, symbolic derivation, or proof-like reasoning. \\
\midrule
\code{code\_generation\_from\_spec} &
Generating executable code from a natural-language or structured specification. \\
\midrule
\code{code\_comprehension\_or\_debugging} &
Understanding code behavior, identifying bugs, or reasoning about edge cases. \\
\midrule
\code{information\_extraction\_or\_transformation} &
Extracting, normalizing, mapping, or transforming information from a given input. \\
\bottomrule
\end{tabularx}
\end{table}

\begin{table}[t]
\centering
\footnotesize
\caption{Examples of task type annotations.}
\label{tab:task_type_annotation_examples}
\renewcommand{\arraystretch}{1.15}
\begin{tabularx}{\linewidth}{@{}L{0.1\linewidth}L{0.6\linewidth}Y@{}}
\toprule
\textbf{Benchmark} & \textbf{Question} & \textbf{Task type} \\
\midrule

\celltop{MMLU}
&
\celltop{
\textbf{Question:}
Say's Law

\medskip
\begin{tabular}{@{}ll@{}}
A. & is the basis of Keynesian economic analysis. \\
B. & is the basis of Classical economic analysis. \\
C. & states that demand creates its own supply. \\
D. & indicates that prices will be stable in capitalist economies.
\end{tabular}
}
&
\celltop{\code{factual\_qa}} \\

\midrule

\celltop{BBH}
&
\celltop{
\textbf{Question:}
Tamika tells the truth. Ka says Tamika tells the truth. Fidel says Ka tells the truth. Elanor says Fidel tells the truth. Amberly says Elanor tells the truth. Does Amberly tell the truth?
}
&
\celltop{\code{logical\_reasoning}} \\

\midrule

\celltop{GSM8K}
&
\celltop{
\textbf{Question:}
100 people apply for a job at Google. Of the people that apply, only 30\% receive interviews. Of those who receive interviews, 20\% receive a job offer. Of those who receive a job offer, a third of the people accept the position. How many people accept the position?
}
&
\celltop{\code{arithmetic\_word\_problem}} \\

\bottomrule
\end{tabularx}
\end{table}

\subsubsection{Solving Operation}
\label{subsubsec:evaluation_solving_operation}

The solving operation category describes the intermediate operations required to transform the input into the answer, answering the question: \textit{how is the answer derived?} Unlike task type, solving operation is multi-label, since a single sample may require a composition of operations; however, annotators are instructed to mark only the operations that lie on the minimal sufficient solving path, rather than every operation that could conceivably be applied. This category constitutes the main bridge between evaluation diagnosis and data intervention: performance on a benchmark may degrade not only because the model lacks the relevant knowledge, but also because it fails to aggregate evidence, track constraints, formulate equations, or handle algorithmic edge cases. By making these distinct failure sources observable at the sample level, operation labels allow a low score to be attributed to the specific operation underlying it, rather than to the task as a whole. Table~\ref{tab:solving_operation_label_space} summarizes the operation labels, and Table~\ref{tab:solving_operation_annotation_examples} lists representative annotated examples.

\begin{table}[t]
\centering
\small
\caption{Solving operation labels.}
\label{tab:solving_operation_label_space}
\begin{tabularx}{\linewidth}{@{}L{0.3\linewidth}Y@{}}
\toprule
\textbf{Solving Operation} & \textbf{Description} \\
\midrule
\code{fact\_recall} &
Recalling a fact from parametric knowledge, without relying on provided context. \\
\midrule
\code{concept\_alignment} &
Mapping an observed phenomenon, expression, or description onto the correct concept, definition, or principle. \\
\midrule
\code{span\_extraction} &
Locating and extracting a relevant evidence span directly from the provided context. \\
\midrule
\code{evidence\_aggregation} &
Combining multiple pieces of evidence, typically drawn from different sentences or regions of a passage. \\
\midrule
\code{compare\_or\_rank} &
Comparing, ranking, or identifying differences among entities, quantities, or options. \\
\midrule
\code{option\_elimination} &
Deriving the answer primarily by ruling out incorrect options rather than by direct computation. \\
\midrule
\code{multi\_hop\_composition} &
Chaining two or more reasoning steps or intermediate relations into a single derivation. \\
\midrule
\code{causal\_inference} &
Inferring or evaluating causal relations between events, actions, or variables. \\
\midrule
\code{temporal\_reasoning} &
Reasoning about event order, duration, or other temporal relations. \\
\midrule
\code{counting} &
Counting entities, occurrences, or events to obtain a numerical answer. \\
\midrule
\code{arithmetic\_computation} &
Performing arithmetic operations such as addition, subtraction, ratio computation, differencing, or summation. \\
\midrule
\code{equation\_formulation} &
Translating a natural-language description into mathematical relations or equations. \\
\midrule
\code{symbolic\_transformation} &
Performing algebraic, logical, or other symbolic transformations. \\
\midrule
\code{constraint\_tracking} &
Tracking multiple conditions, constraints, states, or intermediate results throughout a derivation. \\
\midrule
\code{physical\_state\_simulation} &
Simulating physical, spatial, object, or tool-use states to determine an outcome. \\
\midrule
\code{social\_script\_simulation} &
Simulating social situations, behavioral scripts, or plausible event continuations. \\
\midrule
\code{algorithm\_design} &
Designing program logic, data structures, or algorithmic procedures to satisfy a specification. \\
\midrule
\code{code\_execution\_simulation} &
Mentally executing code to reason about its behavior or output. \\
\midrule
\code{boundary\_case\_reasoning} &
Handling edge cases, exceptional inputs, or boundary conditions that the main solution path may not cover. \\
\bottomrule
\end{tabularx}
\end{table}

\begin{table}[t]
\centering
\footnotesize
\caption{Examples of solving operation annotations.}
\label{tab:solving_operation_annotation_examples}
\renewcommand{\arraystretch}{1.15}
\begin{tabularx}{\linewidth}{@{}L{0.1\linewidth}L{0.6\linewidth}Y@{}}
\toprule
\textbf{Benchmark} & \textbf{Question} & \textbf{Solving operation} \\
\midrule

\celltop{MMLU}
&
\celltop{
\textbf{Question:}
Which of the following styles of fuzzer is more likely to explore paths covering every line of code in the following program?

\medskip
\begin{tabular}{@{}ll@{}}
A. & Generational \\
B. & Blackbox \\
C. & Whitebox \\
D. & Mutation-based
\end{tabular}
}
&
\annlabels{
\code{concept\_alignment} \\[1.5pt]
\code{fact\_recall}
} \\

\midrule

\celltop{GSM8K}
&
\celltop{
\textbf{Question:}
Bubbles collects stuffed animals. She has three stuffed puppies, five stuffed koalas, two stuffed zebras and four stuffed frogs. If she wants to buy enough stuffed goats, such that the percentage of stuffed goats is 30\% of all of her stuffed animals, how many stuffed goats should she buy?
}
&
\annlabels{
\code{arithmetic\_computation} \\[1.5pt]
\code{equation\_formulation} \\[1.5pt]
\code{multi\_hop\_composition} \\[1.5pt]
\code{symbolic\_transformation}
} \\

\midrule

\celltop{HumanEval}
&
\celltop{
\textbf{Question:}

\medskip
{\ttfamily
def iscube(a):\par
\hspace*{1em}'''\par
\hspace*{1em}Write a function that takes an integer a and returns True\par
\hspace*{1em}if this ingeger is a cube of some integer number.\par
\hspace*{1em}Note: you may assume the input is always valid.\par
\hspace*{1em}Examples:\par
\hspace*{1em}iscube(1) ==\textgreater{} True\par
\hspace*{1em}iscube(2) ==\textgreater{} False\par
\hspace*{1em}iscube(-1) ==\textgreater{} True\par
\hspace*{1em}iscube(64) ==\textgreater{} True\par
\hspace*{1em}iscube(0) ==\textgreater{} True\par
\hspace*{1em}iscube(180) ==\textgreater{} False\par
\hspace*{1em}'''
}
}
&
\annlabels{
\code{algorithm\_design} \\[1.5pt]
\code{boundary\_case\_reasoning} \\[1.5pt]
\code{concept\_alignment}
} \\

\bottomrule
\end{tabularx}
\end{table}

\subsubsection{Output Constraint}
\label{subsubsec:evaluation_output_constraint}
The output constraint category describes the form in which the answer must be produced and the rule by which it is judged, answering the question: \textit{how must the answer be expressed and scored?} As the fourth and final dimension of the decomposition, it isolates a source of failure that is orthogonal to the preceding three: a model may correctly identify the relevant background condition, perform the intended task, and execute the required solving operation, yet still be scored incorrect because the generated output is unparseable, incorrectly normalized, numerically inequivalent to the reference, or non-executable. This category therefore separates failures of expression and scoring from failures of underlying capability. We decompose output constraint into six subcategories: answer form, answer space, scoring rule, format rigidity, exactness requirement, and explanation required, each capturing a distinct aspect of how the benchmark protocol constrains and evaluates the model's response. Table~\ref{tab:output_constraint_label_space} summarizes the label space and diagnostic role of each subcategory, and Table~\ref{tab:output_constraint_annotation_examples} lists representative annotated examples.

\begin{table}[t]
\centering
\small
\caption{Output constraint labels.}
\label{tab:output_constraint_label_space}
\begin{tabularx}{\linewidth}{@{}L{0.18\linewidth}L{0.38\linewidth}Y@{}}
\toprule
\textbf{Subcategory} & \textbf{Label space} & \textbf{Diagnostic role} \\
\midrule
Answer form &
\code{mcq\_single}, \code{mcq\_multi}, \code{extractive\_span}, \code{generated\_text}, \code{exact\_numeric}, \code{math\_expression}, \code{code}, \code{boolean} &
Describes the expected surface form of the final answer, and helps separate failures caused by mismatched output type from failures caused by incorrect content. \\
\midrule
Answer space &
\code{closed\_set}, \code{semi\_open}, \code{open} &
Indicates whether the answer is selected from a finite set, admits equivalent surface forms, or lies in an open generation space, and distinguishes selection-style failures from generation-style failures. \\
\midrule
Scoring rule &
\code{option\_match}, \code{exact\_match}, \code{normalized\_text\_match}, \code{semantic\_equivalence}, \code{numeric\_equivalence}, \code{unit\_test\_pass} &
Identifies the criterion the evaluator applies to judge correctness, and clarifies whether an observed error reflects incorrect content or a mismatch with the scoring procedure itself. \\
\midrule
Format rigidity &
\code{0}, \code{1}, \code{2}, \code{3} &
Measures how sensitive the evaluator is to surface-level formatting deviations, and helps isolate formatting-induced failures from genuine reasoning errors. \\
\midrule
Exactness requirement &
\code{0}, \code{1}, \code{2}, \code{3} &
Measures how precisely the generated answer must match the reference under the benchmark scorer, and separates failures of precision from failures of reasoning. \\
\midrule
Explanation required &
\code{no}, \code{optional}, \code{yes} &
Specifies whether the scoring protocol requires an accompanying explanation or rationale, and clarifies whether a missing explanation should itself be counted as an error. \\
\bottomrule
\end{tabularx}
\end{table}

\begin{table}[t]
\centering
\footnotesize
\caption{Examples of output constraint annotations.}
\label{tab:output_constraint_annotation_examples}
\renewcommand{\arraystretch}{1.15}
\begin{tabularx}{\linewidth}{@{}L{0.1\linewidth}L{0.3\linewidth}L{0.2\linewidth}Y@{}}
\toprule
\textbf{Benchmark} & \textbf{Question} & \textbf{Golden answer} & \textbf{Labels} \\
\midrule

\celltop{MMLU}
&
\celltop{
\textbf{Question:}
What is the optimal span of control?

\medskip
\begin{tabular}{@{}ll@{}}
A. & 2 \\
B. & 5 \\
C. & 7 \\
D. & None of the above
\end{tabular}
}
&
\celltop{D}
&
\annlabels{
\textbf{Answer form}: \code{mcq\_single} \\[1.5pt]
\textbf{Answer space}: \code{closed\_set} \\[1.5pt]
\textbf{Scoring rule}: \code{option\_match} \\[1.5pt]
\textbf{Format rigidity}: \code{0} \\[1.5pt]
\textbf{Exactness requirement}: \code{0} \\[1.5pt]
\textbf{Explanation required}: \code{no}
} \\

\midrule

\celltop{GSM8K}
&
\celltop{
\textbf{Question:}
While playing with her friends in their school playground, Katelyn saw 50 fairies flying above the nearby forest. After about twenty minutes, one of her friends saw half as many fairies as Katelyn saw come from the east and join the fairies that were there. Ten minutes later, 30 fairies flew away. How many fairies are remaining?
}
&
\celltop{45}
&
\annlabels{
\textbf{Answer form}: \code{exact\_numeric} \\[1.5pt]
\textbf{Answer space}: \code{semi\_open} \\[1.5pt]
\textbf{Scoring rule}: \code{numeric\_equivalence} \\[1.5pt]
\textbf{Format rigidity}: \code{1} \\[1.5pt]
\textbf{Exactness requirement}: \code{3} \\[1.5pt]
\textbf{Explanation required}: \code{no}
} \\

\midrule

\celltop{HumanEval}
&
\celltop{
\textbf{Question:}

\medskip
{\ttfamily
def add(x: int, y: int):\par
\hspace*{1em}"""Add two numbers x and y\par
\hspace*{1em}>\!>\!> add(2, 3)\par
\hspace*{1em}5\par
\hspace*{1em}>\!>\!> add(5, 7)\par
\hspace*{1em}12\par
\hspace*{1em}"""
}
}
&
\celltop{
{\ttfamily
return x + y
}
}
&
\annlabels{
\textbf{Answer form}: \code{code} \\[1.5pt]
\textbf{Answer space}: \code{semi\_open} \\[1.5pt]
\textbf{Scoring rule}: \code{unit\_test\_pass} \\[1.5pt]
\textbf{Format rigidity}: \code{0} \\[1.5pt]
\textbf{Exactness requirement}: \code{3} \\[1.5pt]
\textbf{Explanation required}: \code{no}
} \\

\bottomrule
\end{tabularx}
\end{table}

\subsubsection{Usage}
\label{subsubsec:evaluation_sample_taxonomy_usage}

The four-dimensional taxonomy developed above is not an end in itself; its value lies in how it converts the sample-level decomposition introduced in Section~\ref{subsec:understanding_evaluation} into concrete analysis operations. In this paper, we use the taxonomy in three complementary ways: to characterize what a benchmark actually measures, to localize where a model's behavior is unstable, and to explain why a benchmark score changes after an intervention. The three uses share the same underlying mechanism—aggregating sample-level labels into structured groups—but differ in what is held fixed and what is allowed to vary.

The first use is to analyze the capability composition of the sixteen benchmarks introduced in Section~\ref{subsubsec:evaluation_sample_taxonomy_construction_annotation}. Rather than treating each benchmark as a homogeneous capability probe, we compute, for each benchmark, the distribution of background conditions, task types, solving operations, and output constraints over its constituent samples. This profile makes explicit, for example, whether a benchmark is dominated by closed-book factual recall, by passage-grounded evidence extraction, by arithmetic computation, by code generation, or by strict, parser-sensitive answer formats. Because this analysis operates below the benchmark name, it also exposes cases in which a single benchmark mixes several distinct capability demands, or in which two differently named benchmarks turn out to probe a similar underlying slice.

The second use is to support the discovery and analysis of weak capability slices, instantiating the notion of a capability slice introduced in Section~\ref{subsec:understanding_model_capability}. Given model predictions and sample-level correctness, we aggregate performance over groups of samples that share similar labels along the four dimensions, rather than over an entire benchmark or an isolated item, and examine these groups to identify slices on which performance is consistently and non-trivially low. A weak capability slice can then be represented as a structured tuple over the four dimensions, for example:
\[
(\texttt{whole\_passage},\;
\texttt{passage\_understanding\_qa},\;
\texttt{evidence\_aggregation},\;
\texttt{exact\_match}).
\]
Such a slice is more actionable than a benchmark name because it specifies the background condition, task type, solving operation, and output constraint under which the model consistently fails, rather than a single aggregate score that conflates many such conditions. This localization step is the analytical core of the second use: it converts a diffuse, benchmark-level weakness into a small number of well-specified, structurally homogeneous failure patterns that can be examined and prioritized individually. Once localized, these weak slices serve as the direct input to the evaluation-to-data mapping rules introduced in Section~\ref{subsec:evaluation_to_data_mapping_rules}, which translate each slice into a concrete optimization direction together with a corresponding data or supervision strategy. In this way, the taxonomy not only localizes where the model fails but also constrains what kind of data intervention is likely to address that failure.

The third use is to explain benchmark changes after optimization. When a model improves or regresses on a benchmark, we decompose the resulting score delta along the same four dimensions rather than reading it as a single undifferentiated signal. This decomposition allows us to attribute an observed gain or drop to a specific source, for example, improved factual recall, more reliable arithmetic computation, better code generation, or a reduction in format-sensitive output failures, and to distinguish such capability-level changes from changes driven mainly by output-constraint compliance. In this way, the taxonomy closes the loop between evaluation and data: it is first used to diagnose weak capability slices, then to map these slices to candidate data or supervision interventions, and finally to verify, after intervention, whether the resulting benchmark improvement originates from the targeted capability slice or from an unrelated source. The case studies in Section~\ref{sec:case_study} demonstrate these three uses.
\subsection{Non-instruction Data Taxonomy}
\label{subsec:non_instruction_data_taxonomy}

\subsubsection{Construction and Annotation}
\label{subsubsec:non_instruction_data_taxonomy_construction_annotation}

The non-instruction data taxonomy is designed to characterize the learnable signal exposed by raw pre-training text. Unlike evaluation samples, which are organized around an explicit task and a scoring rule, non-instruction samples carry no separated task input or target response; they supervise the model only through the language modeling objective. Annotating such samples therefore requires more than a topical label: the taxonomy must describe what a corpus segment is about, how its information is organized, what operation-like patterns it exposes, and whether the resulting signal is sufficiently clean and informative to serve as effective next-token supervision.

We design the taxonomy under three constraints. First, it should reflect properties intrinsic to non-instruction data rather than properties imported from a downstream task. We accordingly organize the data-side description into four dimensions, \emph{content world}, \emph{discourse structure}, \emph{operation opportunity}, and \emph{supervision density}, which respectively characterize the topical and referential content of a sample, the way its information is arranged and connected, the capability-forming operations explicitly present in the text, and the recoverability and informativeness of the resulting supervision signal. Second, the taxonomy should remain compatible with the evaluation-side decomposition introduced in Section~\ref{subsec:understanding_evaluation} and instantiated in Section~\ref{subsec:evaluation_sample_taxonomy}. Since the overall framework aims to close the loop between evaluation failures and data interventions, the four data-side dimensions are deliberately designed to correspond, where such correspondence is meaningful, to the evaluation-side dimensions of \emph{background condition}, \emph{task type}, \emph{solving operation}, and \emph{output constraint}. Third, the taxonomy should remain compatible with the data annotations already used in our data pipeline, including existing domain labels and quality-related labels. Rather than replacing these annotations, we incorporate them into a broader structure in which they serve as reusable signals for corpus composition analysis, filtering, and targeted data construction.

For large-scale annotation, we primarily use compact LLM annotators such as Qwen3 30B-A3B or Gemma4 26B-A4B, rather than conventional text classifiers such as fastText- or BERT-style models. This choice follows from the structure of the annotation problem itself: several dimensions, most notably discourse structure and operation opportunity, require the annotator to reason over multi-sentence dependencies, separate topic from operation, and apply boundary rules that shallow or short-context classifiers struggle to capture. In preliminary prompt-development iterations, compact LLMs produced more faithful boundary decisions and generalized more reliably across domains and discourse forms, whereas smaller discriminative classifiers were more prone to overfitting the annotation examples, missing long-range dependencies, and collapsing structurally complex samples into coarse topic-level labels. At the same time, recent improvements in LLM inference systems have made compact-LLM annotation practical at the scale required for large-scale corpus processing. We therefore use compact LLMs as the main annotation engine, with rule-based checks, sampled human audits, and distribution-level sanity checks used to monitor annotation quality.

The annotation prompts are developed through multiple rounds of human--LLM co-design, following the same general protocol used for the evaluation sample taxonomy in Section~\ref{subsubsec:evaluation_sample_taxonomy_construction_annotation}. In each round, we draft or revise the prompt, annotate a held-out set of corpus segments, manually inspect the resulting labels, and collect recurring boundary errors; the label definitions, decision rules, and few-shot examples are then revised to reduce these errors. Typical revisions include requiring discourse-structure annotation to first identify the sample's main learnable signal before assigning context-scope labels, selecting few-shot examples that more clearly distinguish local from passage-level dependency, and refining the operation-opportunity taxonomy so that its first-level types have sharper boundaries and its second-level sub-types capture the most common operational forms without encouraging over-labeling.

\subsubsection{Content World}
\label{subsubsec:non_instruction_content_world}

The \emph{content world} dimension characterizes what a non-instruction sample is about, abstracting the topical, referential, and distributional semantics of raw text into a small set of complementary labels. Unlike instruction-side annotations, which primarily characterize task form and response behavior, content-world annotations are designed to support corpus composition analysis, data mixture design, and targeted data supplementation. We operationalize this dimension through three complementary subcategories: domain labels, entity--concept labels, and text embeddings. Table~\ref{tab:content_world_description} summarizes these subcategories together with their diagnostic role within the content-world dimension. Table~\ref{tab:content_world_examples} complements this summary with representative annotated examples.

The two-level domain taxonomy is clearly structured, with the coarse-grained and fine-grained tiers deliberately serving distinct purposes. The fine-grained labels are identical to the domain labels used in the background condition dimension of the evaluation sample taxonomy (Section~\ref{subsubsec:evaluation_background_condition}), so that data-side and evaluation-side annotations share a single label space and directly support the evaluation-to-data mapping rule (Section~\ref{subsec:evaluation_to_data_mapping_rules}). The coarse-grained and fine-grained labels are not assigned sequentially; rather, both are produced jointly under a fixed coarse-to-fine mapping that constrains which fine-grained label is admissible under each coarse-grained label, ensuring the two tiers remain mutually consistent by construction rather than one being derived from the other after the fact. This coarse-grained tier nonetheless yields a smaller, more manageable category set that is convenient for data staging, tabulation, and mixture-ratio experimentation. Taken together, the domain subcategory provides the corpus's primary topical-coverage signal, linking evaluation-defined capability slices to controllable units of data composition and mixture design.

The entity--concept annotation extracts salient referential and conceptual units from the text. Entities correspond to concrete, referential objects or named items, such as persons, organizations, locations, products, events, laws, technologies, or scientific objects, whereas concepts correspond to abstract, reusable knowledge units, such as theories, methods, processes, properties, phenomena, skills, or social issues. Compared with the domain subcategory, which situates a sample within a small set of coarse and fine topical categories, entity--concept annotation operates at a more granular level, surfacing the specific instance-level units that populate a given domain and providing more actionable signal for entity- and concept-level grouping and clustering. Taken together, the entity--concept subcategory provides the corpus's primary fine-grained referential and conceptual signal, complementing the topical-coverage signal of the domain subcategory and supporting targeted coverage analysis and data construction at the level of individual entities and concepts.

Beyond these discrete labels, each sample is also encoded by a lightweight embedding model into a fixed-dimensional vector representation. Although continuous rather than categorical, this embedding can be viewed as a complementary, latent content-world signal: samples with similar embeddings tend to share topical or referential content even when they are not assigned identical domain or entity--concept labels. Taken together, the embedding subcategory provides the corpus's primary continuous content-similarity signal, complementing the discrete domain and entity--concept annotations and supporting similarity-based clustering, near-duplicate detection, and retrieval of topically or referentially related samples.

\begin{table*}[t]
\centering
\small
\caption{Content World annotation subcategories for non-instruction data.}
\label{tab:content_world_description}
\renewcommand{\arraystretch}{1.15}
\begin{tabularx}{\textwidth}{@{}L{0.18\textwidth}L{0.26\textwidth}Y@{}}
\toprule
\textbf{Subcategory} & \textbf{Field} & \textbf{Diagnostic role} \\
\midrule

\celltop{Domain}
&
\celltop{Primary coarse domain}
&
\celltop{Assigns the dominant high-level semantic domain of the sample from a fixed set of coarse domains. This field provides a low-dimensional signal for data mixture design, broad corpus stratification, and high-level distribution monitoring.} \\
\cmidrule(lr){2-3}

&
\celltop{Primary fine domain}
&
\celltop{Assigns the dominant fine-grained topical domain, nested under the primary coarse domain. This field supports detailed corpus analysis, domain-specific error attribution, diversity diagnostics, and targeted data supplementation.} \\
\cmidrule(lr){2-3}

&
\celltop{Secondary fine domains}
&
\celltop{Records up to three additional fine-grained domains when substantial secondary topics are present, capturing mixed-domain composition that a single dominant label would otherwise obscure, without weakening the primary topical assignment.} \\

\midrule

\celltop{Entity--Concept}
&
\celltop{Entities}
&
\celltop{Extracts up to five salient, concrete referential items that are explicitly mentioned or strongly implied in the text. Entity labels support entity-centered clustering and help identify overrepresented or undercovered referential regions of the corpus.} \\
\cmidrule(lr){2-3}

&
\celltop{Concepts}
&
\celltop{Extracts up to five salient abstract knowledge units grounded in the text. Concept labels provide a reusable semantic signal for concept-level clustering, diversity analysis, and targeted coverage improvement.} \\

\midrule

\celltop{Embedding}
&
\celltop{Text embedding}
&
\celltop{Encodes the sample into a fixed-dimensional vector using a lightweight embedding model. The resulting representation captures topical and referential similarity beyond what the discrete domain and entity--concept labels express, and supports similarity-based clustering, near-duplicate detection, and retrieval of related samples.} \\

\bottomrule
\end{tabularx}
\end{table*}

\begin{table}[t]
\centering
\footnotesize
\caption{Examples of Content World annotations}
\label{tab:content_world_examples}
\renewcommand{\arraystretch}{1.15}
\begin{tabularx}{\linewidth}{@{}L{0.4\linewidth}Y@{}}
\toprule
\textbf{Text} & \textbf{Labels} \\
\midrule

\celltop{
Diynamic Music's "Picture:" series is a document in time, showcasing an artist's current creative output in order to allow the listener to visualise and understand the artist at a certain moment. It is not an album format, but far more than merely an EP. "Picture:" is various dancefloor-oriented facets of one producer, shaping the imagery that ultimately results in the listeners mind.
}
&
\annlabels{
\textbf{Primary coarse domain}: \code{culture\_media\_humanities} \\
\textbf{Primary fine domain}: \code{Music} \\
\textbf{Secondary fine domains}: \code{Art} \\
\textbf{Entities}: \\
\hspace*{1em}\code{Diynamic Music} \\
\hspace*{1em}\code{Picture: series} \\
\textbf{Concepts}: \\
\hspace*{1em}\code{creative output} \\
\hspace*{1em}\code{album format} \\
\hspace*{1em}\code{EP} \\
\hspace*{1em}\code{dancefloor-oriented music} \\
} \\

\midrule

\celltop{
Established in 1937 by Alfonso Morini, a former racer and skilled designer, Moto Morini gained global renown through its successful participation in international races. By the late 1950s, the company's racing engines were integrated into its standard production motorcycles. The pinnacle of Moto Morini's innovation and technological research came in 1973 with the triumph of the 3 1/2 model.
}
&
\annlabels{
\textbf{Primary coarse domain}: \code{lifestyle\_consumption\_recreation} \\
\textbf{Primary fine domain}: \code{Transportation} \\
\textbf{Secondary fine domains}: \code{History}, \code{Products} \\
\textbf{Entities}: \\
\hspace*{1em}\code{Moto Morini} \\
\hspace*{1em}\code{Alfonso Morini} \\
\hspace*{1em}\code{3 1/2 model} \\
\textbf{Concepts}: \\
\hspace*{1em}\code{motorcycle manufacturing} \\
\hspace*{1em}\code{motorcycle racing} \\
\hspace*{1em}\code{technological research} \\
\hspace*{1em}\code{engine design} \\
} \\

\bottomrule
\end{tabularx}
\end{table}

\subsubsection{Discourse Structure}
\label{subsubsec:non_instruction_discourse_structure}

The \emph{discourse structure} dimension characterizes how a non-instruction sample is organized as a discourse object: whereas \emph{content world} (Section~\ref{subsubsec:non_instruction_content_world}) specifies what a sample is about, discourse structure specifies how that content is presented, connected, and made accessible to the learner, instantiating the second stage of the content-world~$\rightarrow$~discourse-structure~$\rightarrow$~operation-opportunity~$\rightarrow$~supervision-density progression introduced in Section~\ref{subsubsec:understanding_non_instruction_data}. We operationalize this dimension through two complementary subcategories: \emph{discourse form}, which characterizes the sample's external surface format, and \emph{context scope}, which characterizes the internal dependency structure required to recover the signal carried by that form. Table~\ref{tab:discourse_structure_description} summarizes the fields and diagnostic roles of this dimension, and Table~\ref{tab:discourse_structure_examples} provides representative annotated examples.

The discourse-form subcategory identifies the dominant discourse format of a visible text chunk, such as an article, dialogue, table, code fragment, or noisy mixed-format record. Different discourse forms expose the model to different surface conventions, structural regularities, and generation regimes, including turn-taking structure in dialogue, hierarchical organization in tables, or symbolic regularity in code. Because a single chunk may exhibit more than one coherent form, for example a worked example embedded in an expository article, we record one primary form together with up to two secondary forms. Taken together, the discourse-form subcategory provides the corpus's primary signal for surface-level format composition, supporting form-stratified corpus analysis and form-targeted data supplementation.

The context-scope subcategory diagnoses how much contextual integration is required to recover a sample's main learnable signal, i.e., the specific knowledge, relation, process, comparison, or argument structure that the sample is judged to primarily convey, as opposed to a generic topic summary. We decompose this subcategory into three fields. The dependency-pattern field describes the type of relation linking different parts of the text, such as entity tracking, causal chaining, or comparison. The dependency-scope field describes the minimal span over which the signal is distributed, ranging from a single self-contained sentence to the full passage. The dependency-density field describes how frequently and how strongly such dependencies recur throughout the sample. Separating these three fields allows the form, extent, and prevalence of contextual dependency to be assessed independently. Taken together, the context-scope subcategory provides the corpus's primary signal for the internal dependency structure of its samples, complementing the surface-level characterization given by discourse form and supporting dependency-aware data curation for capability slices.

\begin{table*}[t]
\centering
\small
\caption{Discourse Structure annotation subcategories for non-instruction data.}
\label{tab:discourse_structure_description}
\renewcommand{\arraystretch}{1.15}
\begin{tabularx}{\textwidth}{@{}L{0.18\textwidth}L{0.26\textwidth}Y@{}}
\toprule
\textbf{Subcategory} & \textbf{Field} & \textbf{Diagnostic role} \\
\midrule

\celltop{Discourse Form}
&
\celltop{Primary form}
&
\celltop{Identifies the sample's dominant surface format (e.g., \code{narrative\_prose}, \code{dialogue\_or\_transcript}, \code{code\_or\_api\_spec}), independently of the domain-level content captured by \emph{content world}: the same form can host many domains, and the same domain can appear in many forms.} \\
\cmidrule(lr){2-3}

&
\celltop{Secondary forms}
&
\celltop{Records up to two additional coherent formats present in the sample (e.g., a \code{procedural\_or\_howto} segment embedded in an \code{expository\_article}), capturing mixed-format composition that a single primary label would otherwise obscure.} \\

\midrule

\celltop{Context Scope}
&
\celltop{Dependency patterns}
&
\celltop{Identifies the type of relation linking different parts of the text, such as \code{entity\_tracking}, \code{causal\_chain}, or \code{comparison\_contrast}.} \\
\cmidrule(lr){2-3}

&
\celltop{Dependency scope}
&
\celltop{Measures the minimal span required to recover the main learnable signal, ranging from \code{standalone} and \code{local\_span} to \code{multi\_sentence} and \code{whole\_passage} dependency.} \\
\cmidrule(lr){2-3}

&
\celltop{Dependency density}
&
\celltop{Measures how pervasively dependency recurs within the sample, on a four-point scale from \code{0} (almost no contextual dependency) to \code{3} (dense, pervasive dependency).} \\

\bottomrule
\end{tabularx}
\end{table*}

\begin{table}[t]
\centering
\footnotesize
\caption{Examples of Discourse Structure annotations}
\label{tab:discourse_structure_examples}
\renewcommand{\arraystretch}{1.15}
\begin{tabularx}{\linewidth}{@{}L{0.5\linewidth}Y@{}}
\toprule
\textbf{Text} & \textbf{Labels} \\
\midrule

\celltop{
In the wake of the 2014 midterm elections, Rush Limbaugh argues that the Republican Party has a significant mandate, whether they recognize it or not. He believes their victory, if achieved, will be primarily due to voters' desire to stop President Barack Obama's policies. This mandate is not one the Republicans have explicitly campaigned on, but rather, it's the implicit reason behind voters' decisions to support them. Limbaugh predicts that Obama, emboldened by the media's support, will interpret a Republican victory as a national crisis, using it to justify his planned executive actions. Despite the Republican Party's strategic silence during the campaign, Limbaugh asserts that a victory would grant them a substantial mandate to halt Obama's policies and those of the Democratic Party.
}
&
\annlabels{
\textbf{Primary form}: \code{expository\_article} \\
\textbf{Secondary forms}: \code{opinion\_or\_argumentation} \\
\textbf{Dependency scope}: \code{whole\_passage} \\
\textbf{Dependency density}: \code{3} \\
\textbf{Dependency patterns}: \\
\hspace*{1em}\code{argument\_chain} \\
\hspace*{1em}\code{causal\_chain} \\
\hspace*{1em}\code{coreference\_resolution} \\
} \\

\midrule

\celltop{
Painting water is a fairly complex subject. Water, is affected by its surroundings, reflections, depth and clarity. The painting of oceans, rivers, lakes and ponds can be beautiful, but managing to get the water and reflections to look like actual water can be challenging. Reflections in Muddy Water: Layin' Drag on Life's Highway in Cassville, Georgia Paperback – J by Brad Stephens (Author)/5(23). Claude Debussy 's Reflets dans l'eau ("Reflections in the Water") is the first of three piano pieces from his first volume of Images, which are frequently performed separately.
Begin by focusing first on your subject; then, move on to your reflection in the water. This is going to give you subtly different results, nothing too drastic, but enough for you to decide which style you like better. The Poetry book I read was the Reflections on a Gift of a Watermelon pickle. Undercoat the whole water area with white and use plenty of paint. Blend your blue into the white from the bottom up while your white undercoat is still wet. This is the reflection of the sky dark blue at the bottom and white in the distance.
Paint the. Reflection off of smooth surfaces such as mirrors or a calm body of water leads to a type of reflection known as specular reflection. Reflection off of rough surfaces such as clothing, paper, and the asphalt roadway leads to a type of reflection known as diffuse reflection.
}
&
\annlabels{
\textbf{Primary form}: \code{mixed\_or\_fragmented} \\
\textbf{Secondary forms}: \code{qa\_or\_faq}, \code{procedural\_or\_howto} \\
\textbf{Dependency scope}: \code{standalone} \\
\textbf{Dependency density}: \code{0} \\
\textbf{Dependency patterns}: \\
\hspace*{1em}\code{none} \\
} \\

\bottomrule
\end{tabularx}
\end{table}

\subsubsection{Operation Opportunity}
\label{subsubsec:non_instruction_operation_opportunity}

The \emph{operation opportunity} dimension characterizes which capability-forming operations are explicitly exposed by a non-instruction sample. This dimension provides the most direct bridge to the solving-operation dimension of the evaluation sample taxonomy (Section~\ref{subsubsec:evaluation_solving_operation}): many evaluation-side solving operations can only be reliably strengthened if the training corpus exposes a corresponding operational pattern. This correspondence is nonetheless asymmetric. An evaluation sample requires the model to execute an operation in order to produce a correct answer, whereas a non-instruction sample merely exposes the textual conditions under which such an operation can be observed and, potentially, internalized through the language modeling objective. An assigned operation-opportunity label therefore indicates that a learnable operational pattern is explicitly present in the text, not that the operation has been supervised, exercised, or successfully acquired by the model. Table~\ref{tab:operation_opportunity_description} summarizes the resulting taxonomy, and Table~\ref{tab:operation_opportunity_examples} provides representative annotated examples.

We operationalize this dimension as nine independent, multi-label operation types rather than as a single dominant label, since a sample may simultaneously expose several distinct operational patterns. For each type, we judge whether the corresponding pattern is explicitly present in the text; whenever a type is judged present, we assign a sub-type label that specifies its particular form, such as \code{comparison\_or\_ranking} under \code{comparison\_contrast\_and\_exclusion} or \code{causal\_relation\_or\_mechanism} under \code{temporal\_causal\_and\_process\_reasoning}.

The primary analytical value of this taxonomy lies in its correspondence to the solving-operation dimension of the evaluation sample taxonomy (Section~\ref{subsubsec:evaluation_solving_operation}): each operation type is named and defined so that it can be interpreted as a candidate source of training signal for one or more solving operations observed at evaluation time. For instance, the evaluation-side operation \code{fact\_recall} corresponds most naturally to \code{fact\_or\_entity\_relation} or \code{concept\_definition\_or\_explanation} under \code{knowledge\_and\_concept\_grounding}, since both describe a sample whose content is grounded in a single recalled fact or definition rather than in contextual evidence or derivation. Similarly, \code{compare\_or\_rank} corresponds most naturally to \code{comparison\_or\_ranking} or \code{contrastive\_distinction} under \code{comparison\_contrast\_and\_exclusion}, since both describe a sample structured around evaluating two or more entities, quantities, or options against one another. We emphasize that such correspondences are intuitive and heuristic rather than definitive or universally guaranteed: a given solving operation may plausibly be supported by more than one operation-opportunity sub-type, a given sub-type may relate to more than one solving operation, and the alignment need not hold for every individual sample. The mapping is intended as a practical starting point for connecting evaluation-side failure diagnosis to data-side retrieval criteria, rather than as a formal or one-to-one equivalence between the two label spaces. Taken together, the nine operation types provide the corpus's primary signal for the latent reasoning and processing opportunities embedded in non-instruction text, complementing the topical coverage captured by content world and the structural organization captured by discourse structure.

\begin{table*}[t]
\centering
\footnotesize
\setlength{\tabcolsep}{4pt}
\renewcommand{\arraystretch}{0.95}
\caption{Operation Opportunity annotation taxonomy for non-instruction data.}
\label{tab:operation_opportunity_description}
\begin{tabularx}{\textwidth}{p{0.52\textwidth} p{0.40\textwidth}}
\toprule
\textbf{Operation type and description} & \textbf{Sub-type label space} \\
\midrule

\code{knowledge\_and\_concept\_grounding} \newline
The text provides factual statements, entity attributes, definitions, concept explanations, category relations, or example-to-concept mappings.
&
\begin{tabular}[t]{@{}l@{}}
\code{fact\_or\_entity\_relation} \\
\code{concept\_definition\_or\_explanation} \\
\code{taxonomy\_or\_category\_relation} \\
\code{example\_to\_concept\_mapping} \\
\code{other}
\end{tabular} \\
\midrule

\code{evidence\_localization\_and\_integration} \newline
The text exposes local evidence anchors, distributed evidence, cross-reference resolution, structured evidence, or multi-hop relation chains.
&
\begin{tabular}[t]{@{}l@{}}
\code{local\_evidence\_anchor} \\
\code{distributed\_evidence} \\
\code{cross\_reference\_resolution} \\
\code{multi\_hop\_relation\_or\_structured\_evidence} \\
\code{other}
\end{tabular} \\
\midrule

\code{comparison\_contrast\_and\_exclusion} \newline
The text compares, ranks, contrasts, distinguishes, negates, excludes, corrects, refutes, or provides counterexamples.
&
\begin{tabular}[t]{@{}l@{}}
\code{comparison\_or\_ranking} \\
\code{contrastive\_distinction} \\
\code{negation\_or\_exclusion} \\
\code{correction\_refutation\_or\_counterexample} \\
\code{other}
\end{tabular} \\
\midrule

\code{temporal\_causal\_and\_process\_reasoning} \newline
The text describes temporal order, event progression, causal relations, mechanisms, consequence chains, workflows, procedures, or process stages.
&
\begin{tabular}[t]{@{}l@{}}
\code{temporal\_sequence\_or\_event\_progression} \\
\code{causal\_relation\_or\_mechanism} \\
\code{consequence\_chain} \\
\code{workflow\_or\_procedure} \\
\code{other}
\end{tabular} \\
\midrule

\code{quantitative\_and\_symbolic\_reasoning} \newline
The text exposes countable structures, numerical relations, calculations, formulas, equations, symbolic transformations, formal proofs, or logical arguments.
&
\begin{tabular}[t]{@{}l@{}}
\code{counting\_or\_enumeration} \\
\code{arithmetic\_or\_quantity\_relation} \\
\code{equation\_or\_formula\_application} \\
\code{symbolic\_derivation\_or\_formal\_proof} \\
\code{other}
\end{tabular} \\
\midrule

\code{constraint\_and\_state\_tracking} \newline
The text contains multiple conditions, rules, constraints, dependencies, state changes, intermediate states, variables, or eligibility requirements that must be tracked consistently.
&
\begin{tabular}[t]{@{}l@{}}
\code{conditional\_rule\_or\_requirement} \\
\code{multi\_constraint\_interaction} \\
\code{state\_or\_variable\_tracking} \\
\code{dependency\_or\_configuration\_rule} \\
\code{other}
\end{tabular} \\
\midrule

\code{simulation\_and\_script\_reasoning} \newline
The text describes physical, spatial, tool-use, social, behavioral, or commonsense scenarios that expose state, action, role, intention, or event-continuation simulation patterns.
&
\begin{tabular}[t]{@{}l@{}}
\code{physical\_or\_spatial\_simulation} \\
\code{tool\_use\_or\_manipulation} \\
\code{social\_interaction\_script} \\
\code{intention\_emotion\_or\_commonsense\_script} \\
\code{other}
\end{tabular} \\
\midrule

\code{algorithm\_and\_code\_reasoning} \newline
The text contains algorithmic logic, pseudocode, program structure, code snippets, execution traces, debugging information, input-output behavior, variable updates, data structures, or complexity analysis.
&
\begin{tabular}[t]{@{}l@{}}
\code{algorithm\_or\_pseudocode} \\
\code{data\_structure\_or\_complexity} \\
\code{code\_snippet\_or\_io\_behavior} \\
\code{execution\_trace\_or\_debugging} \\
\code{other}
\end{tabular} \\
\midrule

\code{boundary\_exception\_and\_failure\_reasoning} \newline
The text explicitly discusses edge cases, corner cases, invalid inputs, exception handling, failure modes, limitations, rule exceptions, robustness issues, or cases where a general rule does not apply.
&
\begin{tabular}[t]{@{}l@{}}
\code{edge\_or\_corner\_case} \\
\code{invalid\_input\_or\_exception} \\
\code{failure\_mode\_or\_limitation} \\
\code{rule\_exception\_or\_robustness\_issue} \\
\code{other}
\end{tabular} \\
\bottomrule
\end{tabularx}
\end{table*}

\begin{table}[t]
\centering
\footnotesize
\caption{Examples of Operation Opportunity annotations}
\label{tab:operation_opportunity_examples}
\renewcommand{\arraystretch}{1.15}
\begin{tabularx}{\linewidth}{@{}L{0.5\linewidth}Y@{}}
\toprule
\textbf{Text} & \textbf{Labels} \\
\midrule

\celltop{
The Shining Bronze-Cuckoo is similar to Horsfield’s Bronze-Cuckoo, but it is more brightly marked with iridescent green-brown on the wings, back and neck. The barring on the chest of adult birds is complete, without the central break of the Horsfield’s Bronze-Cuckoo. The call is a whistle with an upward inflection – similar to a human whistling a dog. They feed mainly on insects and their larvae, especially hairy caterpillars.
}
&
\annlabels{
\textbf{Operation type}: \\
\hspace*{1em}\code{knowledge\_and\_concept\_grounding} \\
\hspace*{1em}\code{comparison\_contrast\_and\_exclusion} \\
\textbf{Sub-type}: \\
\hspace*{1em}\code{fact\_or\_entity\_relation} \\
\hspace*{1em}\code{contrastive\_distinction} \\
} \\

\midrule

\celltop{
In the aftermath of natural disasters such as earthquakes or tsunamis, the paramount importance of sanitation becomes undeniably clear. Past events, notably the Great East Japan earthquake and the Kumamoto earthquake, illuminated the necessity for uninterrupted access to basic amenities like toilets. These pivotal experiences emphasised the dire need for a sanitation solution that is versatile, sustainable, and reliable, even amidst the unpredictabilities of crises.
}
&
\annlabels{
\textbf{Operation type}: \\
\hspace*{1em}\code{evidence\_localization\_and\_integration} \\
\hspace*{1em}\code{temporal\_causal\_and\_process\_reasoning} \\
\textbf{Sub-type}: \\
\hspace*{1em}\code{multi\_hop\_relation\_or\_structured\_evidence} \\
\hspace*{1em}\code{causal\_relation\_or\_mechanism} \\
} \\

\bottomrule
\end{tabularx}
\end{table}

\subsubsection{Supervision Density}
\label{subsubsec:non_instruction_supervision_density}

The \emph{supervision density} dimension characterizes how effectively a non-instruction sample can serve as training signal under the language modeling objective, jointly considering whether its surface form is clean and recoverable and whether its underlying content is substantive, verifiable, and transferable. This dimension instantiates the fourth and final stage of the content-world~$\rightarrow$~discourse-structure~$\rightarrow$~operation-opportunity~$\rightarrow$~supervision-density progression introduced in Section~\ref{subsubsec:understanding_non_instruction_data}: whereas the preceding three perspectives jointly characterize what a sample is about, how it is organized, and what operations it exposes, supervision density asks a logically prior question—whether the resulting signal is usable as next-token supervision at all. We operationalize this dimension through two complementary subcategories, \emph{cleanliness} and \emph{knowledge density}, corresponding respectively to the recoverability and the informativeness of the underlying signal. The two subcategories form a sequential gate rather than two independent scores: a sample must first pass a cleanliness check before its knowledge density can be meaningfully assessed, since knowledge density presupposes that the underlying content has already been recovered from the surface text. Table~\ref{tab:supervision_density_description} summarizes the annotation fields and diagnostic roles of these two subcategories, and Table~\ref{tab:supervision_density_examples} provides representative annotated examples.

The cleanliness subcategory is designed to determine whether a corpus segment contains a usable learning signal once extraction noise and surface corruption are accounted for. The noise-profile field separates distinct noise sources, including boilerplate navigation, formatting or markup artifacts, and OCR or encoding corruption, among others. This separation is useful because different noise patterns imply different data actions: boilerplate-heavy samples may call for source-level filtering, whereas OCR corruption may call for preprocessing repair. Building on this profile, the overall text-cleanliness label summarizes whether the main content is mostly clean, partially recoverable, or heavily corrupted, while the learnable-signal field separately records whether the sample still carries coherent language, factual knowledge, reasoning structure, code, mathematics, dialogue, narrative, or other domain-specific content despite its surface noise. This separation matters because a sample with noticeable formatting residue can still expose a strong underlying signal; the two labels are therefore allowed to diverge rather than being collapsed into a single composite judgment. Taken together, the cleanliness subcategory provides the corpus's primary signal for whether a sample's supervision is recoverable at all, supporting noise-aware filtering and preprocessing decisions.

The knowledge-density subcategory is designed to assess whether a clean or recoverable sample is also substantively informative. A formally clean text may still provide weak supervision if it consists mainly of generic promotion, shallow commentary, isolated titles, or low-value templates, whereas a technical document with minor formatting residue may still carry dense and useful learning signal. We therefore score knowledge density on an ordinal scale according to the amount of learnable, verifiable, and transferable information in the text, together with its structural organization, explanatory depth, and proportion of valid main content. Taken together, the knowledge-density subcategory provides the corpus's primary signal for a sample's substantive informativeness, complementing cleanliness's focus on surface recoverability and supporting the targeted retention of high-value training segments.

Unlike content world, discourse structure, and operation opportunity, supervision density has no direct counterpart among the four dimensions used to decompose evaluation samples (Section~\ref{subsec:understanding_evaluation}). Evaluation samples are typically curated to be readable and well-formed, so the only partial analogue, the ambiguity-noise field under background condition, captures surface-level interference rather than the broader notions of extraction noise and informativeness that supervision density targets. Supervision density instead plays a corpus-internal role: it determines whether the content-world, discourse-structure, and operation-opportunity signals identified for a given sample are actually learnable under the language modeling objective, rather than describing a condition to be matched against the evaluation side. In this sense, supervision density closes the four-perspective progression by gating, rather than extending, the signals captured by the preceding three perspectives.

\begin{table*}[t]
\centering
\small
\caption{Supervision Density annotation subcategories for non-instruction data.}
\label{tab:supervision_density_description}
\renewcommand{\arraystretch}{1.15}
\begin{tabularx}{\textwidth}{@{}L{0.18\textwidth}L{0.26\textwidth}Y@{}}
\toprule
\textbf{Subcategory} & \textbf{Field} & \textbf{Diagnostic role} \\
\midrule

\celltop{Cleanliness}
&
\celltop{Noise profile}
&
\celltop{Records the strength of major noise types, including boilerplate navigation, formatting or markup artifacts, and OCR or encoding corruption, among others. This field supports targeted filtering, preprocessing repair, and source-level quality auditing.} \\
\cmidrule(lr){2-3}

&
\celltop{Text cleanliness}
&
\celltop{Assigns an overall ordinal label indicating whether the main content is clean, partially recoverable, or heavily corrupted. This field provides a compact signal for corpus filtering and for separating usable samples from samples whose supervision signal is dominated by noise.} \\
\cmidrule(lr){2-3}

&
\celltop{Learnable signal}
&
\celltop{Assesses whether the sample still carries a coherent and useful learning signal after accounting for noise. This field prevents the mechanical discarding of samples that contain minor artifacts but still expose valuable language, knowledge, reasoning structure, code, mathematics, dialogue, or domain-specific content.} \\

\midrule

\celltop{Knowledge Density}
&
\celltop{Knowledge density score}
&
\celltop{Assigns an ordinal score that reflects the amount of learnable, verifiable, and transferable information in the sample. This field provides the primary signal for distinguishing substantive content from text that is clean but shallow or low in value.} \\

\bottomrule
\end{tabularx}
\end{table*}

\begin{table}[t]
\centering
\footnotesize
\caption{Examples of Supervision Density annotations}
\label{tab:supervision_density_examples}
\renewcommand{\arraystretch}{1.15}
\begin{tabularx}{\linewidth}{@{}L{0.6\linewidth}Y@{}}
\toprule
\textbf{Text} & \textbf{Labels} \\
\midrule

\celltop{
In 1902, renowned German opera singer Lilli Lehmann published "How to Sing", an illustrated guide detailing her singing techniques. She emphasized the importance of maintaining a long, resonating breath that vibrates both forward and backward in the mouth. The singer should focus on preparing a powerful, flexible, and mobile vocal apparatus, filling it entirely with a continuous vocal mixture. This mixture, composed of indistinguishable components, ensures a smooth and consistent tone.
}
&
\celltop{
\textbf{Noise profile}: \par
\hspace*{1em}\code{none} \par
\textbf{Text cleanliness}: \code{high} \par
\textbf{Learnable signal}: \code{strong} \par
\textbf{Knowledge density score}: \code{3}
} \\

\midrule

\celltop{
Skip to main content \par
Image \par
An official website of the United States government \par
Here's how you know \par
Here's how you know \par
Image \par
Official websites use .\_gov \par
A .\_gov website belongs to an official government organization in the United States. \par
Image \par
Secure .\_gov websites use HTTPS \par
A lock ( Lock Locked padlock icon ) or https:\/\/\_means you’ve safely connected to the .\_gov website. Share sensitive information only on official, secure websites. \par
Home Close \par
Search Image: Search \par
1-844-USAGOV1 \par
$\bullet$ All topics and services \par
$\bullet$ The U.S. and its government \par
$\bullet$ Complaints \par
$\bullet$ Disability services \par
$\bullet$ Disasters and emergencies
}
&
\celltop{
\textbf{Noise profile}: \par
\hspace*{1em}\code{boilerplate\_navigation} \par
\hspace*{1em}\code{duplication\_repetition} \par
\textbf{Text cleanliness}: \code{low} \par
\textbf{Learnable signal}: \code{weak} \par
\textbf{Knowledge density score}: \code{1}
} \\

\bottomrule
\end{tabularx}
\end{table}

\subsubsection{Usage}
\label{subsubsec:non_instruction_data_taxonomy_usage}

The non-instruction data taxonomy serves as the data-side counterpart of the evaluation sample taxonomy described in Section~\ref{subsubsec:evaluation_sample_taxonomy_usage}. Whereas the evaluation taxonomy decomposes benchmark samples into structured capability demands, the non-instruction taxonomy decomposes raw corpus segments into structured data affordances. In this paper, we use the taxonomy in four complementary ways: to characterize the affordance composition of a corpus, to support quality-aware filtering and repair, to translate a diagnosed weak capability slice into a concrete data action, and to validate whether a data intervention has changed the intended part of the model's behavior. These four uses share the same underlying mechanism, aggregating sample-level data labels into interpretable groups, but differ in whether the goal is corpus diagnosis, data selection, intervention design, or post-intervention attribution.

The first use is to characterize the affordance composition of a corpus. Rather than treating a training corpus as a mixture of sources or domains alone, we aggregate labels along the four data-side dimensions introduced above. This profile makes it possible to ask, for example, whether a corpus is dominated by standalone factual statements, whether it contains enough whole-passage dependency structures, whether comparison or causal-process patterns are underrepresented, or whether some sources provide text that is clean but shallow. Such analysis offers a more operational view of corpus composition than domain statistics alone, since it describes not only what the corpus is about, but also what kinds of learning signal it can plausibly supply.

The second use is to support quality-aware filtering, repair, and retention. The supervision-density dimension is especially important for this purpose: cleanliness labels identify whether a sample is clean, partially recoverable, or dominated by extraction noise, while the noise profile separates distinct failure modes. These distinctions map to different data actions, for example, samples dominated by boilerplate or formatting noise may call for source-level filtering, whereas samples whose surface corruption is superficial but whose underlying content remains recoverable may instead call for preprocessing repair. Knowledge-density labels further distinguish substantive text from samples that are formally clean but provide little learnable, verifiable, or transferable information. This separation prevents quality control from collapsing into a single scalar score: a sample may be surface-noisy but still carry a valuable learnable signal, or surface-clean but too shallow to justify retention. The taxonomy therefore supports filtering and repair decisions that are sensitive to both recoverability and informativeness.

The third use is to translate a diagnosed weak capability slice into a concrete data action. A weak capability slice is specified along the four evaluation-side dimensions introduced in Section~\ref{subsec:understanding_model_capability}; using the approximate correspondence between these dimensions and the four data-side dimensions established in Section~\ref{subsubsec:understanding_non_instruction_data}, we translate this specification into a data affordance profile, namely the content world, discourse structure, operation opportunities, and supervision density that the supporting data should exhibit. The mapping rules detailed in Section~\ref{subsec:evaluation_to_data_mapping_rules} formalize this translation step. Depending on how well the existing corpus already satisfies the target profile, the resulting data action can take the form of retrieval from the existing corpus, reweighting of already available segments, targeted synthesis of underrepresented affordances, or filtering of distracting low-signal data.

The final use is post-intervention validation and attribution. After retrieval, reweighting, synthesis, filtering, or repair has been applied, we verify whether the intervention enriched the intended data affordances and whether the corresponding weak capability slice improved. This creates a symmetric analysis loop with the evaluation sample taxonomy: before training, evaluation labels identify the weak capability slice and data labels define the data affordance profile; after training, evaluation labels decompose the score change by capability slice, while data labels confirm whether the intended corpus region was modified. This makes it possible to distinguish targeted improvements from unrelated benchmark-level fluctuations, to detect regressions on non-target slices, and to compare a structured intervention against simpler baselines such as domain-only supplementation or random data addition. In this sense, the non-instruction data taxonomy is not merely a corpus annotation scheme; it is the data-side instrument that enables the evaluation--data loop to move from diagnosis, to intervention, to controlled validation.
\subsection{Evaluation-to-Data Mapping Rules}
\label{subsec:evaluation_to_data_mapping_rules}

\subsubsection{Motivation and Workflow}
\label{subsubsec:mapping_rules_motivation}

The evaluation sample taxonomy (Section~\ref{subsec:evaluation_sample_taxonomy}) and the non-instruction data taxonomy (Section~\ref{subsec:non_instruction_data_taxonomy}) characterize two complementary sides of the same underlying loop: the former specifies what capability demand a benchmark sample places on the model, while the latter specifies what learnable affordances a corpus segment can supply. Neither, on its own, indicates which data affordance is likely to remedy a given failure. The evaluation-to-data mapping rules are the bridge that connects the two sides.

This bridge is not equally direct on both ends. Because instruction data is itself decomposed, from its evaluation-side perspective (Section~\ref{subsubsec:understanding_instruction_data}), along the same four dimensions used to characterize evaluation samples, a weak capability slice can specify a target instruction-data condition in essentially the same vocabulary as the slice itself; the mapping from evaluation samples to instruction data is therefore comparatively direct. Non-instruction data, in contrast, is characterized along a structurally different taxonomy whose correspondence to the evaluation-side dimensions is only approximate (Section~\ref{subsubsec:understanding_non_instruction_data}); translating a weak capability slice into a data affordance profile on this side therefore requires an explicit, separately justified translation step. The mapping rules developed in this section are accordingly not restricted to the evaluation-sample-to-non-instruction-data direction: they span the bridge in its broader sense, with the comparatively direct instruction-data mapping and the more indirect non-instruction-data mapping forming its two ends.

Integrating these mapping rules into the conceptual framework yields a concrete five-step workflow that operationalizes the data--evaluation closed loop. A weak capability slice is first localized through the evaluation sample taxonomy and specified along background condition, task type, solving operation, and output constraint. The mapping rules then translate this slice into a data affordance profile: on the non-instruction-data side, a target content world, discourse structure, operation opportunities, and supervision density; on the instruction-data side, the corresponding background condition, task type, solving operation, and output constraint. Given this profile, a data action (retrieval, reweighting, targeted synthesis, or filtering) is selected according to how well the existing corpus already satisfies it. The resulting intervention is then applied and evaluated under controlled experimental conditions, and the outcome is analyzed by returning to the evaluation sample taxonomy to determine whether the targeted slice has improved. This workflow,
\[
\text{Weak Capability Slice} \rightarrow \text{Data Affordance Profile} \rightarrow \text{Data Action} \rightarrow \text{Experimental Validation} \rightarrow \text{Result Analysis},
\]
is the operational form taken by the data--evaluation closed loop once the mapping rules below are in place.

Because the instruction-data side of this mapping reduces, in the limit, to restating the slice specification as a synthesis target in the vocabulary it already shares with evaluation samples (Section~\ref{subsubsec:understanding_instruction_data}), it requires comparatively little additional translation machinery beyond that restatement. The remainder of this section therefore concentrates on developing rules for the more indirect, non-instruction-data side of the bridge: Section~\ref{subsubsec:taxonomy_aligned_mapping_rules} introduces a set of explicit correspondences between dimensions that were aligned by design when the two taxonomies were constructed, and Section~\ref{subsubsec:llm_assisted_mapping_rules} introduces a complementary, LLM-assisted mechanism for slices that fall outside those correspondences.

Importantly, the mapping rules that drive this workflow are not intended to establish deterministic causal relations between training data and benchmark performance. A benchmark failure may arise from multiple, potentially compounding, factors, including insufficient knowledge coverage, missing reasoning patterns, weak format control, decoding effects, or scoring artifacts, and a single data affordance profile cannot be assumed to address all of them. We therefore treat mapping rules as structured heuristics for generating data-intervention hypotheses rather than as guarantees of improvement: given a diagnosed evaluation weakness, they indicate what kind of data signal should be inspected, retrieved, strengthened, or constructed.

\subsubsection{Taxonomy-Aligned Mapping Rules}
\label{subsubsec:taxonomy_aligned_mapping_rules}

The mapping rules introduced here connect the evaluation sample taxonomy and the non-instruction data taxonomy through the dimensions that were intentionally aligned during taxonomy design (Section~\ref{subsubsec:non_instruction_data_taxonomy_construction_annotation}). We restrict the core mapping rules to four explicit, dimension-level correspondences that were built into the two taxonomies by design. Table~\ref{tab:taxonomy_aligned_mapping_rules} summarizes these correspondences.

\begin{table}[t]
\centering
\small
\caption{Taxonomy-aligned mapping rules between the evaluation sample taxonomy and the non-instruction data taxonomy.}
\label{tab:taxonomy_aligned_mapping_rules}
\begin{tabular}{p{0.23\linewidth} p{0.25\linewidth} p{0.36\linewidth}}
\toprule
\textbf{Evaluation-side label} & \textbf{Data-side label} & \textbf{Mapping rationale} \\
\midrule
domain &
domain &
A capability weakness concentrated in a specific domain motivates retrieving, upweighting, or synthesizing corpus segments under the matching domain label. \\
\midrule
discourse form &
discourse form &
The dominant information carrier of an evaluation sample, such as dialogue, exposition, a table, or a code specification, can be matched to corpus segments sharing the same discourse form. \\
\midrule
context scope &
dependency scope &
The information span required to solve an evaluation sample is aligned with the dependency span over which a corpus segment carries its main learnable signal. \\
\midrule
solving operation &
operation opportunity &
The operation an evaluation sample requires the model to perform is mapped to corpus segments that expose an analogous operation-opportunity type. \\
\bottomrule
\end{tabular}
\end{table}

The domain correspondence is the most direct of the four, since the fine-grained domain labels used on the data side are, by construction, identical to those used in the background condition dimension of the evaluation taxonomy (Section~\ref{subsubsec:non_instruction_content_world}). If an evaluation weakness is concentrated in mathematics, code, or another explicitly annotated domain, the corresponding data-side domain label provides a first-pass retrieval or reweighting criterion. This mapping is useful but intentionally coarse, since domain alone carries no information about the operation or discourse pattern the data should supply. A weakness on mathematical evaluation samples, for instance, is equally consistent with needing more symbolic derivation, more equation formulation, more numerical computation, or more proof-like discourse, and domain matching alone cannot distinguish among these fine-grained data strategies; doing so requires the discourse-form and operation-to-opportunity correspondences introduced below.

The discourse-form correspondence maps the evaluation-side discourse form to its data-side counterpart, which was likewise designed to share a common label space (Section~\ref{subsubsec:non_instruction_discourse_structure}), and supports form-preserving retrieval and analysis. For example, weaknesses on table-based evaluation samples motivate inspecting list- or table-like corpus segments. The mapping remains approximate rather than exact, since naturally occurring corpus segments are typically noisier and more hybrid in form than curated benchmark samples, but it still provides a useful structural constraint beyond domain matching alone.

The context-scope correspondence maps context scope on the evaluation side to dependency scope on the data side (Section~\ref{subsubsec:non_instruction_discourse_structure}). The key criterion in both cases is not surface length but the scope of information dependency: an evaluation sample labeled \code{standalone} should be solvable primarily from the question itself, whereas a sample labeled \code{whole\_passage} requires integrating information across the entire provided context; the corresponding data-side label records whether a corpus segment's main learnable signal is local, multi-sentence, or passage-level. This correspondence is particularly useful for diagnosing failures that stem from long-range dependency use rather than from missing domain knowledge.

The operation-to-opportunity correspondence is the central mapping for data--evaluation analysis, linking what an evaluation sample requires the model to do with what a corpus segment gives the model the opportunity to learn (Section~\ref{subsubsec:non_instruction_operation_opportunity}). Unlike the domain and discourse-form correspondences, this mapping is typically many-to-one or many-to-many. Table~\ref{tab:operation_to_opportunity_mapping_examples} provides representative examples.

Not every evaluation-side dimension admits a table-level non-instruction-data counterpart, and the four correspondences above are intentionally exhaustive under this constraint rather than a partial list awaiting completion. Output constraint has no non-instruction-data analogue at all, since raw corpus text carries no explicit answer slot or scoring rule for it to align with; task type, together with the language, parametric--contextual need, and ambiguity-noise subcategories of background condition, is likewise omitted here because it describes a property of an explicit task presentation rather than of a raw corpus segment. These remaining evaluation-side dimensions are instead addressed by the comparatively direct instruction-data mapping previewed in Section~\ref{subsubsec:mapping_rules_motivation}, since instruction data's evaluation-side decomposition already specifies background condition, task type, solving operation, and output constraint in the same vocabulary used to define a weak capability slice.

\begin{table}[t]
\centering
\small
\caption{Mapping from evaluation-side solving operations to data-side operation-opportunity labels.}
\label{tab:operation_to_opportunity_mapping_examples}
\begin{tabular}{p{0.42\linewidth} p{0.48\linewidth}}
\toprule
\textbf{Evaluation-side solving operation} & \textbf{Data-side operation opportunity} \\
\midrule
\code{fact\_recall} &
\multirow{2}{=}{\code{knowledge\_and\_concept\_grounding}} \\
\code{concept\_alignment} & \\
\midrule
\code{span\_extraction} &
\multirow{3}{=}{\code{evidence\_localization\_and\_integration}} \\
\code{evidence\_aggregation} & \\
\code{multi\_hop\_composition} & \\
\midrule
\code{compare\_or\_rank} &
\multirow{2}{=}{\code{comparison\_contrast\_and\_exclusion}} \\
\code{option\_elimination} & \\
\midrule
\code{causal\_inference} &
\multirow{2}{=}{\code{temporal\_causal\_and\_process\_reasoning}} \\
\code{temporal\_reasoning} & \\
\midrule
\code{counting} &
\multirow{4}{=}{\code{quantitative\_and\_symbolic\_reasoning}} \\
\code{arithmetic\_computation} & \\
\code{equation\_formulation} & \\
\code{symbolic\_transformation} & \\
\midrule
\code{constraint\_tracking} &
\code{constraint\_and\_state\_tracking} \\
\midrule
\code{physical\_state\_simulation} &
\multirow{2}{=}{\code{simulation\_and\_script\_reasoning}} \\
\code{social\_script\_simulation} & \\
\midrule
\code{algorithm\_design} &
\multirow{2}{=}{\code{algorithm\_and\_code\_reasoning}} \\
\code{code\_execution\_simulation} & \\
\midrule
\code{boundary\_case\_reasoning} &
\code{boundary\_exception\_and\_failure\_reasoning} \\
\bottomrule
\end{tabular}
\end{table}

\subsubsection{LLM-Assisted Mapping Rules}
\label{subsubsec:llm_assisted_mapping_rules}

The taxonomy-aligned mapping rules introduced in Section~\ref{subsubsec:taxonomy_aligned_mapping_rules} are deliberately restricted to the four dimension-level correspondences that were built into the two taxonomies by design. This restriction makes the rules transparent and easy to audit, but it also means they cannot resolve cases that fall outside a clean one-to-one correspondence: a weak capability slice may combine several dimensions in a way no single table row captures, or it may depend on domain-specific judgment about which operation or discourse pattern is actually missing, as in the mathematics example discussed above. Resolving such cases requires semantic reasoning over the specific slice and its failed examples, rather than a fixed lookup. We therefore introduce LLM-assisted mapping rules as a complementary, more flexible mechanism: where the taxonomy-aligned rules provide a transparent default for the dimensions they cover, the LLM-assisted rules extend coverage to the remaining, more context-dependent cases within that same non-instruction-data scope.

Concretely, we use a strong LLM as a rule-instantiation assistant. Given a weak capability slice, the definitions of both taxonomies, a set of representative failed examples, and optionally corpus-level tag statistics, the LLM proposes a candidate data affordance profile for that slice, comprising target data-side labels, retrieval or reweighting criteria, synthesis specifications for the missing affordance, and validation slices against which the resulting intervention should later be checked.

This LLM-assisted step functions as a planning mechanism rather than as evidence: the LLM does not define the mapping system, and its suggestions are not treated as conclusions. Consistent with the broader treatment of mapping rules as heuristics rather than as causal claims (Section~\ref{subsubsec:mapping_rules_motivation}), its role is to resolve, on a case-by-case basis, the ambiguity that the taxonomy-aligned correspondences leave open, rather than to replace them. The resulting suggestions are reviewed manually before use and, like all mapping-rule outputs, remain subject to experimental validation rather than being accepted at face value.

\section{Case Study}
\label{sec:case_study}
\subsection{Case Study 1: Output-Constraint Diagnosis Reveals an EOS Supervision Bug}
\label{subsec:case_study_output_constraint}

\subsubsection{Background and Initial Observation}
\label{subsubsec:output_constraint_background}

This case study illustrates the closed-loop workflow introduced in Section~\ref{sec:analysis_toolkit}: we localize a weak capability slice using the evaluation sample taxonomy, trace its root cause to a specific property of the continued pre-training pipeline, and validate the resulting fix under controlled experimental conditions.

We begin from a warm-start checkpoint and continue pre-training it for 200B tokens to obtain a checkpoint denoted \emph{base}. The continued pre-training corpus combines open-source data and internally synthesized data processed through a unified internal data pipeline, using a mixture of 50\% general-domain data, 30\% mathematics data, and 20\% code data.

Table~\ref{tab:output_constraint_initial_observation_benchmarks} compares benchmark performance before and after this continued pre-training stage. For each benchmark, we report the relative change
\[
\Delta_{\mathrm{rel}}
=
\frac{s_{\mathrm{base}} - s_{\mathrm{warm}}}{s_{\mathrm{warm}}}
\times 100\%,
\]
where \(s_{\mathrm{warm}}\) and \(s_{\mathrm{base}}\) denote the scores of the warm-start and base checkpoints, respectively.

\begin{table}[t]
\centering
\caption{Benchmark performance of the warm-start and base checkpoints.}
\label{tab:output_constraint_initial_observation_benchmarks}
\begin{tabular}{lccc}
\toprule
\textbf{Benchmark} & \textbf{Warm-start} & \textbf{Base} & \(\bm{\Delta_{\mathrm{rel}}}\) \\
\midrule
PIQA & 78.13 & \textbf{79.16} & +1.32\% \\
HellaSwag & 72.76 & \textbf{78.24} & +7.53\% \\
MMLU & 60.43 & \textbf{66.21} & +9.57\% \\
TriviaQA & 39.78 & \textbf{51.53} & +29.54\% \\
RACE & 42.49 & \textbf{44.11} & +3.81\% \\
DROP & 38.56 & \textbf{42.74} & +10.84\% \\
MMLU-Pro & 28.50 & \textbf{38.96} & +36.70\% \\
BBH & \textbf{47.27} & 25.14 & -46.82\% \\
C-Eval & 63.22 & \textbf{68.50} & +8.35\% \\
CMMLU & 65.21 & \textbf{70.90} & +8.73\% \\
AGIEval-CN & 44.59 & \textbf{51.22} & +14.87\% \\
MBPP & 34.00 & \textbf{45.40} & +33.53\% \\
HumanEval & 28.05 & \textbf{32.32} & +15.22\% \\
GSM8K & 53.60 & \textbf{68.54} & +27.87\% \\
MathQA & 43.79 & \textbf{54.44} & +24.32\% \\
MATH (Minerva) & 29.52 & \textbf{43.04} & +45.80\% \\
\bottomrule
\end{tabular}
\end{table}

Continued pre-training improves nearly every individual benchmark, with relative gains ranging from +1.32\% on PIQA to +45.80\% on MATH (Minerva). BBH is the sole exception: its score drops from 47.27 to 25.14, a relative change of \(-46.82\%\). Because this regression runs counter to the otherwise uniform improvement, a natural first hypothesis is that continued pre-training has degraded the model's underlying multi-step reasoning capability, the capability BBH is designed to probe. We take this discrepancy as the entry point for the diagnostic analysis that follows, and examine whether the regression in fact reflects a loss of reasoning capability or, instead, a more localized failure that an aggregate benchmark score cannot distinguish.

\subsubsection{Output-Constraint Regression on BBH}
\label{subsubsec:output_constraint_bbh_regression}

Because an aggregate BBH score conflates many distinct capability demands, we turn to the evaluation sample taxonomy to disentangle them along separately analyzable dimensions. Among these dimensions, output constraint is the one most directly coupled to the scored outcome, since it governs how a generated response is parsed, normalized, and ultimately judged correct or incorrect; we therefore initiate our diagnosis from this dimension. Specifically, we revisit the regression through the \emph{output constraint} dimension of the evaluation sample taxonomy (Section~\ref{subsubsec:evaluation_output_constraint}), which characterizes each instance by its expected answer form, the size of its admissible answer space, the rigidity of its required output format, the strictness of its exactness requirement, and its scoring rule. Table~\ref{tab:output_constraint_bbh_characterization} reports BBH accuracy aggregated over these output-constraint attributes, with \(\Delta_{\mathrm{rel}}\) denoting the relative change from the warm-start checkpoint to the base checkpoint.

\begin{table*}[t]
\centering
\small
\caption{BBH performance grouped by output-constraint attributes.}
\label{tab:output_constraint_bbh_characterization}
\begin{tabular}{lrrrr}
\toprule
\textbf{Dimension} & \textbf{Sample Count} & \textbf{Warm-start} & \textbf{Base} & \(\bm{\Delta_{\mathrm{rel}}}\) \\
\midrule
\multicolumn{5}{l}{Answer form} \\
\midrule
\texttt{mcq\_single} & 4760 & 47.48 & 21.18 & -55.39\% \\
\texttt{boolean} & 751 & 69.64 & 55.79 & -19.89\% \\
\texttt{generated\_text} & 500 & 9.20 & 1.80 & -80.43\% \\
\texttt{exact\_numeric} & 500 & 49.80 & 40.20 & -19.28\% \\
\midrule
\multicolumn{5}{l}{Answer space} \\
\midrule
\texttt{closed\_set} & 5545 & 50.30 & 25.86 & -48.59\% \\
\texttt{semi\_open} & 763 & 34.99 & 24.64 & -29.58\% \\
\texttt{open} & 203 & 10.84 & 7.39 & -31.83\% \\
\midrule
\multicolumn{5}{l}{Format rigidity} \\
\midrule
\texttt{3} & 5667 & 46.92 & 23.20 & -50.55\% \\
\texttt{0} & 503 & 53.08 & 41.15 & -22.48\% \\
\texttt{1} & 284 & 45.77 & 37.32 & -18.46\% \\
\texttt{2} & 57 & 38.60 & 15.79 & -59.09\% \\
\midrule
\multicolumn{5}{l}{Exactness requirement} \\
\midrule
\texttt{3} & 6511 & 47.27 & 25.14 & -46.82\% \\
\midrule
\multicolumn{5}{l}{Scoring rule} \\
\midrule
\texttt{exact\_match} & 6099 & 47.11 & 24.32 & -48.38\% \\
\texttt{numeric\_equivalence} & 412 & 49.76 & 37.38 & -24.88\% \\
\bottomrule
\end{tabular}
\end{table*}

This characterization confirms that, by construction, BBH is a strict-output benchmark: 87.04\% of its instances carry the highest format-rigidity level (\(\texttt{format\_rigidity}=3\)), all instances carry the highest exactness requirement (\(\texttt{exactness\_requirement}=3\)), and 93.67\% are scored by exact matching. Under this profile, BBH accuracy reflects not only whether the model can solve the underlying reasoning task, but also whether its final response conforms exactly to the output form the scorer accepts.

The degradation is broad across this strict-output regime: the base checkpoint underperforms the warm-start checkpoint for every observed value of every output-constraint attribute. The decline is most pronounced for \texttt{generated\_text} answers, where accuracy falls from 9.20\% to 1.80\%, and for the dominant high-rigidity subset, where accuracy falls from 46.92\% to 23.20\%. This breadth suggests that the regression is not confined to a single task type, answer form, or scoring rule.

The closed-set and multiple-choice subsets, however, reveal a more diagnostic anomaly. Accuracy on \texttt{closed\_set} instances drops from 50.30\% to 25.86\%, and accuracy on \texttt{mcq\_single} instances drops from 47.48\% to 21.18\%. Because these instances draw answers from a finite, typically small set of options, a regression of this magnitude is difficult to attribute solely to increased reasoning difficulty under open-ended generation. This discrepancy motivates a prediction-level audit of parser-sensitive failures, to which we turn next.

\subsubsection{Prediction-Level Audit of Parser-Sensitive Failures}
\label{subsubsec:output_constraint_parser_sensitive_audit}

To test this hypothesis directly, we conduct a prediction-level audit restricted to instances with \(\texttt{answer\_form}=\texttt{mcq\_single}\) or \(\texttt{answer\_space}=\texttt{closed\_set}\), yielding a subset of 5,545 examples. For each example, we compare the raw model response, denoted \(\texttt{resps}\), with the scorer-facing filtered response, denoted \(\texttt{filtered\_resps}\), i.e., the text obtained after the evaluation pipeline's response-filtering logic, which is then passed to the downstream answer parser or exact-match scorer.

\begin{figure}[t]
    \centering
    \includegraphics[width=\linewidth]{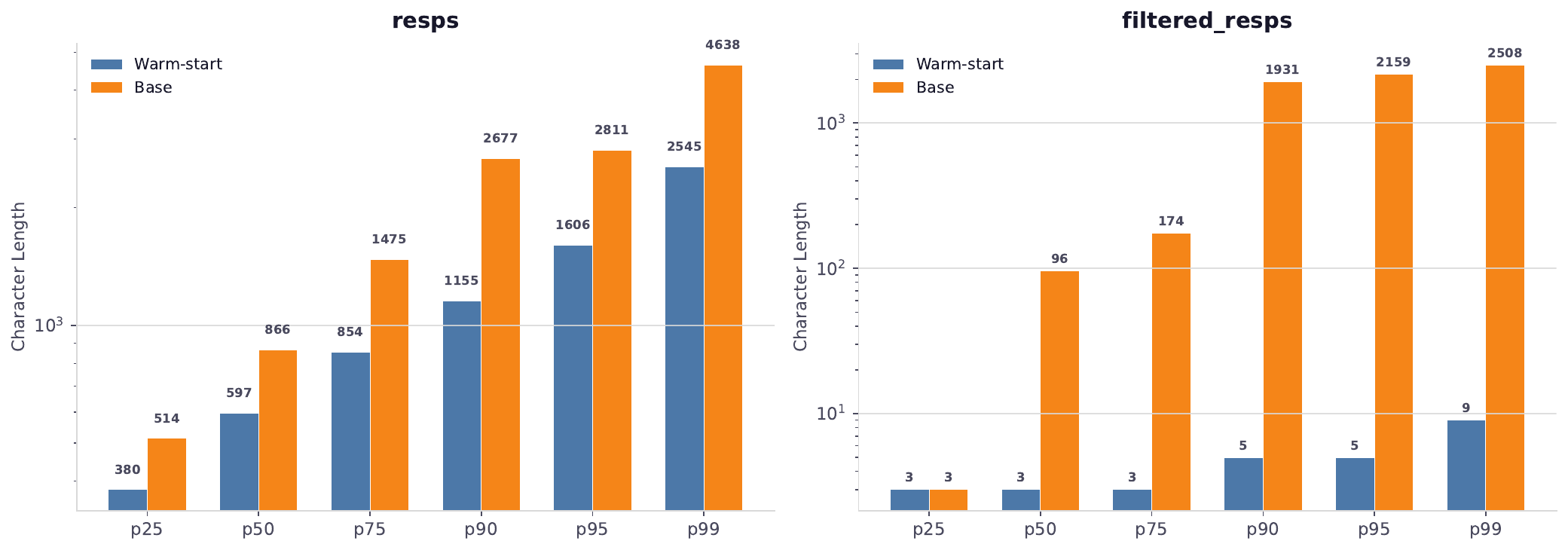}
    \caption{
        Character-length quantiles of \(\texttt{resps}\) and \(\texttt{filtered\_resps}\) on the audited BBH subset, for the warm-start and base checkpoints.
    }
    \label{fig:output_constraint_resp_length_before}
\end{figure}

Figure~\ref{fig:output_constraint_resp_length_before} reports the character-length quantiles of \(\texttt{resps}\) and \(\texttt{filtered\_resps}\) for both checkpoints. The base checkpoint produces substantially longer raw responses than the warm-start checkpoint, indicating that continued pre-training shifts generation behavior toward longer completions even on closed-set and multiple-choice BBH instances.

The more salient anomaly appears in the filtered responses. For the warm-start checkpoint, \(\texttt{filtered\_resps}\) are characteristically short, consistent with the expected behavior on closed-set and multiple-choice tasks, where the filtered answer should reduce to a short option label or canonical answer. For the base checkpoint, however, the filtered-response length is abnormally large: the median length is 96 characters, the 75th percentile is 174 characters, and the 90th percentile reaches 1931 characters. Because the filtering stage is designed to extract a compact answer string rather than an extended continuation, this length distribution is itself evidence of an output-formatting irregularity rather than of a content-level reasoning failure.

We manually inspect examples with unusually long \(\texttt{filtered\_resps}\); Table~\ref{tab:bbh_abnormal_filtered_resp_examples} lists representative cases. The common failure pattern is that the model first emits the correct gold answer and then continues generating additional content beyond it. Because BBH applies strict exactness requirements and exact-match-oriented scoring, such trailing content can render an otherwise correct answer unacceptable to the parser.

\begin{table*}[t]
\centering
\scriptsize
\caption{Representative examples of abnormal \(\texttt{filtered\_resps}\) with excessive suffixes.}
\label{tab:bbh_abnormal_filtered_resp_examples}
\begin{tabularx}{\textwidth}{
  >{\raggedright\arraybackslash}p{0.45\textwidth}
  >{\raggedright\arraybackslash}p{0.35\textwidth}
  >{\raggedright\arraybackslash}p{0.20\textwidth}
}
\toprule
\textbf{\(\texttt{resps}\)} & \textbf{\(\texttt{filtered\_resps}\)} & \textbf{Golden answer} \\
\midrule
If today is 3/5, and it is Jane's second time in the year 1973 to see a meteor shower, then today's date is March 5, 1973. The date one week ago from today is March 5, 1973 minus 7 days, that is March 5, 1973 minus 7 days, that is March 5, 1973 minus 7 days, that is March 5, 1973 minus 7 days
&
\texttt{[invalid]}
&
(E) \\
\midrule
If we look at (A), it says that working the printer is too complex for me, implying that the printer is difficult to work with. However, we know that printers are typically not complex to work with, and the speaker is likely being sarcastic. If we look at (B), it says that working the microprocessor is too complex for me, implying that the microprocessor is difficult to work with. \ldots{} So the answer is (A) and (B).
\par
I have a project that I'm working on that has a lot of different types of files. I'm trying to figure out how to organize them.
&
(A) and (B)
&
(A) \\
\midrule
Here in this question, we are told that ``John was allergic to the drug and he immediately went into shock and experienced cardiac arrest (a heart attack).'' A typical person would therefore think that the misadministration of medication caused John's premature death. So the answer is Yes.<begin\_url>https://www.analystforum.com/t/calculating-expected-return/10116<end\_url>\# Calculating expected return
&
Yes.<begin\_url>https://www.analystforum.com/t/calculating-expected-return/10116<end\_url>\# Calculating expected retur
&
Yes \\
\bottomrule
\end{tabularx}
\end{table*}

To quantify this pattern, we measure two complementary statistics on the audited subset: the rate at which the gold answer occurs as a prefix of \(\texttt{filtered\_resps}\), and, conditional on this match, the amount of trailing content that follows it. Table~\ref{tab:bbh_prefix_suffix_statistics} reports both statistics for the warm-start and base checkpoints.

\begin{table}[t]
\centering
\small
\caption{Gold-answer prefix and suffix diagnostics for \(\texttt{filtered\_resps}\) on the audited BBH subset.}
\label{tab:bbh_prefix_suffix_statistics}
\begin{tabular}{lrr}
\toprule
\textbf{Metric} & \textbf{Warm-start} & \textbf{Base} \\
\midrule
Gold answer appears as prefix & 50.39\% & 69.25\% \\
Accepted when gold answer is prefix & 99.82\% & 37.34\% \\
Cases with extra suffix & 0.18\% & 62.66\% \\
Extra suffix length, mean \(\pm\) std & \(0.01 \pm 0.28\) & \(394.18 \pm 733.48\) \\
\bottomrule
\end{tabular}
\end{table}

As shown in the table, the base checkpoint achieves a higher prefix rate than the warm-start checkpoint (69.25\% versus 50.39\%), i.e., it reaches the correct answer string at least as often. However, among the cases where the filtered response begins with the gold answer, the base checkpoint's prediction is accepted by the original scorer in only 37.34\% of cases, compared with 99.82\% for the warm-start checkpoint, indicating that the base checkpoint frequently produces the correct answer but fails to conform to the format the scorer accepts. This gap coincides with a marked increase in trailing content: only 0.18\% of warm-start prefix cases contain extra characters after the gold answer, versus 62.66\% for the base checkpoint, and the mean suffix length grows correspondingly, from \(0.01 \pm 0.28\) to \(394.18 \pm 733.48\) characters. A substantial fraction of base-checkpoint predictions therefore contain the correct answer as a prefix but append additional text that invalidates the prediction under strict exact matching.

A particularly frequent suffix pattern is content beginning with the special token \(\texttt{\textless begin\_url\textgreater}\), which marks the start of a web document in our pre-training data format and accounts for approximately 60\% of the extra-suffix cases. Its presence suggests that the base checkpoint sometimes continues generating beyond the intended short answer into document-like text, rather than terminating once the required answer has been produced. This constitutes a concrete parser-sensitive failure mode: the model often identifies the correct closed-set answer, but the subsequent continuation violates BBH's strict output contract and is therefore scored as incorrect, plausibly accounting for a large portion of the observed degradation. This failure pattern raises a further question, namely why the base checkpoint fails to terminate generation after producing an otherwise correct, well-formed answer; we examine this question next by inspecting the continued pre-training pipeline itself.

\subsubsection{Missing Termination Supervision at Document Boundaries}
\label{subsubsec:missing_termination_supervision}

The prediction-level audit in Section~\ref{subsubsec:output_constraint_parser_sensitive_audit} identifies a recurring parser-sensitive failure mode on BBH: the base checkpoint frequently emits the correct answer string but then continues generating irrelevant content beyond it. The most common such suffix begins with the special token \(\texttt{\textless begin\_url\textgreater}\), accounting for approximately 60\% of the extra-suffix cases. Because this token marks the start of a new web document in our pre-training data format, its recurrence as an unwanted continuation suggests that the failure may originate in how the model has learned to model document boundaries, rather than in the semantic content of the BBH tasks themselves.

Motivated by this observation, we audit the pre-training pipeline from the perspective of document-boundary representation and termination supervision. In the training corpus, the \(\texttt{\textless EOS\textgreater}\) token accounts for 0.1\% of all tokens, and manual inspection confirms that these tokens are correctly placed at the boundaries between consecutive documents. The serialized data therefore contains explicit and correctly positioned document-boundary markers; we find no evidence that the regression originates from a data-construction error in how boundaries are marked.

The training-code audit, however, reveals a mismatch between the serialized data and the optimization objective applied to it. Although \(\texttt{\textless EOS\textgreater}\) is present as a document separator in the input sequence, its loss is masked during continued pre-training. As a result, the model can attend to \(\texttt{\textless EOS\textgreater}\) tokens in its context, but it never receives direct next-token supervision to predict \(\texttt{\textless EOS\textgreater}\) itself. This creates a termination-supervision gap: the model continues to receive a dense training signal for modeling in-document content, but the objective never explicitly rewards predicting the termination token at the positions where generation should stop.

This gap provides a plausible root cause for the BBH regression observed under strict-output scoring. On closed-set and multiple-choice instances, the base checkpoint often reaches the correct answer, but the absence of explicit termination supervision makes it more likely to continue generating beyond the required answer span. When this continuation happens to contain document-start patterns such as \(\texttt{\textless begin\_url\textgreater}\), the filtered response no longer matches the format the scorer accepts, and an otherwise correct prediction is counted as incorrect. The BBH regression is therefore better characterized as a mismatch between the training objective and the strict termination requirements imposed by the evaluation protocol, rather than as a genuine degradation in the model's underlying task-solving ability — directly addressing the hypothesis raised in Section~\ref{subsubsec:output_constraint_background}.

\subsubsection{Restoring EOS Supervision and Validating Failure-Mode Reduction}
\label{subsubsec:restoring_eos_supervision_validation}

Motivated by this diagnosis, we design a controlled intervention that modifies only the loss mask used during continued pre-training. In the base training recipe, \(\texttt{\textless EOS\textgreater}\) appears in the serialized sequence as a document separator, but its loss is masked; in the intervention, we keep the data, model architecture, optimizer, and all other training hyperparameters unchanged, and restore loss computation and back-propagation on \(\texttt{\textless EOS\textgreater}\) alone. We denote the resulting checkpoint \emph{exp}. This single-variable design isolates the effect of termination supervision from other potential confounds in the training recipe.

Table~\ref{tab:eos_fix_benchmark_results} reports benchmark-level performance for the warm-start, base, and exp checkpoints. The intervention preserves the broad gains obtained from continued pre-training while substantially reversing the BBH regression: BBH rises from 25.14\% (base) to 66.44\% (exp), surpassing not only the base checkpoint but also the original warm-start score of 47.27\%. These results indicate that restoring \(\texttt{\textless EOS\textgreater}\) supervision corrects the strict-output generative regression without materially degrading performance on the other benchmark categories.

\begin{table}[t]
\centering
\small
\caption{Benchmark performance of the warm-start, base, and exp checkpoints.}
\label{tab:eos_fix_benchmark_results}
\begin{tabular}{lccc}
\toprule
\textbf{Benchmark} & \textbf{Warm-start} & \textbf{Base} & \textbf{Exp} \\
\midrule
PIQA & 78.13 & \textbf{79.16} & 79.00 \\
HellaSwag & 72.76 & 78.24 & \textbf{78.42} \\
MMLU & 60.43 & 66.21 & \textbf{67.06} \\
TriviaQA & 39.78 & \textbf{51.53} & 51.43 \\
RACE & 42.49 & \textbf{44.11} & 43.35 \\
DROP & 38.56 & 42.74 & \textbf{46.47} \\
MMLU-Pro & 28.50 & 38.96 & \textbf{39.10} \\
BBH & 47.27 & 25.14 & \textbf{66.44} \\
C-Eval & 63.22 & 68.50 & \textbf{68.57} \\
CMMLU & 65.21 & 70.90 & \textbf{71.29} \\
AGIEval-CN & 44.59 & \textbf{51.22} & 50.38 \\
MBPP & 34.00 & 45.40 & \textbf{47.40} \\
HumanEval & 28.05 & 32.32 & \textbf{42.07} \\
GSM8K & 53.60 & 68.54 & \textbf{71.34} \\
MathQA & 43.79 & 54.44 & \textbf{54.84} \\
MATH (Minerva) & 29.52 & \textbf{43.04} & 42.30 \\
\bottomrule
\end{tabular}
\end{table}

To verify that this improvement is concentrated where the diagnosis predicts, we re-examine BBH performance along the same output-constraint dimensions used in Section~\ref{subsubsec:output_constraint_bbh_regression}. Table~\ref{tab:eos_fix_output_constraint_results} shows that the intervention improves performance across every observed value of every output-constraint attribute; here, \(\Delta_{\mathrm{rel}}\) is computed analogously to the definition in Section~\ref{subsubsec:output_constraint_background}, but relative to the base checkpoint rather than the warm-start checkpoint. The largest gains occur precisely on the subsets most affected by the parser-sensitive failure mode identified earlier: accuracy on \(\texttt{mcq\_single}\) examples rises from 21.18\% to 66.28\%, accuracy on \(\texttt{closed\_set}\) examples rises from 25.86\% to 69.09\%, and accuracy on the dominant high-rigidity subset (\(\texttt{format\_rigidity}=3\)) rises from 23.20\% to 65.29\%. This alignment between where the regression was concentrated and where the recovery is concentrated supports the hypothesis that a substantial part of the original BBH regression stemmed from a failure to terminate after the intended answer, rather than from a uniform loss of task-solving ability.

\begin{table*}[t]
\centering
\small
\caption{BBH performance after restoring \(\texttt{\textless EOS\textgreater}\) supervision, grouped by output-constraint attributes. Scores are reported as percentages.}
\label{tab:eos_fix_output_constraint_results}
\begin{tabular}{lrrrr}
\toprule
\textbf{Dimension} & \textbf{Sample Count} & \textbf{Base} & \textbf{Exp} & \(\bm{\Delta_{\mathrm{rel}}}\) \\
\midrule
\multicolumn{5}{l}{Answer form} \\
\midrule
\texttt{mcq\_single} & 4760 & 21.18 & 66.28 & +212.94\% \\
\texttt{boolean} & 751 & 55.79 & 87.62 & +57.05\% \\
\texttt{generated\_text} & 500 & 1.80 & 19.60 & +988.89\% \\
\texttt{exact\_numeric} & 500 & 40.20 & 83.00 & +106.47\% \\
\midrule
\multicolumn{5}{l}{Answer space} \\
\midrule
\texttt{closed\_set} & 5545 & 25.86 & 69.09 & +167.17\% \\
\texttt{semi\_open} & 763 & 24.64 & 59.50 & +141.48\% \\
\texttt{open} & 203 & 7.39 & 20.20 & +173.34\% \\
\midrule
\multicolumn{5}{l}{Format rigidity} \\
\midrule
\texttt{3} & 5667 & 23.20 & 65.29 & +181.42\% \\
\texttt{0} & 503 & 41.15 & 80.72 & +96.16\% \\
\texttt{1} & 284 & 37.32 & 68.31 & +83.04\% \\
\texttt{2} & 57 & 15.79 & 45.61 & +188.85\% \\
\midrule
\multicolumn{5}{l}{Exactness requirement} \\
\midrule
\texttt{3} & 6511 & 25.14 & 66.44 & +164.28\% \\
\midrule
\multicolumn{5}{l}{Scoring rule} \\
\midrule
\texttt{exact\_match} & 6099 & 24.32 & 65.29 & +168.46\% \\
\texttt{numeric\_equivalence} & 412 & 37.38 & 83.50 & +123.38\% \\
\bottomrule
\end{tabular}
\end{table*}

\begin{figure}[t]
    \centering
    \includegraphics[width=\linewidth]{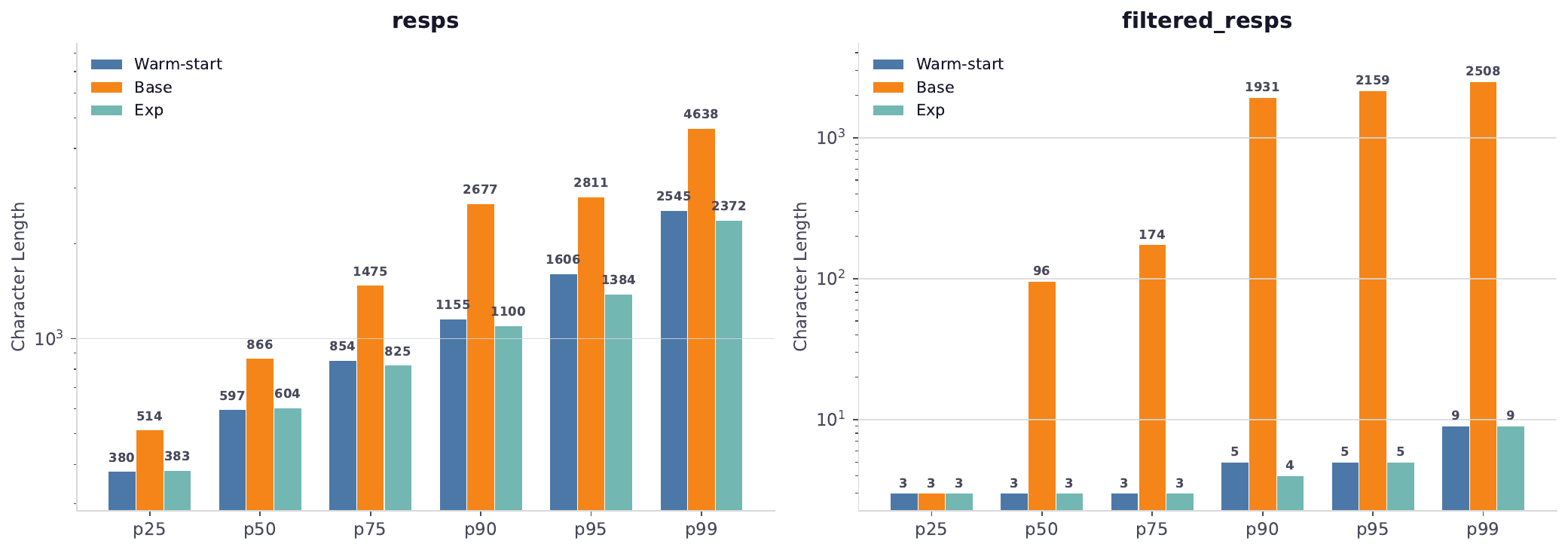}
    \caption{
        Character-length quantiles of \(\texttt{resps}\) and \(\texttt{filtered\_resps}\) on the audited BBH subset, for the warm-start, base, and exp checkpoints.
    }
    \label{fig:output_constraint_resp_length_after}
\end{figure}

We then revisit the prediction-level diagnostics on the same audited subset of 5,545 BBH examples. Figure~\ref{fig:output_constraint_resp_length_after} compares the character-length quantiles of \(\texttt{resps}\) and \(\texttt{filtered\_resps}\) across the warm-start, base, and exp checkpoints. For the raw \(\texttt{resps}\), the exp checkpoint shortens generation length relative to base at every quantile, bringing the distribution back in line with the warm-start checkpoint (e.g., the 99th-percentile length falls from 4638 characters for base to 2372 characters for exp, close to the warm-start value of 2545). The effect is even more pronounced for \(\texttt{filtered\_resps}\): the median length falls from 96 characters for base to 3 characters for exp, matching the warm-start median, and the 90th percentile falls from 1931.40 characters for base to 4 characters for exp, close to the warm-start value of 5. This directly resolves the abnormal scorer-facing long outputs observed in the base checkpoint and indicates that restoring \(\texttt{\textless EOS\textgreater}\) supervision affects generation length broadly, rather than only at the specific point where the filtered answer is extracted.

Table~\ref{tab:eos_fix_prefix_suffix_statistics} provides a more direct confirmation of the proposed mechanism. The rate at which the gold answer appears as a prefix of \(\texttt{filtered\_resps}\) is essentially unchanged between base and exp (69.25\% versus 69.31\%), indicating that the intervention does not change how often the model reaches the correct answer. What changes is whether that correct answer is accepted: among prefix cases, scorer-accepted accuracy rises from 37.34\% for base to 99.69\% for exp, closely matching the warm-start rate of 99.82\%. Correspondingly, the proportion of prefix cases with extra trailing characters falls from 62.66\% to 0.31\%, and the mean suffix length falls from \(394.18 \pm 733.48\) characters to \(0.02 \pm 0.42\) characters, both closely matching the warm-start statistics. The intervention therefore does not act by improving the model's ability to solve BBH tasks; it acts by suppressing the unwanted continuations that previously caused otherwise correct answers to be rejected by the scorer.

\begin{table}[t]
\centering
\small
\caption{Prefix and suffix diagnostics on the audited BBH subset.}
\label{tab:eos_fix_prefix_suffix_statistics}
\begin{tabular}{lrrr}
\toprule
\textbf{Metric} & \textbf{Warm-start} & \textbf{Base} & \textbf{Exp} \\
\midrule
Gold answer appears as prefix & 50.39\% & 69.25\% & 69.31\% \\
Accepted when gold answer is prefix & 99.82\% & 37.34\% & 99.69\% \\
Cases with extra suffix & 0.18\% & 62.66\% & 0.31\% \\
Extra suffix length, mean \(\pm\) std & \(0.01 \pm 0.28\) & \(394.18 \pm 733.48\) & \(0.02 \pm 0.42\) \\
\bottomrule
\end{tabular}
\end{table}

Taken together, the controlled intervention supports the proposed root cause. Masking the loss on \(\texttt{\textless EOS\textgreater}\) leaves the model without direct supervision to terminate at document or answer boundaries, a gap that manifests as long, irrelevant continuations under strict-output evaluation protocols such as BBH. Restoring \(\texttt{\textless EOS\textgreater}\) supervision substantially reduces these continuations, recovers BBH performance beyond both the base and warm-start checkpoints, and improves several other generative benchmarks, while leaving the rest of the training pipeline unchanged. This case study illustrates the broader value of the analysis toolkit: an aggregate benchmark score that initially appeared to indicate a regression in reasoning capability was, on closer inspection through the output-constraint dimension, traced to a narrow and fully correctable issue in training-objective construction, rather than to a deficiency in the underlying data mixture or model capacity.
\subsection{Case Study 2: Solving-Operation Diagnosis Reveals an Operation-Composition Bottleneck in Mathematical Reasoning}
\label{subsec:case_study_solving_operation}

\subsubsection{Background}
\label{subsubsec:solving_operation_background}

We start from a warm-start checkpoint and evaluate it on two complementary benchmark suites. The first suite consists of 16 widely used benchmarks during pre-training, covering general language understanding, knowledge, mathematical reasoning, and code generation. As shown in Table~\ref{tab:solving_operation_warm_start_benchmark}, the checkpoint attains non-trivial performance on several general-purpose benchmarks, such as PIQA, HellaSwag, MMLU, C-Eval, and CMMLU. The same aggregate profile, however, also reveals a comparative weakness on reasoning-intensive mathematical tasks: the checkpoint scores 69.67 on GSM8K, 52.16 on MathQA, and 41.92 on MATH (Minerva).

This benchmark-level observation is informative but not directly actionable. A low score on a mathematical benchmark does not, by itself, indicate whether the underlying failure originates from missing mathematical knowledge, unstable arithmetic execution, weak equation formulation, or insufficient symbolic transformation. The benchmark name alone is therefore too coarse a unit to determine what type of training data should be added, reweighted, or otherwise targeted.

To obtain a more sensitive view of this weakness, we further evaluate the checkpoint under zero-shot Pass@K protocols on several hard mathematical problem-solving and code-generation benchmarks. Table~\ref{tab:solving_operation_warm_start_pass_at_k} shows that the model attains low Pass@1 accuracy on these hard mathematical benchmarks: 7.47 on MATH500, 0.05 on AIME2025, and 0.00 on AIME2026. Although Pass@128 improves with larger sampling budgets on MATH500, the AIME results remain close to zero even at Pass@128, where the model is allowed 128 independent attempts per problem. This indicates that the weakness cannot be attributed to an unfavorable decoding configuration or an insufficient sampling budget: increasing the number of attempts does not surface a correct trajectory, because the model rarely assigns non-negligible probability to a valid reasoning path for these problems in the first place. The gap therefore reflects a limitation in the model's underlying mathematical problem-solving capability rather than one that test-time scaling could resolve.

These results motivate an operation-level diagnosis. Rather than treating mathematics as a monolithic domain, we apply the solving-operation dimension of the evaluation sample taxonomy (Section~\ref{subsubsec:evaluation_solving_operation}) to decompose this aggregate failure into more specific capability slices. This decomposition provides the bridge from benchmark-level weakness to data intervention: once the weak operations are identified, they can be translated, following the mapping rules introduced in Section~\ref{subsec:evaluation_to_data_mapping_rules}, into corresponding data-side affordances and data action.

\begin{table}[t]
\centering
\caption{Performance of the warm-start checkpoint on widely used benchmarks during pre-training.}
\label{tab:solving_operation_warm_start_benchmark}
\begin{tabular}{lc}
\toprule
\textbf{Benchmark} & \textbf{Warm-Start} \\
\midrule
PIQA & 78.24 \\
HellaSwag & 77.22 \\
DROP & 46.15 \\
MMLU-Pro & 37.61 \\
BBH & 60.51 \\
MMLU & 65.50 \\
TriviaQA & 48.28 \\
RACE & 44.69 \\
C-Eval & 67.61 \\
CMMLU & 69.44 \\
AGIEval-CN & 49.52 \\
GSM8K & 69.67 \\
MathQA & 52.16 \\
MATH (Minerva) & 41.92 \\
MBPP & 48.00 \\
HumanEval & 36.59 \\
\bottomrule
\end{tabular}
\end{table}

\begin{table}[t]
\centering
\caption{Zero-shot Pass@K performance of the warm-start checkpoint on hard mathematical and code-generation benchmarks.}
\label{tab:solving_operation_warm_start_pass_at_k}
\begin{tabular}{lccccc}
\toprule
\textbf{Benchmark} & \textbf{Pass@1} & \textbf{Pass@4} & \textbf{Pass@16} & \textbf{Pass@32} & \textbf{Pass@128} \\
\midrule
MATH500 & 7.47 & 22.39 & 46.12 & 57.94 & 76.00 \\
AIME2025 & 0.05 & 0.21 & 0.83 & 1.67 & 6.67 \\
AIME2026 & 0.00 & 0.00 & 0.00 & 0.00 & 0.00 \\
HumanEval & 18.22 & 37.71 & 58.14 & 67.62 & 84.76 \\
\bottomrule
\end{tabular}
\end{table}

\subsubsection{Operation-Level Diagnosis of the Mathematical Weakness}
\label{subsubsec:solving_operation_diagnosis}

Benchmark-level results indicate that the warm-start model is weak at mathematical problem solving, but benchmark names alone are too coarse to identify which capability is weak. We therefore decompose the evaluation set using the solving-operation labels introduced in the evaluation sample taxonomy. Let \(\mathcal{D}\) denote the evaluation set and let \(L(x)\) denote the set of solving-operation labels assigned to sample \(x \in \mathcal{D}\). For an operation subset \(O\), we define the inclusive slice
\[
\mathcal{D}(O) = \{x \in \mathcal{D}: O \subseteq L(x)\}.
\]
A single-operation slice is defined by one operation, an operation-pair slice by a pair of operations, and an operation-triple slice by a triple of operations; under this inclusive definition, a sample with multiple labels contributes to every operation subset it contains. In addition, we report frequent exact operation sets, in which the full label set \(L(x)\) matches a recurring operation combination, rather than merely containing it. For each slice, we report its support and accuracy, and treat the corresponding error rate as a descriptive indicator of weakness.

\paragraph{Operation-set coverage and raw error rates.}

We first aggregate samples from three mathematical evaluation benchmarks—GSM8K, MathQA, and MATH (Minerva)—comprising 9,304 items in total, organizing them by single-operation, operation-pair, operation-triple, and frequent exact operation sets. The single-operation statistics in Table~\ref{tab:warm_start_uni_op_stats} show that the most frequent operations are \texttt{arithmetic\_computation}, \texttt{concept\_alignment}, \texttt{multi\_hop\_composition}, and \texttt{equation\_formulation}; together, these operations cover a large fraction of the mathematical evaluation distribution and therefore yield diagnostic slices with substantial support. Among the high-frequency operations, \texttt{constraint\_tracking}, \texttt{boundary\_case\_reasoning}, \texttt{counting}, and \texttt{symbolic\_transformation} exhibit particularly high error rates, suggesting that the weakness is not confined to numerical computation alone.

\begin{table}[t]
\centering
\caption{Top-10 single-operation slices for the warm-start model. Slices are defined inclusively: a sample is counted if its solving-operation labels include the operation.}
\label{tab:warm_start_uni_op_stats}
\begin{tabular}{lccc}
\toprule
\textbf{Operation} & \textbf{Count} & \textbf{Acc. (\%)} & \textbf{Err. (\%)} \\
\midrule
\texttt{arithmetic\_computation} & 8858 & 49.14 & 50.86 \\
\texttt{concept\_alignment} & 5312 & 40.64 & 59.36 \\
\texttt{multi\_hop\_composition} & 5234 & 40.24 & 59.76 \\
\texttt{equation\_formulation} & 4541 & 41.09 & 58.91 \\
\texttt{symbolic\_transformation} & 2867 & 38.75 & 61.25 \\
\texttt{fact\_recall} & 1961 & 47.02 & 52.98 \\
\texttt{constraint\_tracking} & 1685 & 24.39 & 75.61 \\
\texttt{counting} & 970 & 33.51 & 66.49 \\
\texttt{boundary\_case\_reasoning} & 913 & 25.74 & 74.26 \\
\texttt{compare\_or\_rank} & 392 & 41.07 & 58.93 \\
\bottomrule
\end{tabular}
\end{table}

The operation-pair slices in Table~\ref{tab:warm_start_bi_op_stats} further show that the dominant mathematical samples typically require arithmetic computation to be combined with another operation, such as multi-hop composition, concept alignment, equation formulation, or symbolic transformation. Several high-support pairs have substantially higher error rates than arithmetic alone, especially pairs involving \texttt{multi\_hop\_composition}, \texttt{equation\_formulation}, or \texttt{symbolic\_transformation}. This indicates that the model's mathematical weakness is better characterized as an operation-composition problem than as a failure of isolated arithmetic execution.

\begin{table}[t]
\centering
\caption{Top-10 operation-pair slices for the warm-start model. Slices are defined inclusively: a sample is counted if its solving-operation labels include both operations in the pair.}
\label{tab:warm_start_bi_op_stats}
\begin{tabular}{lccc}
\toprule
\textbf{Operation Pair} & \textbf{Count} & \textbf{Acc. (\%)} & \textbf{Err. (\%)} \\
\midrule
(\texttt{arithmetic\_computation}, \texttt{multi\_hop\_composition}) & 5074 & 40.52 & 59.48 \\
(\texttt{arithmetic\_computation}, \texttt{concept\_alignment}) & 4956 & 40.42 & 59.58 \\
(\texttt{arithmetic\_computation}, \texttt{equation\_formulation}) & 4399 & 40.99 & 59.01 \\
(\texttt{concept\_alignment}, \texttt{multi\_hop\_composition}) & 3155 & 29.54 & 70.46 \\
(\texttt{concept\_alignment}, \texttt{equation\_formulation}) & 2932 & 34.62 & 65.38 \\
(\texttt{arithmetic\_computation}, \texttt{symbolic\_transformation}) & 2669 & 37.92 & 62.08 \\
(\texttt{equation\_formulation}, \texttt{multi\_hop\_composition}) & 2569 & 29.08 & 70.92 \\
(\texttt{concept\_alignment}, \texttt{symbolic\_transformation}) & 2166 & 34.07 & 65.93 \\
(\texttt{equation\_formulation}, \texttt{symbolic\_transformation}) & 2131 & 32.94 & 67.06 \\
(\texttt{arithmetic\_computation}, \texttt{fact\_recall}) & 1793 & 45.79 & 54.21 \\
\bottomrule
\end{tabular}
\end{table}

The operation-triple slices in Table~\ref{tab:warm_start_tri_op_stats} offer a more realistic view of difficult mathematical samples, since many benchmark problems require at least three operations to solve. The lowest-accuracy high-support triples concentrate around combinations of \texttt{concept\_alignment}, \texttt{equation\_formulation}, \texttt{multi\_hop\_composition}, and \texttt{symbolic\_transformation}; these slices correspond to problems in which the model must align concepts, formulate mathematical relations, transform symbolic expressions, and sustain a multi-step derivation simultaneously.

\begin{table}[t]
\centering
\caption{Top-10 operation-triple slices for the warm-start model. Slices are defined inclusively: a sample is counted if its solving-operation labels include all three operations in the triple.}
\label{tab:warm_start_tri_op_stats}
\small
\setlength{\tabcolsep}{4pt}
\begin{tabular}{@{}L{0.43\linewidth}rrr@{}}
\toprule
\textbf{Operation Triple} & \textbf{Count} & \textbf{Acc. (\%)} & \textbf{Err. (\%)} \\
\midrule
\annlabels{
\code{arithmetic\_computation}\\
\code{concept\_alignment}\\
\code{multi\_hop\_composition}
}
& \labellist{3007} & \labellist{29.50} & \labellist{70.50} \\

\midrule

\annlabels{
\code{arithmetic\_computation}\\
\code{concept\_alignment}\\
\code{equation\_formulation}
}
& \labellist{2806} & \labellist{34.25} & \labellist{65.75} \\

\midrule

\annlabels{
\code{arithmetic\_computation}\\
\code{equation\_formulation}\\
\code{multi\_hop\_composition}
}
& \labellist{2496} & \labellist{29.01} & \labellist{70.99} \\

\midrule

\annlabels{
\code{arithmetic\_computation}\\
\code{equation\_formulation}\\
\code{symbolic\_transformation}
}
& \labellist{2006} & \labellist{32.10} & \labellist{67.90} \\

\midrule

\annlabels{
\code{arithmetic\_computation}\\
\code{concept\_alignment}\\
\code{symbolic\_transformation}
}
& \labellist{2003} & \labellist{33.05} & \labellist{66.95} \\

\midrule

\annlabels{
\code{concept\_alignment}\\
\code{equation\_formulation}\\
\code{multi\_hop\_composition}
}
& \labellist{1912} & \labellist{24.27} & \labellist{75.73} \\

\midrule

\annlabels{
\code{concept\_alignment}\\
\code{equation\_formulation}\\
\code{symbolic\_transformation}
}
& \labellist{1710} & \labellist{29.82} & \labellist{70.18} \\

\midrule

\annlabels{
\code{arithmetic\_computation}\\
\code{multi\_hop\_composition}\\
\code{symbolic\_transformation}
}
& \labellist{1616} & \labellist{25.37} & \labellist{74.63} \\

\midrule

\annlabels{
\code{equation\_formulation}\\
\code{multi\_hop\_composition}\\
\code{symbolic\_transformation}
}
& \labellist{1468} & \labellist{23.98} & \labellist{76.02} \\

\midrule

\annlabels{
\code{concept\_alignment}\\
\code{multi\_hop\_composition}\\
\code{symbolic\_transformation}
}
& \labellist{1429} & \labellist{23.44} & \labellist{76.56} \\
\bottomrule
\end{tabular}
\end{table}

Finally, Table~\ref{tab:warm_start_comb_op_stats} reports the most frequent exact operation sets. This view complements the inclusive single-operation, operation-pair, and operation-triple statistics by isolating recurring, fully specified operation configurations rather than operation subsets. The exact-set statistics show that lower-complexity combinations, such as \texttt{arithmetic\_computation} alone or \texttt{arithmetic\_computation} paired with \texttt{concept\_alignment}, retain comparatively high accuracy, whereas exact sets that combine arithmetic computation, concept alignment, equation formulation, multi-hop composition, and symbolic transformation have markedly higher error rates. This contrast supports the hypothesis that the warm-start model's mathematical weakness is driven by compositional derivation rather than by a uniform failure across all math-related samples.

\begin{table}[t]
\centering
\caption{Top-10 frequent exact operation sets for the warm-start model. Unlike the inclusive single-operation, operation-pair, and operation-triple slices, each row corresponds to a full recurring operation-label set.}
\label{tab:warm_start_comb_op_stats}
\small
\setlength{\tabcolsep}{4pt}
\begin{tabular}{@{}L{0.43\linewidth}rrr@{}}
\toprule
\textbf{Operation Combination} & \textbf{Count} & \textbf{Acc. (\%)} & \textbf{Err. (\%)} \\
\midrule
\annlabels{
\code{arithmetic\_computation}\\
\code{multi\_hop\_composition}
}
& \labellist{1045} & \labellist{68.90} & \labellist{31.10} \\

\midrule

\annlabels{
\code{arithmetic\_computation}\\
\code{equation\_formulation}
}
& \labellist{638} & \labellist{60.19} & \labellist{39.81} \\

\midrule

\annlabels{
\code{arithmetic\_computation}\\
\code{concept\_alignment}\\
\code{equation\_formulation}
}
& \labellist{407} & \labellist{58.48} & \labellist{41.52} \\

\midrule

\annlabels{
\code{arithmetic\_computation}
}
& \labellist{388} & \labellist{74.48} & \labellist{25.52} \\

\midrule

\annlabels{
\code{arithmetic\_computation}\\
\code{concept\_alignment}\\
\code{equation\_formulation}\\
\code{multi\_hop\_composition}\\
\code{symbolic\_transformation}
}
& \labellist{368} & \labellist{27.45} & \labellist{72.55} \\

\midrule

\annlabels{
\code{arithmetic\_computation}\\
\code{equation\_formulation}\\
\code{multi\_hop\_composition}
}
& \labellist{342} & \labellist{49.42} & \labellist{50.58} \\

\midrule

\annlabels{
\code{arithmetic\_computation}\\
\code{concept\_alignment}\\
\code{multi\_hop\_composition}
}
& \labellist{332} & \labellist{53.31} & \labellist{46.69} \\

\midrule

\annlabels{
\code{arithmetic\_computation}\\
\code{concept\_alignment}
}
& \labellist{328} & \labellist{70.43} & \labellist{29.57} \\

\midrule

\annlabels{
\code{arithmetic\_computation}\\
\code{concept\_alignment}\\
\code{equation\_formulation}\\
\code{multi\_hop\_composition}
}
& \labellist{277} & \labellist{35.74} & \labellist{64.26} \\

\midrule

\annlabels{
\code{arithmetic\_computation}\\
\code{concept\_alignment}\\
\code{equation\_formulation}\\
\code{symbolic\_transformation}
}
& \labellist{237} & \labellist{55.70} & \labellist{44.30} \\
\bottomrule
\end{tabular}
\end{table}

\paragraph{Pairwise composition effects.}

The raw operation-set statistics identify candidate weak operations and recurring difficult combinations, but they do not quantify how the performance of an anchor operation changes once it is composed with another operation. We therefore analyze the pairwise effect of companion operations over high-frequency anchor operations. For an anchor operation \(u\) and a companion operation \(o_i\), we compare the accuracy of the corresponding operation-pair slice against the accuracy of the single-operation slice:
\[
\Delta_{\mathrm{pair}}(u; o_i)
=
\frac{
\operatorname{Acc}(u,o_i)-\operatorname{Acc}(u)
}{
\operatorname{Acc}(u)
}
\times 100,
\]
where \(\operatorname{Acc}(u)\) denotes the accuracy on \(\mathcal{D}(\{u\})\), and \(\operatorname{Acc}(u,o_i)\) denotes the accuracy on \(\mathcal{D}(\{u,o_i\})\). A negative value of \(\Delta_{\mathrm{pair}}(u;o_i)\) indicates that samples containing both \(u\) and \(o_i\) are harder than the average sample containing the anchor operation \(u\). Conversely, a positive value indicates that the companion operation is associated with higher accuracy relative to the anchor baseline. The corresponding figures are provided in Appendix~\ref{app:pairwise_figures}.

The pairwise-effect results show a consistent pattern across the major mathematical anchors. For \texttt{arithmetic\_computation}, whose single-operation accuracy is \(49.14\%\), the addition of a relatively simple companion such as \texttt{fact\_recall} causes only a moderate drop (\(-6.8\%\)), whereas composition with \texttt{symbolic\_transformation}, \texttt{counting}, \texttt{boundary\_case\_reasoning}, and \texttt{constraint\_tracking} leads to much larger relative decreases of \(-22.8\%\), \(-31.2\%\), \(-50.6\%\), and \(-51.3\%\), respectively. This indicates that arithmetic is not uniformly weak in isolation; rather, it becomes substantially less reliable once embedded in symbolic, counting-intensive, boundary-sensitive, or constraint-heavy contexts.

A similar pattern holds for \texttt{equation\_formulation} and \texttt{multi\_hop\_composition}. For \texttt{equation\_formulation}, whose single-operation accuracy is \(41.09\%\), adding \texttt{arithmetic\_computation} has almost no effect (\(-0.3\%\)), but adding \texttt{multi\_hop\_composition}, \texttt{counting}, \texttt{constraint\_tracking}, and \texttt{boundary\_case\_reasoning} yields substantially larger drops of \(-29.2\%\), \(-37.9\%\), \(-47.2\%\), and \(-49.0\%\). For \texttt{multi\_hop\_composition}, whose single-operation accuracy is \(40.24\%\), the companion \texttt{arithmetic\_computation} is nearly neutral (\(+0.7\%\)), while \texttt{symbolic\_transformation}, \texttt{constraint\_tracking}, and \texttt{boundary\_case\_reasoning} induce large decreases of \(-35.8\%\), \(-46.7\%\), and \(-51.6\%\). These results suggest that multi-step reasoning failures are amplified when the model must simultaneously maintain formal transformations, constraints, or exceptional cases.

For \texttt{symbolic\_transformation}, whose single-operation accuracy is \(38.75\%\), adding \texttt{fact\_recall} is associated with a positive relative change (\(+5.6\%\)), and adding \texttt{arithmetic\_computation} is nearly neutral (\(-2.2\%\)); in contrast, \texttt{multi\_hop\_composition}, \texttt{constraint\_tracking}, and \texttt{boundary\_case\_reasoning} produce large drops of \(-33.4\%\), \(-41.9\%\), and \(-44.0\%\).

Overall, the pairwise-effect analysis refines the raw operation-set diagnosis in two ways. First, it shows that \texttt{arithmetic\_computation} is not the principal source of difficulty in isolation: its weakness is amplified specifically when paired with symbolic transformation, counting, constraint tracking, or boundary-case reasoning. Second, it identifies \texttt{constraint\_tracking} and \texttt{boundary\_case\_reasoning} as systematic difficulty amplifiers across the major mathematical anchors, including arithmetic computation, equation formulation, multi-hop composition, and symbolic transformation. These observations motivate the composition-effect analysis below, which tests whether operation-triple combinations exhibit non-additive degradation beyond what the pairwise effects alone would predict.

\paragraph{Non-additive composition effects.}

The pairwise effect measures how the accuracy of an anchor operation changes when it is composed with a single additional companion operation. It does not, however, determine whether adding two companion operations together produces an approximately additive difficulty or introduces further compositional interference. To make this distinction, we compute a composition effect for each high-frequency anchor operation \(u\) and companion pair \((o_i,o_j)\), using the anchor accuracy \(\operatorname{Acc}(u)\) as the normalization baseline. We first define the relative accuracy changes of the two operation-pair slices and the corresponding operation-triple slice:
\[
r(u,o_i)
=
\frac{\operatorname{Acc}(u,o_i)-\operatorname{Acc}(u)}
{\operatorname{Acc}(u)}
\times 100,
\]
\[
r(u,o_j)
=
\frac{\operatorname{Acc}(u,o_j)-\operatorname{Acc}(u)}
{\operatorname{Acc}(u)}
\times 100,
\]
\[
r(u,o_i,o_j)
=
\frac{\operatorname{Acc}(u,o_i,o_j)-\operatorname{Acc}(u)}
{\operatorname{Acc}(u)}
\times 100.
\]
The composition effect is then defined as
\[
C(u;o_i,o_j)
=
r(u,o_i,o_j)
-
\left(r(u,o_i) + r(u,o_j)\right).
\]
A negative value of \(C(u;o_i,o_j)\) indicates that the operation-triple slice performs worse than expected from the sum of the two pairwise relative changes, suggesting non-additive compositional difficulty. A value close to zero indicates that the triple's difficulty is approximately additive under this metric. A positive value indicates that the full operation-triple slice is less harmful than the sum of the two pairwise effects, which may occur when the two companion operations introduce overlapping rather than independent difficulty. The composition-effect figures are provided in Appendix~\ref{app:non_additive_figures}.

The composition-effect analysis shows that non-additive degradation is not uniform across anchors. For \texttt{arithmetic\_computation}, whose single-operation accuracy is \(49.14\%\), most observed composition effects are modest: the strongest negative effects occur when \texttt{multi\_hop\_composition} is combined with \texttt{fact\_recall}, \texttt{symbolic\_transformation}, or \texttt{equation\_formulation}, with composition effects of \(-9.5\), \(-8.0\), and \(-6.8\), respectively, while other pairs are close to additive and the pair \((\texttt{equation\_formulation}, \texttt{compare\_or\_rank})\) even shows a small positive composition effect of \(+1.8\). Thus, for arithmetic-centered slices, the additional triple-level difficulty is present but relatively bounded; the larger weakness observed in the pairwise analysis is mostly explained by pairwise composition rather than by strong higher-order interference.

The pattern differs for anchors that require comparison, counting, or symbolic manipulation. For \texttt{compare\_or\_rank}, several companion pairs produce large negative composition effects, especially \((\texttt{equation\_formulation}, \texttt{counting})\) at \(-37.1\), \((\texttt{equation\_formulation}, \texttt{fact\_recall})\) at \(-32.8\), and \((\texttt{symbolic\_transformation}, \texttt{counting})\) at \(-31.6\). For \texttt{counting}, the strongest negative composition effects similarly involve comparison-oriented companions: \((\texttt{equation\_formulation}, \texttt{compare\_or\_rank})\) reaches \(-42.7\), and \((\texttt{symbolic\_transformation}, \texttt{compare\_or\_rank})\) reaches \(-38.0\). These results indicate that counting and comparison become substantially more fragile when integrated with equation formulation or symbolic transformation, beyond what the corresponding pairwise difficulties alone would predict.

For \texttt{symbolic\_transformation}, whose single-operation accuracy is \(38.75\%\), most composition effects are negative. The largest degradation appears for \((\texttt{fact\_recall}, \texttt{compare\_or\_rank})\), with a composition effect of \(-33.9\), followed by \((\texttt{counting}, \texttt{compare\_or\_rank})\) at \(-15.0\). Arithmetic- and constraint-related pairs are closer to zero, such as \((\texttt{arithmetic\_computation}, \texttt{compare\_or\_rank})\) at \(-0.9\) and \((\texttt{arithmetic\_computation}, \texttt{constraint\_tracking})\) at \(-1.6\). This suggests that symbolic transformation is especially brittle when the sample additionally requires comparison or ranking, whereas arithmetic-centered symbolic combinations behave closer to additive.

By contrast, \texttt{multi\_hop\_composition} and \texttt{equation\_formulation} show weaker higher-order interference in most arithmetic-related combinations. For \texttt{multi\_hop\_composition}, arithmetic-centered composition effects are close to zero, such as \((\texttt{arithmetic\_computation}, \texttt{counting})\) at \(-0.4\) and \((\texttt{arithmetic\_computation}, \texttt{equation\_formulation})\) at \(-0.9\), while some pairs are even positive, such as \((\texttt{fact\_recall}, \texttt{constraint\_tracking})\) at \(+9.4\) and \((\texttt{equation\_formulation}, \texttt{fact\_recall})\) at \(+9.3\). For \texttt{equation\_formulation}, the majority of arithmetic-related composition effects are likewise near zero, while the largest negative effect is concentrated in \((\texttt{fact\_recall}, \texttt{compare\_or\_rank})\) at \(-24.5\). Although \texttt{multi\_hop\_composition} and \texttt{equation\_formulation} are weak under the raw and pairwise statistics, their higher-order failures are therefore not uniformly super-additive; the strongest non-additive effects arise from more specific operation mixtures rather than from arithmetic-centered combinations.

Overall, the composition-effect analysis refines the pairwise-effect findings above: some weaknesses are primarily pairwise, while others involve stronger higher-order interference. The most pronounced non-additive degradation appears in slices involving \texttt{counting}, \texttt{compare\_or\_rank}, and \texttt{symbolic\_transformation}, particularly when combined with \texttt{equation\_formulation} or other comparison-oriented operations, whereas many arithmetic-centered operation triples remain close to additive relative to the anchor baseline. These findings support a targeted data profile that emphasizes not only arithmetic computation and equation formulation, but also operation combinations that require counting, comparison, symbolic manipulation, and multi-condition reasoning to be executed jointly.

\subsubsection{Diagnosis-Aligned Targeted Sampling from Synthetic Instruction Data}
\label{subsubsec:sampling_from_synthetic_instruction_data}

The operation-level diagnosis in Section~\ref{subsubsec:solving_operation_diagnosis} characterizes the warm-start model's mathematical weakness along three complementary levels: raw error rates over inclusive operation slices, pairwise degradation induced by companion operations, and non-additive degradation over operation triples. Rather than treating these three levels as separate descriptive statistics, we use them jointly to construct a targeted sampling policy over a pool of synthetic instruction data, so that the operation combinations most strongly associated with the diagnosed weakness are over-represented in the resulting training subset.

We reuse the notation introduced in Section~\ref{subsubsec:solving_operation_diagnosis}. Let $\mathcal{D}$ denote the diagnostic mathematical evaluation set, with $N=|\mathcal{D}|$ samples; let $L(x_i)$ denote the solving-operation label set assigned to sample $x_i\in\mathcal{D}$; and let $c_i\in\{0,1\}$ denote its correctness indicator. We define a \emph{feature} $F$ as an operation combination with $|F|\in\{1,2,3\}$, corresponding respectively to single-operation, operation-pair, and operation-triple slices, and write its inclusive slice as
\begin{equation}
\mathcal{D}(F)=\{x_i\in\mathcal{D}:F\subseteq L(x_i)\}.
\end{equation}
Its support, accuracy, and error rate are given by
\begin{equation}
n(F)=|\mathcal{D}(F)|,
\qquad
\operatorname{Acc}(F)=\frac{1}{n(F)}\sum_{i:\,F\subseteq L(x_i)}c_i,
\qquad
\operatorname{Err}(F)=1-\operatorname{Acc}(F).
\end{equation}
Following the inclusive convention of Section~\ref{subsubsec:solving_operation_diagnosis}, a sample carrying multiple solving-operation labels contributes to every feature it contains, so the number of samples containing $F$ serves simultaneously as its support and its frequency in $\mathcal{D}$.

Building on these units, the sampling procedure proceeds through four stages. First, each feature $F$ is assigned a diagnosis-aligned weakness score that combines its raw error rate, pairwise effect, and composition effect, weighted by a support-reliability factor. Second, a feature pool is constructed by retaining the highest-scoring features under support and composition filters. Third, synthetic instruction examples are matched to the retained features through their annotated solving-operation labels, and each matched example is assigned a sampling weight derived from its dominant feature's score. Fourth, the matched pool is sampled via bounded-duplication weighted reservoir sampling, with a small portion of the budget reserved as a fallback to preserve coverage of unmatched examples.

\paragraph{Diagnosis-aligned weakness score.}
For each feature $F$, we compute three diagnostic signals that mirror the three levels of the operation-level diagnosis.

The first signal, \emph{raw weakness}, measures the extent to which a slice's error rate exceeds the overall diagnostic error rate:
\begin{equation}
A_{\mathrm{raw}}(F)=\bigl[\operatorname{Err}(F)-\operatorname{Err}(\mathcal{D})\bigr]_{+},
\end{equation}
where $[z]_{+}=\max(z,0)$. This term is positive only when $F$ is over-represented among errors.

The second signal, \emph{pairwise-degradation weakness}, captures the anchor-level effects analyzed in Section~\ref{subsubsec:solving_operation_diagnosis}. For an operation pair $F=\{u,o\}$, the pairwise effect $\Delta_{\mathrm{pair}}$ is anchor-dependent, so we take the larger of the two anchor-directional degradations,
\begin{equation}
A_{\mathrm{pair}}(\{u,o\})=\max\!\Bigl(\bigl[-\Delta_{\mathrm{pair}}(u;o)\bigr]_{+},\ \bigl[-\Delta_{\mathrm{pair}}(o;u)\bigr]_{+}\Bigr),
\end{equation}
so that the term is large whenever \emph{either} operation becomes substantially less reliable once composed with the other. For single-operation and operation-triple features, $A_{\mathrm{pair}}(F)=0$.

The third signal, \emph{non-additive composition weakness}, captures the higher-order interactions identified in Section~\ref{subsubsec:solving_operation_diagnosis}. For an operation triple $F=\{u,o_i,o_j\}$, we use the composition effect $C$ defined therein. Because the same unordered triple admits three possible anchor choices, we take
\begin{equation}
A_{\mathrm{comp}}(F)=\max_{u\in F}\bigl[-C(u;\,F\setminus\{u\})\bigr]_{+}.
\end{equation}
This term is positive only when the triple performs worse than the sum of its two pairwise effects would predict. For single-operation and operation-pair features, $A_{\mathrm{comp}}(F)=0$.

To prevent low-support features from being ranked highly on the basis of noisy estimates, we introduce a support-reliability factor,
\begin{equation}
\omega(F)=\log\!\bigl(1+n(F)\bigr)\sqrt{\frac{n(F)}{n(F)+\kappa}},
\end{equation}
with shrinkage constant $\kappa$; we set $\kappa=50$ throughout, which attenuates features with support on the order of a few tens of samples while leaving high-support features essentially unscaled. The diagnosis-aligned weakness score is then
\begin{equation}
\operatorname{Weak}(F)=\omega(F)\Bigl(A_{\mathrm{raw}}(F)+A_{\mathrm{pair}}(F)+A_{\mathrm{comp}}(F)\Bigr),
\label{eq:weakness_score}
\end{equation}
where the three signals are combined with equal weight, and $\omega(F)$ is relied upon to stabilize the resulting ranking against support-driven noise. By construction, Eq.~\eqref{eq:weakness_score} directly operationalizes the diagnosis: high-error slices contribute through $A_{\mathrm{raw}}$, operation pairs that destabilize their anchor contribute through $A_{\mathrm{pair}}$, and triples whose three operations interact super-additively contribute through $A_{\mathrm{comp}}$. This score also serves, in the matching step below, as the basis for each matched example's sampling weight.

\paragraph{Feature-pool construction.}
The score in Eq.~\eqref{eq:weakness_score} is defined for features of any order, but enumerating all triples is both computationally wasteful and statistically unreliable: many triples carry no weakness beyond their constituent pairs, and many low-support triples yield unstable composition-effect estimates. We therefore construct the feature pool $\mathcal{F}$ in tiers.

We first score all single operations and all operation pairs, retaining the highest-scoring pairs by $\operatorname{Weak}(F)$. Triples are then formed by extending the retained pair structure and are filtered according to two criteria. A \emph{support filter} removes triples with insufficient evidence,
\begin{equation}
n(T)\geq\theta,\qquad\theta=30,
\end{equation}
which guards against unreliable estimates of the composition effect. A \emph{non-additive composition filter} requires the triple to exhibit super-additive degradation relative to its pairwise sub-features,
\begin{equation}
A_{\mathrm{comp}}(T)>0,
\end{equation}
so that triples whose difficulty is fully explained by their constituent pairs are discarded as redundant. Surviving triples are scored using Eq.~\eqref{eq:weakness_score}, and the top-$K$ ($K=30$) are retained. The final feature pool $\mathcal{F}$ comprises all single operations together with the retained pairs and triples, each carrying its diagnosis-aligned weakness score.

\paragraph{Matching synthetic instruction data to diagnostic features.}
Let $\mathcal{T}$ denote the candidate pool of synthetic instruction examples after standard quality, deduplication, and contamination filtering. Each example $s\in\mathcal{T}$ is annotated with a solving-operation label set $L(s)$ under the taxonomy introduced in Section~\ref{subsubsec:evaluation_solving_operation}.

A synthetic example matches a feature $F\in\mathcal{F}$ when $F\subseteq L(s)$, and its hit set is
\begin{equation}
\operatorname{Hits}(s)=\{F\in\mathcal{F}:F\subseteq L(s)\}.
\end{equation}
Examples for which $\operatorname{Hits}(s)\neq\emptyset$ form the \emph{matched} pool $\mathcal{T}_{\mathrm{m}}$, and the remainder form the \emph{unmatched} pool $\mathcal{T}_{\mathrm{u}}$.

For attribution purposes, each matched example is assigned to a single dominant feature $F^{\star}(s)$ via a two-step rule that gives precedence to higher-order features. Let
\[
r^{\star}(s)=\max_{F\in\operatorname{Hits}(s)}|F|
\]
denote the maximum feature order present in the hit set, and let
\[
\operatorname{Hits}^{\star}(s)=\{F\in\operatorname{Hits}(s):|F|=r^{\star}(s)\}
\]
denote the subset of hit features of that maximum order. The dominant feature is then the highest-scoring feature within this subset,
\[
F^{\star}(s)=\arg\max_{F\in\operatorname{Hits}^{\star}(s)}\operatorname{Weak}(F),
\]
with any residual ties broken by an arbitrary fixed ordering over features. This rule ensures that every matched example is attributed to exactly one feature: higher-order operation combinations take precedence as the more informative attribution, and the weakness score resolves ambiguity within a given order.

Beyond reporting, this dominant feature also defines each matched example's sampling weight: we set
\begin{equation}
w_s=\operatorname{Weak}\bigl(F^{\star}(s)\bigr),
\label{eq:matched_weight}
\end{equation}
so that an example inherits the weakness score of the single most diagnostic feature it exhibits, rather than an aggregate over all features in $\operatorname{Hits}(s)$. This weight is used directly by the weighted reservoir sampling procedure described below.

\paragraph{Bounded weighted sampling with coverage fallback.}
Let $N_{\mathrm{target}}$ denote the desired number of sampled examples, and let $n_{\mathrm{m}}=|\mathcal{T}_{\mathrm{m}}|$ and $n_{\mathrm{u}}=|\mathcal{T}_{\mathrm{u}}|$ denote the sizes of the matched and unmatched pools, respectively. We allocate the sampling budget by first drawing as many matched examples as possible, up to a fraction $\rho\in(0,1]$ of the target, and then filling any residual budget from the unmatched pool. Let $d_{\max}=2$ denote the per-example duplication bound. The matched quota is
\begin{equation}
Q_{\mathrm{match}}
=\min\!\Bigl(\bigl\lceil\rho N_{\mathrm{target}}\bigr\rceil,\
n_{\mathrm{m}}d_{\max}\Bigr),
\end{equation}
where the cap $n_{\mathrm{m}}d_{\max}$ ensures that no matched example is selected more than $d_{\max}$ times. The unmatched draw absorbs only the residual budget, capped by the unmatched-pool size:
\begin{equation}
Q_{\mathrm{cov}}=\min\!\bigl(N_{\mathrm{target}}-Q_{\mathrm{match}},\
n_{\mathrm{u}}\bigr),
\end{equation}
where $\rho$ controls the share of the budget reserved for the diagnosis-aligned matched draw. In our experiments, we set $\rho=1.0$, directing the entire budget to the matched pool so as to up-sample the diagnosed weak operation combinations as aggressively as the duplication bound allows; the unmatched draw activates only as a fallback when $n_{\mathrm{m}}d_{\max}<N_{\mathrm{target}}$, in which case $Q_{\mathrm{cov}}=N_{\mathrm{target}}-n_{\mathrm{m}}d_{\max}$ examples are drawn uniformly from $\mathcal{T}_{\mathrm{u}}$ to preserve distributional coverage. Smaller values of $\rho$ would trade matched intensity for guaranteed coverage of operation combinations outside the diagnosed feature pool; we leave this trade-off to future work.

We draw the $Q_{\mathrm{match}}$ matched examples using Efraimidis--Spirakis weighted reservoir sampling without replacement~\cite{efraimidis2006weighted}, using the sampling weight $w_s=\operatorname{Weak}(F^{\star}(s))$ defined in Eq.~\eqref{eq:matched_weight}. Each matched example $s$ contributes up to $d_{\max}$ candidate slots; for each slot, we draw $u\sim\operatorname{Uniform}(0,1)$ and assign the key
\begin{equation}
k_s=-\frac{\log u}{w_s}.
\end{equation}
The $Q_{\mathrm{match}}$ slots with the smallest keys are retained. Since $-\log u\sim\operatorname{Exp}(1)$, the key follows an exponential race with rate $w_s$, so larger weights yield stochastically smaller keys and a correspondingly higher selection probability; selecting over per-slot keys realizes sampling without replacement in which each matched example appears at most $d_{\max}$ times. The $Q_{\mathrm{cov}}$ coverage examples, if any, are drawn uniformly at random from $\mathcal{T}_{\mathrm{u}}$. The final sampled subset is
\begin{equation}
\mathcal{S}=\operatorname{WeightedSample}\!\bigl(\mathcal{T}_{\mathrm{m}};\,w_s,\,Q_{\mathrm{match}},\,d_{\max}\bigr)\ \cup\ \operatorname{UniformSample}\!\bigl(\mathcal{T}_{\mathrm{u}};\,Q_{\mathrm{cov}}\bigr),
\end{equation}
with $|\mathcal{S}|=N_{\mathrm{target}}$ whenever
$n_{\mathrm{m}}d_{\max}+n_{\mathrm{u}}\geq N_{\mathrm{target}}$.

\subsubsection{Controlled Experiment and Benchmark-Level Validation}
\label{subsubsec:solving_operation_controlled_validation}

\paragraph{Experimental setup.}
We validate the targeted-sampling intervention of Section~\ref{subsubsec:sampling_from_synthetic_instruction_data} under a single-variable design. Starting from the same warm-start checkpoint analyzed in Section~\ref{subsubsec:solving_operation_background}, we continue training under a fixed architecture and optimizer, performing continued pre-training (CPT) on a fixed budget of 100B tokens drawn from a mixture of 50\% general-domain, 30\% math, and 20\% code data. Within the 30B-token math portion of this mixture, we sample \(N_{\mathrm{target}} = 10\)B tokens from a pool of 15B tokens of synthetic math instruction data and substitute them for an equal volume of the original math data, holding the overall 100B-token budget and the 50/30/20 composition fixed across runs; only the construction of this 10B-token instruction subset is varied. The \emph{base} checkpoint is trained on a 10B-token subset drawn from the synthetic instruction pool without weakness-targeted reweighting, whereas the \emph{exp} checkpoint is trained on the importance-sampled 10B-token subset produced by the procedure of Section~\ref{subsubsec:sampling_from_synthetic_instruction_data}, using the same target size \(N_{\mathrm{target}}\) and the same duplication bound \(d_{\max}\). Because the two runs differ only in how this subset is sampled from the same 15B-token pool, any systematic difference between base and exp is attributable to the targeting itself rather than to the amount of instruction data or to the training recipe.

We evaluate all three checkpoints along three complementary axes. The first is the standard 16-benchmark suite used in Section~\ref{subsubsec:solving_operation_background} (Table~\ref{tab:solving_operation_after_benchmark}). The second is the zero-shot Pass@K protocol on MATH500, AIME2025, AIME2026, and HumanEval (Table~\ref{tab:solving_operation_after_pass_at_k}), evaluated under a fixed decoding configuration across checkpoints. The third is a slice-level re-evaluation along the solving-operation dimension.

\begin{table}[t]
\centering
\caption{Performance of the warm-start, base, and exp checkpoints on widely used benchmarks during pre-training.}
\label{tab:solving_operation_after_benchmark}
\begin{tabular}{lccc}
\toprule
\textbf{Benchmark} & \textbf{Warm-start} & \textbf{Base} & \textbf{Exp} \\
\midrule
PIQA & 78.24 & \textbf{78.73} & 78.45 \\
HellaSwag & 77.22 & 77.37 & \textbf{77.50} \\
DROP & 46.15 & \textbf{49.62} & 48.90 \\
MMLU-Pro & 37.61 & 37.74 & \textbf{37.92} \\
BBH & 60.51 & 62.46 & \textbf{62.66} \\
MMLU & \textbf{65.50} & 65.48 & 65.35 \\
TriviaQA & 48.28 & \textbf{48.92} & 48.53 \\
RACE & \textbf{44.69} & 43.44 & 43.73 \\
C-Eval & \textbf{67.61} & 67.53 & 67.38 \\
CMMLU & 69.44 & 70.21 & \textbf{70.22} \\
AGIEval-CN & 49.52 & 49.59 & \textbf{50.84} \\
GSM8K & 69.67 & \textbf{69.90} & 68.61 \\
MathQA & 52.16 & 52.53 & \textbf{53.27} \\
MATH (Minerva) & 41.92 & 42.48 & \textbf{42.82} \\
MBPP & \textbf{48.00} & 47.80 & 46.00 \\
HumanEval & 36.59 & 37.80 & \textbf{40.85} \\
\bottomrule
\end{tabular}
\end{table}

\begin{table}[t]
\centering
\caption{Zero-shot Pass@K performance of the warm-start, base, and exp checkpoints on hard mathematical and code-generation benchmarks.}
\label{tab:solving_operation_after_pass_at_k}
\scriptsize
\setlength{\tabcolsep}{2.5pt}
\begin{tabular}{lccccccccccccccc}
\toprule
\multirow{2}{*}{\textbf{Benchmark}}
& \multicolumn{3}{c}{\textbf{Pass@1}}
& \multicolumn{3}{c}{\textbf{Pass@4}}
& \multicolumn{3}{c}{\textbf{Pass@16}}
& \multicolumn{3}{c}{\textbf{Pass@32}}
& \multicolumn{3}{c}{\textbf{Pass@128}} \\
\cmidrule(lr){2-4}
\cmidrule(lr){5-7}
\cmidrule(lr){8-10}
\cmidrule(lr){11-13}
\cmidrule(lr){14-16}
& \textbf{Warm-start} & \textbf{Base} & \textbf{Exp}
& \textbf{Warm-start} & \textbf{Base} & \textbf{Exp}
& \textbf{Warm-start} & \textbf{Base} & \textbf{Exp}
& \textbf{Warm-start} & \textbf{Base} & \textbf{Exp}
& \textbf{Warm-start} & \textbf{Base} & \textbf{Exp} \\
\midrule
MATH500
& 7.47 & 7.33 & \textbf{7.97}
& 22.39 & 21.95 & \textbf{23.89}
& 46.12 & 45.73 & \textbf{48.97}
& 57.94 & 57.87 & \textbf{61.39}
& 76.00 & 77.20 & \textbf{81.00} \\

AIME2025
& 0.05 & 0.05 & \textbf{0.73}
& 0.21 & 0.21 & \textbf{2.73}
& 0.83 & 0.83 & \textbf{8.65}
& 1.67 & 1.67 & \textbf{13.44}
& 6.67 & 6.67 & \textbf{26.67} \\

AIME2026
& 0.00 & 0.08 & \textbf{0.47}
& 0.00 & 0.31 & \textbf{1.81}
& 0.00 & 1.20 & \textbf{6.32}
& 0.00 & 2.30 & \textbf{10.84}
& 0.00 & 6.67 & \textbf{26.67} \\

HumanEval
& \textbf{18.22} & 16.65 & 17.82
& \textbf{37.71} & 35.42 & 37.63
& 58.14 & 55.54 & \textbf{58.94}
& 67.62 & 65.06 & \textbf{68.54}
& \textbf{84.76} & 80.49 & 83.54 \\
\bottomrule
\end{tabular}
\end{table}

\paragraph{Standard-suite results.}
On the standard suite (Table~\ref{tab:solving_operation_after_benchmark}), the targeted intervention produces modest gains concentrated on the mathematical benchmarks while leaving general-domain performance essentially unchanged. Relative to the warm-start checkpoint, exp improves MathQA from \(52.16\) to \(53.27\) and MATH (Minerva) from \(41.92\) to \(42.82\), in both cases exceeding the base checkpoint as well; GSM8K is the exception, decreasing slightly from \(69.67\) to \(68.61\), even though base itself improves on this benchmark (\(69.90\)). Across the eleven general-language, knowledge, and reading-comprehension benchmarks, exp remains within approximately one point of warm-start in most cases, the largest movements being gains on DROP (\(+2.75\)) and BBH (\(+2.15\)) and a decrease on RACE (\(-0.96\)); we observe no broad regression of the kind that would indicate the intervention had displaced general capability. The code benchmarks show a small, mixed effect: HumanEval improves from \(36.59\) to \(40.85\), while MBPP decreases from \(48.00\) to \(46.00\). Because these benchmarks are evaluated with limited sampling, they offer only a coarse view of the hard mathematical slices that motivated the intervention; we therefore turn first to the Pass@K evaluation below, and subsequently to the operation-level slice analysis on which the link between diagnosis and data construction in Section~\ref{subsubsec:sampling_from_synthetic_instruction_data} ultimately rests.

\paragraph{Pass@K results on hard mathematical benchmarks.}
The Pass@K results in Table~\ref{tab:solving_operation_after_pass_at_k} reveal the intervention's effect most clearly. On MATH500, exp improves over base at every sampling budget, and the gain widens with \(K\), from \(+0.64\) at Pass@1 (\(7.33\to 7.97\)) to \(+3.80\) at Pass@128 (\(77.20\to 81.00\)). The effect is far more pronounced on the AIME benchmarks, which stress exactly the compositional operations flagged in Section~\ref{subsubsec:solving_operation_diagnosis}. On AIME2025, base tracks warm-start at every budget (e.g., \(6.67\) at Pass@128), indicating that continued training without targeted sampling does not by itself surface valid trajectories on these problems; exp, in contrast, raises Pass@128 from \(6.67\) to \(26.67\) and lifts every lower budget correspondingly (e.g., Pass@16 from \(0.83\) to \(8.65\)). On AIME2026, where warm-start scores \(0.00\) at every budget, exp reaches \(0.47\) at Pass@1 and \(26.67\) at Pass@128, again well above base (\(6.67\) at Pass@128). Across all three benchmarks, exp exceeds base at every sampling budget, while base itself remains close to warm-start, varying by only a few points in either direction; this pattern indicates that the observed gains track the targeted sampling procedure itself rather than the continued-training recipe alone.

These results directly address the gap identified in Section~\ref{subsubsec:solving_operation_background}, where the warm-start model failed to surface correct trajectories on AIME even at Pass@128 because it rarely assigned non-negligible probability to a valid reasoning path. After the intervention, additional sampling attempts do surface correct trajectories: the widening of the exp--base gap with increasing \(K\) is the expected signature of a model whose output distribution now places non-negligible probability mass on valid reasoning paths that were previously absent or vanishingly rare. The large gap between Pass@1 and Pass@K---for instance, \(0.73\) versus \(26.67\) on AIME2025---further suggests that the model has acquired the relevant compositional operations without yet consolidating them into its highest-probability completion, consistent with an intervention that broadens trajectory coverage rather than one that merely sharpens an already-correct mode. This distinction matters for how the operation-level attribution below should be read: an intervention that broadens trajectory coverage is expected to produce a stronger signal under Pass@K-style evaluation than under Pass@1-style evaluation, and the operation-level contrast reported below realizes exactly this prediction.

\subsubsection{Operation-Level Attribution of the Gains}
\label{subsubsec:solving_operation_attribution}

\paragraph{Operation-level attribution on the standard suite.}

We now turn to the third evaluation axis introduced in Section~\ref{subsubsec:solving_operation_controlled_validation}: a slice-level re-evaluation along the solving-operation dimension that asks whether the benchmark-level gains reported above are concentrated on the operation combinations targeted by the sampling procedure of Section~\ref{subsubsec:sampling_from_synthetic_instruction_data}, or are instead distributed uniformly across the mathematical evaluation set. We restrict this analysis to the three mathematical benchmarks of the standard suite, GSM8K, MathQA, and MATH (Minerva), and reuse the diagnostic evaluation set \(\mathcal{D}\) and the operation-feature definitions introduced in Section~\ref{subsubsec:solving_operation_diagnosis}. For each retained feature \(F\) in the diagnosis-aligned feature pool \(\mathcal{F}\) of Section~\ref{subsubsec:sampling_from_synthetic_instruction_data}, comprising single operations, operation pairs, and operation triples, we report its support \(n(F)\), its diagnosis-aligned weakness score, and the accuracy change
\[
\Delta\mathrm{acc}(F)=\operatorname{Acc}_{\mathrm{exp}}(F)-\operatorname{Acc}_{\mathrm{base}}(F)
\]
on the inclusive slice \(\mathcal{D}(F)\), where \(\operatorname{Acc}_{\mathrm{exp}}(F)\) and \(\operatorname{Acc}_{\mathrm{base}}(F)\) denote the slice accuracy of the exp and base checkpoints introduced in Section~\ref{subsubsec:solving_operation_controlled_validation}.

\begin{figure}[t]
\centering
\includegraphics[width=0.95\linewidth]{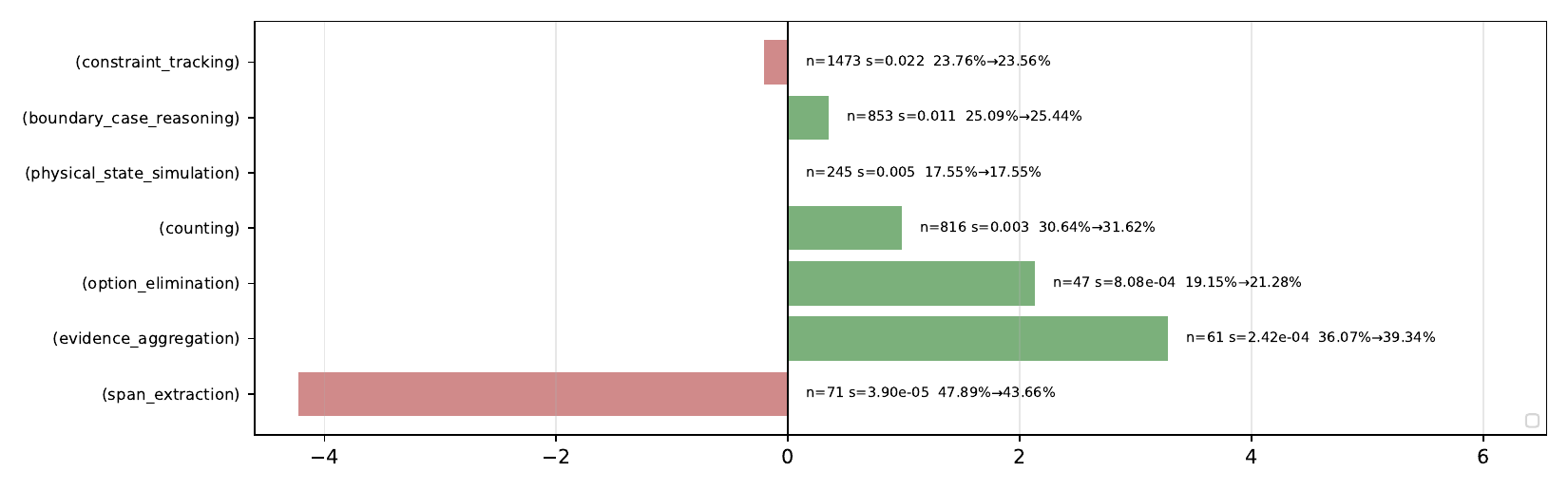}
\caption{Operation-level accuracy change on the standard-suite mathematical evaluation set \(\mathcal{D}\) (GSM8K, MathQA, and MATH (Minerva)), comparing the exp and base checkpoints across the top-7 retained single-operation slices in the diagnosis-aligned feature pool \(\mathcal{F}\) of Section~\ref{subsubsec:solving_operation_diagnosis}. The y-axis lists the single operation defining each inclusive slice \(\mathcal{D}(F)\); the x-axis reports the accuracy change \(\Delta\mathrm{acc}=\mathrm{acc}_{\mathrm{exp}}-\mathrm{acc}_{\mathrm{base}}\), with bars to the right of zero (green) indicating an improvement of exp over base and bars to the left (red) indicating a regression. For each slice, \(n\) denotes its support \(n(F)\) and \(s\) denotes its diagnosis-aligned weakness score, equivalently the sampling weight \(w_s=\operatorname{Weak}(F)\) assigned to matched examples whose dominant feature is \(F\) (Eq.~\eqref{eq:matched_weight}); the accompanying percentages report the base\(\to\)exp accuracy transition, with the arrow indicating the direction of change. Mean \(\Delta\mathrm{acc}=+0.33\)pt.}
\label{fig:op_level_single_std}
\end{figure}

\begin{figure}[t]
\centering
\includegraphics[width=0.95\linewidth]{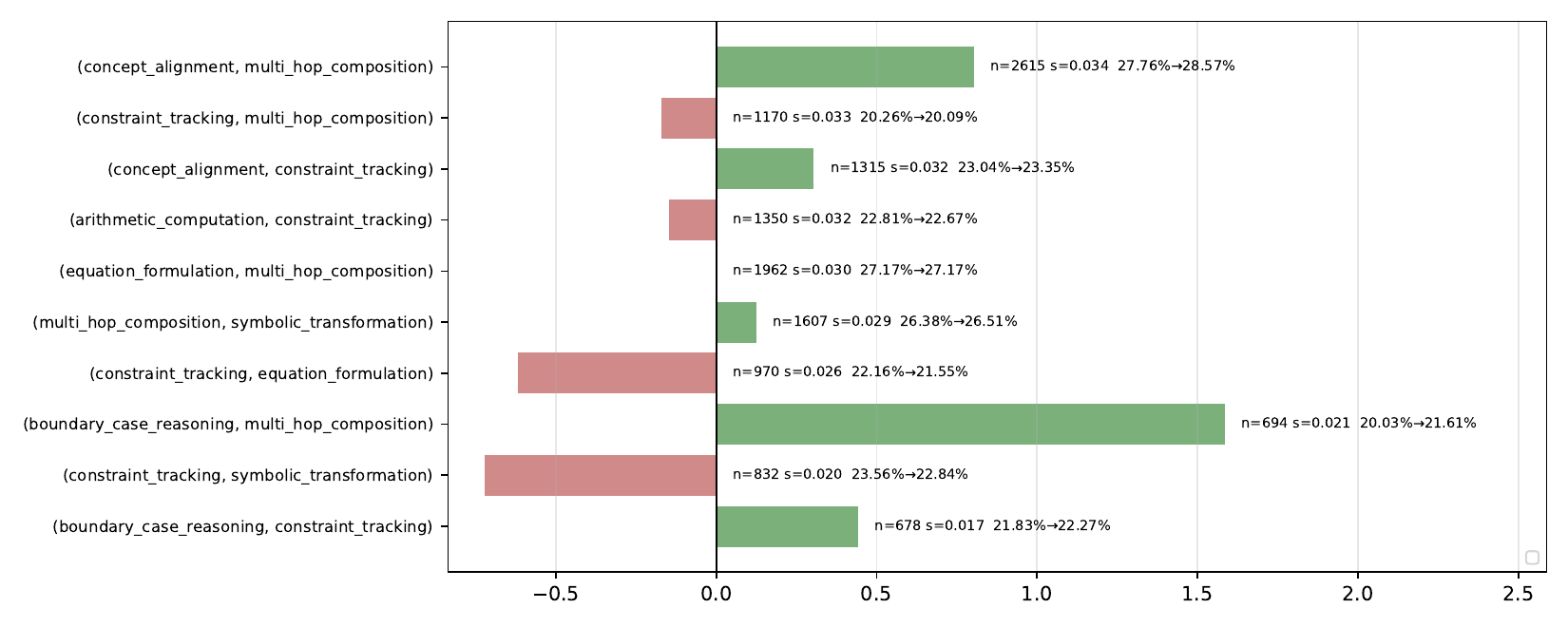}
\caption{Operation-level accuracy change over the top-10 highest-scoring operation pairs in \(\mathcal{F}\); axes, bar coloring, and annotation conventions as in Figure~\ref{fig:op_level_single_std}. Mean \(\Delta\mathrm{acc}=+0.16\)pt.}
\label{fig:op_level_bi_std}
\end{figure}

\begin{figure}[t]
\centering
\includegraphics[width=0.95\linewidth]{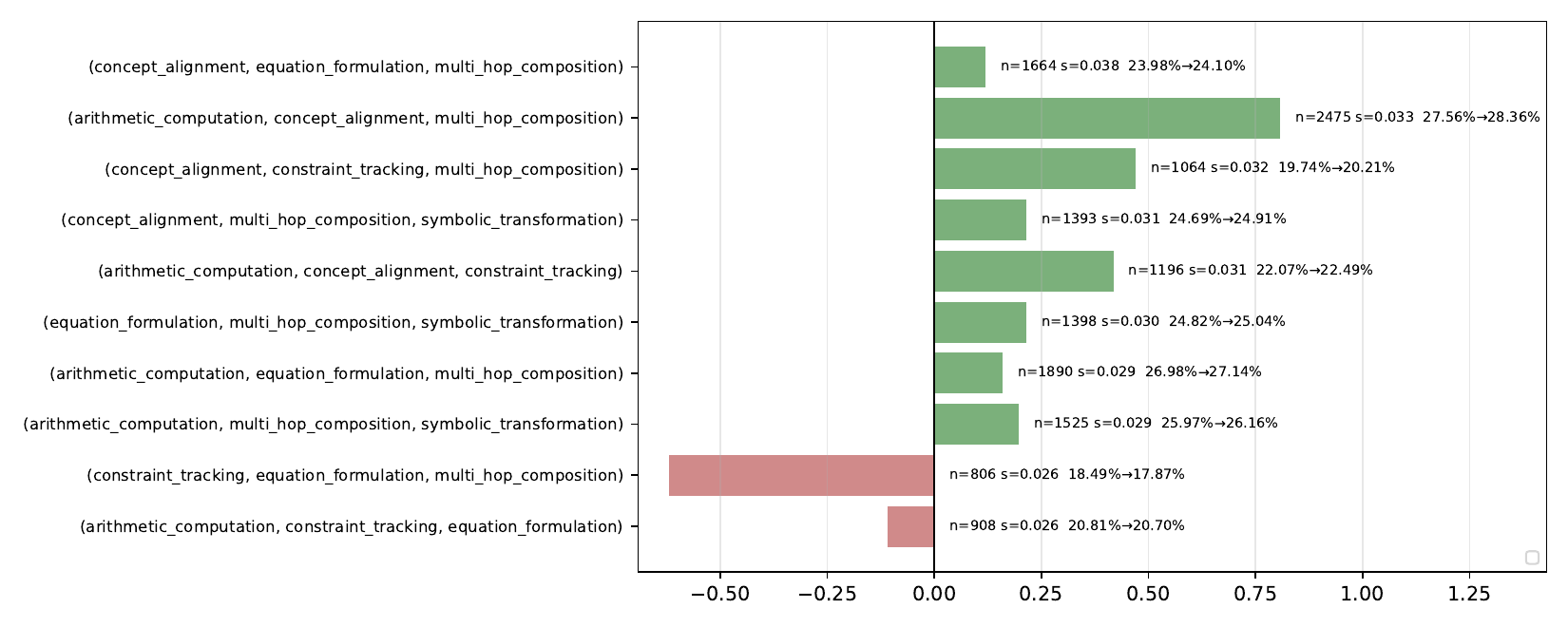}
\caption{Operation-level accuracy change over the top-10 highest-scoring operation triples in \(\mathcal{F}\); conventions as in Figure~\ref{fig:op_level_single_std}. Mean \(\Delta\mathrm{acc}=+0.19\)pt.}
\label{fig:op_level_tri_std}
\end{figure}

The single-operation comparison in Figure~\ref{fig:op_level_single_std} shows a positive but modest mean change of \(+0.33\)pt across the seven retained single-operation slices. Among the operations whose participation in higher-order compositional features made them targets of the sampling policy, \(\texttt{counting}\) gains \(+1.0\)pt on \(816\) samples and \(\texttt{boundary\_case\_reasoning}\) gains \(+0.4\)pt on \(853\) samples; the largest absolute movements occur on low-support slices whose operations fall outside the diagnosed compositional core, namely \(\texttt{evidence\_aggregation}\) (\(+3.3\)pt on \(61\) samples) and \(\texttt{option\_elimination}\) (\(+2.1\)pt on \(47\) samples), which we treat as informative about direction rather than magnitude given their limited support. \(\texttt{constraint\_tracking}\), the slice with both the largest support and the highest weakness score among the single-operation features, regresses by \(-0.2\)pt on \(1{,}473\) samples, and the largest single negative movement is a decrease of \(-4.2\)pt on \(\texttt{span\_extraction}\) (\(n=71\)), an operation that does not appear among the high-scoring pair or triple features retained in \(\mathcal{F}\). This pattern is consistent with the diagnosis of Section~\ref{subsubsec:solving_operation_diagnosis}, which attributed the warm-start weakness to operation composition rather than to uniform single-operation failure: because the sampling policy of Eq.~\eqref{eq:weakness_score} places most of its mass on pair and triple features, we expect the slice-level signal at the single-operation level to be correspondingly small.

The operation-pair comparison in Figure~\ref{fig:op_level_bi_std} gives a more textured picture, with a mean change of \(+0.16\)pt across the ten highest-scoring pairs. The largest gain, \((\texttt{boundary\_case\_reasoning},\texttt{multi\_hop\_composition})\) at \(+1.6\)pt on \(694\) samples, falls on a slice that the diagnosis flagged for both substantial pairwise degradation and low absolute accuracy; the second-largest gain, \((\texttt{concept\_alignment},\texttt{multi\_hop\_composition})\) at \(+0.8\)pt on \(2{,}615\) samples, falls on the highest-support compositional slice in the panel. The slices that fail to improve all involve \(\texttt{constraint\_tracking}\) as one of the two operations: \((\texttt{constraint\_tracking},\texttt{symbolic\_transformation})\) regresses by \(-0.7\)pt on \(832\) samples, \((\texttt{constraint\_tracking},\texttt{equation\_formulation})\) by \(-0.6\)pt on \(970\) samples, \((\texttt{constraint\_tracking},\texttt{multi\_hop\_composition})\) by \(-0.2\)pt on \(1{,}170\) samples, and \((\texttt{arithmetic\_computation},\texttt{constraint\_tracking})\) by \(-0.1\)pt on \(1{,}350\) samples, with \((\texttt{equation\_formulation},\texttt{multi\_hop\_composition})\) essentially flat on \(1{,}962\) samples. Of the six constraint-tracking-bearing pairs in the panel, only \((\texttt{boundary\_case\_reasoning},\texttt{constraint\_tracking})\) (\(+0.4\)pt on \(678\) samples) and \((\texttt{concept\_alignment},\texttt{constraint\_tracking})\) (\(+0.3\)pt on \(1{,}315\) samples) improve at the pair level. We return to this constraint-tracking asymmetry below.

The operation-triple comparison in Figure~\ref{fig:op_level_tri_std} confirms the picture observed at the pair level. The mean change is \(+0.19\)pt across the ten highest-scoring triples, and eight of the ten triples improve over base. The largest gains involve multi-step compositional derivation: \((\texttt{arithmetic\_computation},\texttt{concept\_alignment},\texttt{multi\_hop\_composition})\) gains \(+0.8\)pt on \(2{,}475\) samples, \((\texttt{concept\_alignment},\texttt{constraint\_tracking},\texttt{multi\_hop\_composition})\) gains \(+0.5\)pt on \(1{,}064\) samples, and \((\texttt{arithmetic\_computation},\texttt{concept\_alignment},\texttt{constraint\_tracking})\) gains \(+0.4\)pt on \(1{,}196\) samples. The two triples that regress, \((\texttt{constraint\_tracking},\texttt{equation\_formulation},\texttt{multi\_hop\_composition})\) (\(-0.6\)pt on \(806\) samples) and \((\texttt{arithmetic\_computation},\texttt{constraint\_tracking},\texttt{equation\_formulation})\) (\(-0.1\)pt on \(908\) samples), both pair \(\texttt{constraint\_tracking}\) with \(\texttt{equation\_formulation}\) as their shared core, and both retain low absolute accuracy under both checkpoints. Taken together with the pair-level pattern, the triple-level result indicates that the intervention is most effective on operation combinations whose dominant failure mode is multi-step compositional derivation across \(\texttt{concept\_alignment}\), \(\texttt{equation\_formulation}\), and \(\texttt{symbolic\_transformation}\), and least effective when \(\texttt{constraint\_tracking}\) co-occurs with \(\texttt{equation\_formulation}\).

\paragraph{Operation-level attribution under Pass@128.}

We next re-evaluate the same single-operation, operation-pair, and operation-triple slices from the diagnosis-aligned feature pool \(\mathcal{F}\) of Section~\ref{subsubsec:sampling_from_synthetic_instruction_data}, this time measuring slice accuracy under the zero-shot Pass@128 protocol on MATH500\footnote{We restrict this operation-level Pass@128 analysis to MATH500. AIME2025 and AIME2026 contain too few evaluation samples for the resulting per-slice supports to be statistically meaningful once partitioned by operation, operation pair, and operation triple.} rather than on the standard-suite mathematical evaluation set used above. Because Pass@128 credits a problem as solved once any of \(128\) independent samples succeeds, this view is informative about whether the targeted operation combinations are reachable by the model's output distribution at all, complementing the standard-suite analysis rather than restating it.

\begin{figure}[t]
\centering
\includegraphics[width=0.95\linewidth]{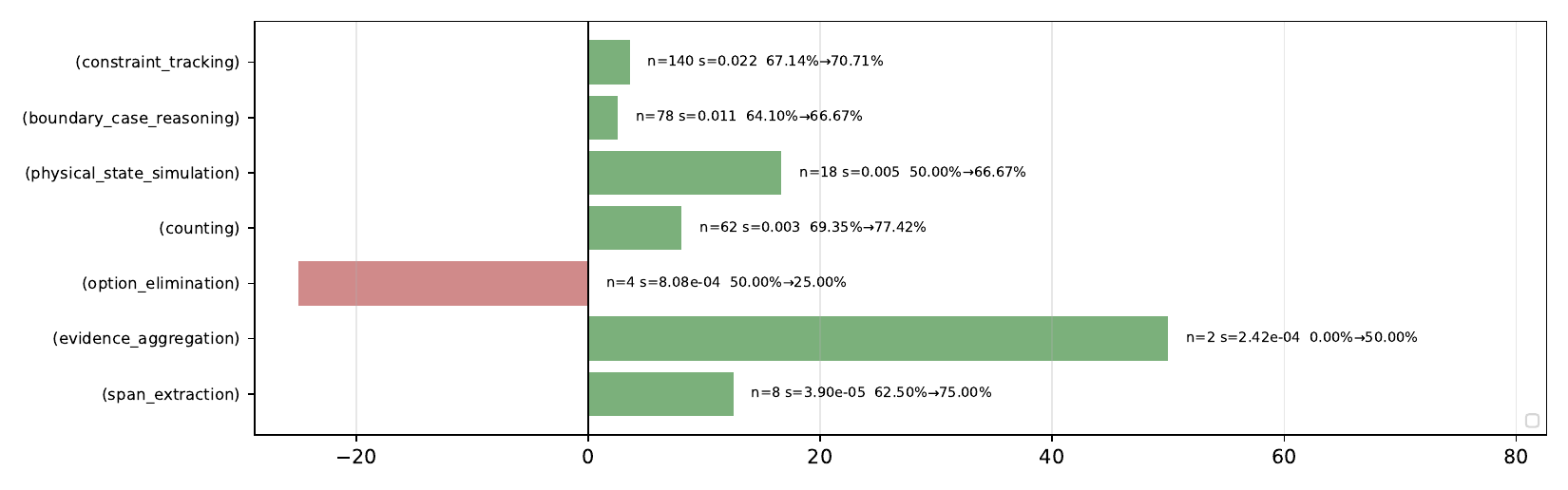}
\caption{Operation-level accuracy change on MATH500 under the zero-shot Pass@128 protocol, over the same top-7 single-operation slices as in Figure~\ref{fig:op_level_single_std}. Axes, bar coloring, and the \(n\)/\(s\)/arrow annotation conventions follow Figure~\ref{fig:op_level_single_std}, with \(\Delta\mathrm{acc}\) and the base\(\to\)exp transition now computed under Pass@128. Mean \(\Delta\mathrm{acc}=+9.77\)pt; this all-slice mean is dominated by very low-support slices, and the mean restricted to the three slices with \(n\ge 50\) is \(+4.74\)pt.}
\label{fig:op_level_single_passk}
\end{figure}

\begin{figure}[t]
\centering
\includegraphics[width=0.95\linewidth]{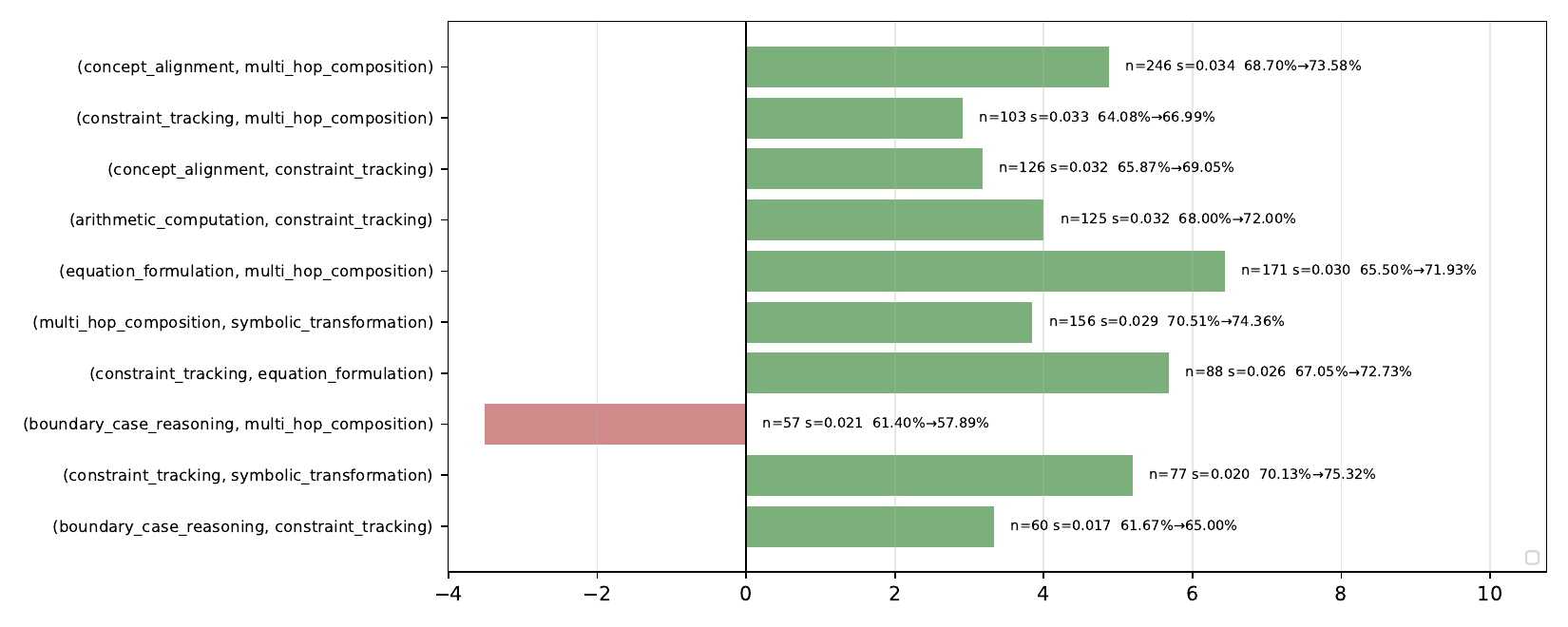}
\caption{Operation-level accuracy change on MATH500 under Pass@128, over the same top-10 operation pairs as in Figure~\ref{fig:op_level_bi_std}; conventions as in Figure~\ref{fig:op_level_single_passk}. Mean \(\Delta\mathrm{acc}=+3.59\)pt; nine of the ten slices improve.}
\label{fig:op_level_bi_passk}
\end{figure}

\begin{figure}[t]
\centering
\includegraphics[width=0.95\linewidth]{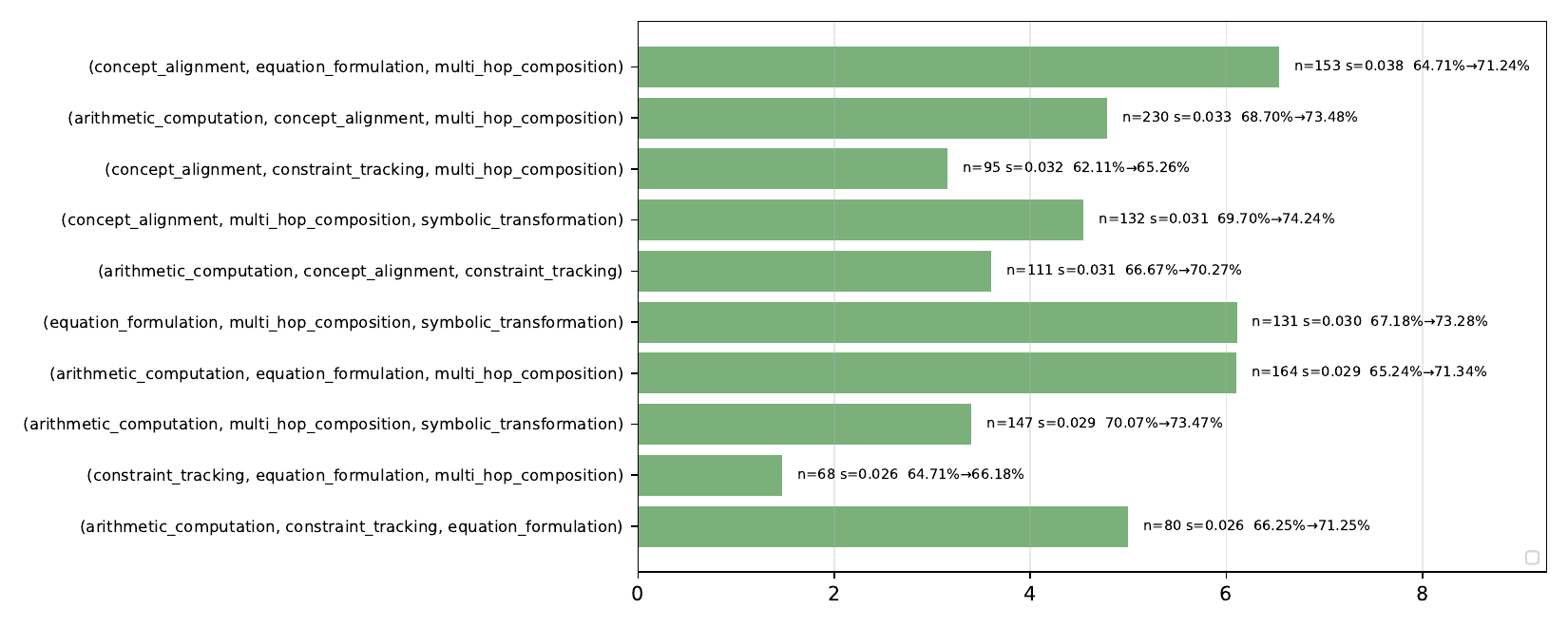}
\caption{Operation-level accuracy change on MATH500 under Pass@128, over the same top-10 operation triples as in Figure~\ref{fig:op_level_tri_std}; conventions as in Figure~\ref{fig:op_level_single_passk}. Mean \(\Delta\mathrm{acc}=+4.47\)pt; all ten slices improve.}
\label{fig:op_level_tri_passk}
\end{figure}

The single-operation comparison in Figure~\ref{fig:op_level_single_passk} shows a mean change of \(+9.77\)pt across the seven retained slices, with six of the seven slices improving. The sole regression is \(\texttt{option\_elimination}\) at \(-25.00\)pt on \(n=4\) samples, where, given the slice size, a single flipped prediction already corresponds to a \(25\)-percentage-point movement. The all-slice mean is heavily influenced by very low-support slices: \(\texttt{evidence\_aggregation}\) (\(n=2\)) contributes \(+50.00\)pt, \(\texttt{physical\_state\_simulation}\) (\(n=18\)) contributes \(+16.67\)pt, and \(\texttt{span\_extraction}\) (\(n=8\)) contributes \(+12.50\)pt; at these sample sizes, a handful of flipped predictions can produce large nominal percentage changes. Restricting attention to the three higher-support single operations (\(n\ge 50\): \(\texttt{counting}\), \(\texttt{boundary\_case\_reasoning}\), and \(\texttt{constraint\_tracking}\)) yields a more stable mean of \(+4.74\)pt, with every one of the three slices improving: \(\texttt{counting}\) gains \(+8.07\)pt on \(62\) samples, \(\texttt{boundary\_case\_reasoning}\) gains \(+2.57\)pt on \(78\) samples, and \(\texttt{constraint\_tracking}\) gains \(+3.57\)pt on \(140\) samples. We treat this higher-support estimate as the load-bearing summary of the single-operation panel; the all-slice mean is reported for completeness but should be read with its low-support sensitivity in mind.

The operation-pair comparison in Figure~\ref{fig:op_level_bi_passk} shows a mean change of \(+3.59\)pt across the ten highest-scoring pairs, with nine of the ten slices improving. All six constraint-tracking-bearing pairs in the panel post gains of several percentage points: \((\texttt{constraint\_tracking},\texttt{equation\_formulation})\) gains \(+5.68\)pt on \(88\) samples, \((\texttt{constraint\_tracking},\texttt{symbolic\_transformation})\) gains \(+5.19\)pt on \(77\) samples, \((\texttt{arithmetic\_computation},\texttt{constraint\_tracking})\) gains \(+4.00\)pt on \(125\) samples, \((\texttt{boundary\_case\_reasoning},\texttt{constraint\_tracking})\) gains \(+3.33\)pt on \(60\) samples, \((\texttt{concept\_alignment},\texttt{constraint\_tracking})\) gains \(+3.18\)pt on \(126\) samples, and \((\texttt{constraint\_tracking},\texttt{multi\_hop\_composition})\) gains \(+2.91\)pt on \(103\) samples. The largest gain in the panel is on \((\texttt{equation\_formulation},\texttt{multi\_hop\_composition})\), at \(+6.43\)pt on \(171\) samples, and the highest-support slice, \((\texttt{concept\_alignment},\texttt{multi\_hop\_composition})\), gains \(+4.88\)pt on \(246\) samples. The only regression in the panel is \((\texttt{boundary\_case\_reasoning},\texttt{multi\_hop\_composition})\) at \(-3.51\)pt on \(57\) samples; given its limited support, we treat this as a within-noise fluctuation rather than as evidence of a systematic effect.

The operation-triple comparison in Figure~\ref{fig:op_level_tri_passk} is the strongest result in the panel set: all ten of the highest-scoring operation triples improve over base, with a mean change of \(+4.47\)pt and individual gains ranging from \(+1.47\)pt to \(+6.53\)pt. The two triples with the smallest gains in the panel, \((\texttt{constraint\_tracking},\texttt{equation\_formulation},\texttt{multi\_hop\_composition})\) at \(+1.47\)pt on \(68\) samples and \((\texttt{arithmetic\_computation},\texttt{constraint\_tracking},\texttt{equation\_formulation})\) at \(+5.00\)pt on \(80\) samples, both pair \(\texttt{constraint\_tracking}\) with \(\texttt{equation\_formulation}\); even so, both remain solidly positive. The largest gain in the panel is on the highest-scoring triple, \((\texttt{concept\_alignment},\texttt{equation\_formulation},\texttt{multi\_hop\_composition})\), at \(+6.53\)pt on \(153\) samples; the next-largest gains involve symbolic-composition triples that the diagnosis of Section~\ref{subsubsec:solving_operation_diagnosis} flagged as compositionally fragile, with \((\texttt{equation\_formulation},\texttt{multi\_hop\_composition},\texttt{symbolic\_transformation})\) and \((\texttt{arithmetic\_computation},\texttt{equation\_formulation},\texttt{multi\_hop\_composition})\) each gaining \(+6.10\)pt. With no regressions in the panel and a smallest individual gain of \(+1.47\)pt, the operation-triple results offer the clearest evidence that the targeted operation combinations are reliably reachable under repeated sampling.

\subsubsection{Cross-Protocol Synthesis and Assessment of the Data--Evaluation Closed Loop}
\label{subsubsec:solving_operation_synthesis}

\paragraph{Standard suite versus Pass@128: consistency and amplification at the operation level.}

The two slice-level evaluations in Sections~\ref{subsubsec:solving_operation_diagnosis} and above target the same diagnosis-aligned feature pool \(\mathcal{F}\) but differ in what they measure, the standard suite credits only the model's single, modal completion, while Pass@128 credits a problem as solved once any of \(128\) independent samples succeeds, and read together they give a sharper picture than either protocol does alone.

First, the operation-level signal is consistent in direction across both protocols on the slices that the diagnosis identified as compositional and underweight in the untargeted baseline: \((\texttt{concept\_alignment},\texttt{multi\_hop\_composition})\), \((\texttt{multi\_hop\_composition},\texttt{symbolic\_transformation})\), \((\texttt{arithmetic\_computation},\texttt{concept\_alignment},\texttt{multi\_hop\_composition})\), and \((\texttt{equation\_formulation},\texttt{multi\_hop\_composition},\texttt{symbolic\_transformation})\) all improve under both the standard suite and Pass@128. This makes it unlikely that the slice-level gains in Figures~\ref{fig:op_level_bi_passk} and \ref{fig:op_level_tri_passk} are an artifact specific to the smaller Pass@128 evaluation set.

Second, the magnitude and uniformity of the operation-level signal differ sharply between protocols at the pair and triple levels. The mean accuracy change is roughly an order of magnitude larger under Pass@128 than under the standard suite (\(+3.59\)pt vs.\ \(+0.16\)pt for pairs; \(+4.47\)pt vs.\ \(+0.19\)pt for triples), and the proportion of improving slices is correspondingly higher (\(9/10\) vs.\ \(5/10\) for pairs; \(10/10\) vs.\ \(8/10\) for triples). The operation-triple result under Pass@128 is the cleanest single piece of evidence that the gains in Table~\ref{tab:solving_operation_after_pass_at_k} are not a generic continued-training effect but track the diagnostic-to-data link that Eq.~\eqref{eq:weakness_score} encodes: all ten top-scoring triples improve, with individual gains ranging from \(+1.47\)pt to \(+6.53\)pt.

Third, the constraint-tracking-bearing slices that limited the standard-suite attribution are recovered, and substantially so, under Pass@128. This is the most informative protocol contrast in the controlled experiment, since it concerns exactly the slices whose standard-suite regression would otherwise be read as a failure of the intervention on the operations it was meant to address. All six constraint-tracking-bearing pairs improve under Pass@128, by between \(+2.91\)pt and \(+5.68\)pt, and the two constraint-tracking-bearing triples that regressed under the standard suite improve by \(+1.47\)pt and \(+5.00\)pt under Pass@128. We interpret this in line with the Pass@1--Pass@K gap discussed in Section~\ref{subsubsec:solving_operation_controlled_validation}: the intervention places non-negligible probability mass on previously absent or vanishingly rare reasoning trajectories for constraint-tracking-heavy problems, but those trajectories have not yet become the model's modal completion. A standard-suite, Pass@1-style evaluation, which credits only the modal completion, is therefore systematically less sensitive to this kind of trajectory broadening than a Pass@K-style evaluation, and the protocol-dependence of the constraint-tracking signal is the operation-level signature of this gap. We accordingly do not read constraint tracking as resistant to the intervention; rather, the standard-suite regression reflects a Pass@1-style measurement of an intervention whose effect on these slices is, at this stage, dominated by trajectory broadening rather than modal sharpening.

\paragraph{Benchmark-level versus operation-level results.}

The operation-level decomposition also helps reconcile two features of the benchmark-level evidence in Section~\ref{subsubsec:solving_operation_controlled_validation} that are difficult to interpret from aggregate scores alone.

The first concerns the relative scale of the standard-suite and Pass@K benchmark deltas. On the standard suite, exp improves over warm-start by a comparatively small margin on MathQA (\(52.16\to53.27\)) and MATH (Minerva) (\(41.92\to42.82\)); under Pass@K, the gap between exp and base instead widens sharply with the sampling budget, reaching \(+3.80\)pt on MATH500 and \(+20.00\)pt on AIME2025 at Pass@128 (Table~\ref{tab:solving_operation_after_pass_at_k}). This asymmetry has the same shape as what we observe at the operation level, where the standard-suite pair and triple means are an order of magnitude smaller than their Pass@128 counterparts. The benchmark-level and operation-level views are therefore mutually consistent on this point, and the operation-level decomposition supplies a candidate mechanism: the targeted operation combinations appear to become reachable under repeated sampling well before they become the modal completion, so a metric that credits only the modal output registers a markedly weaker signal than one that credits any of \(128\) attempts. We restrict this reading to the three mathematical benchmarks analyzed at the operation level; the code-generation benchmarks (HumanEval, MBPP), which show a small and mixed standard-suite effect, fall outside the solving-operation diagnosis used here and are not addressed by this comparison. We also note that the operation-level Pass@128 results are themselves drawn only from MATH500, so they bear most directly on the MATH500 benchmark-level gain and only suggestively, rather than directly, on the AIME results, which the per-slice annotation could not support (see footnote above).

The second concerns an apparent tension: GSM8K is the one standard-suite mathematical benchmark on which exp regresses relative to warm-start (\(69.67\to68.61\)), even though base itself improves (\(69.90\)). Read in isolation, this could suggest the intervention is mildly harmful on at least one mathematical benchmark. Two non-exclusive readings are consistent with the operation-level evidence above, and we cannot adjudicate between them with the evidence collected in this case study. One is that the regression is largely within ordinary evaluation noise for a single-checkpoint comparison, given that the standard-suite operation-level results already show several slices moving by less than one point in either direction. The other is that it reflects the same protocol-dependence identified in the constraint-tracking case: if GSM8K's problem mix draws more heavily on constraint-tracking-style compositions than MathQA or MATH (Minerva) do, a standard-suite, Pass@1-style read-out would be expected to under-detect any underlying improvement on this benchmark in particular. We have not separately verified the per-benchmark composition of operation labels, so we present this only as a plausible, operation-level-motivated hypothesis rather than as an established explanation. What the operation-level results do support is the narrower claim that the GSM8K regression is consistent with a benchmark-level instance of the same protocol-sensitivity already documented at the slice level, rather than with a broad, intervention-induced loss of mathematical capability: no comparable regression appears across the eleven general-language, knowledge, and reading-comprehension benchmarks in Table~\ref{tab:solving_operation_after_benchmark}, and the operation-level regressions on the standard suite are themselves concentrated in a small, identifiable subset of slices rather than spread broadly across the feature pool.

\paragraph{Assessment of the targeted-sampling method and the data--evaluation loop.}

Taken together, these comparisons support a specific and appropriately qualified judgment about the diagnosis-aligned weighted-sampling method of Section~\ref{subsubsec:sampling_from_synthetic_instruction_data}. The method succeeds at its stated objective: improvement is concentrated on the operation combinations that the weakness score in Eq.~\eqref{eq:weakness_score} up-weighted, the direction of the effect is stable across two evaluation protocols that differ substantially in what they measure, and the size of the effect tracks how directly each protocol can detect a broadened-but-not-yet-modal output distribution rather than any property of the sampling procedure itself. This is the pattern a targeted intervention should produce if the underlying diagnosis is correct, and it is difficult to reconcile with the alternative hypothesis that the gains in Tables~\ref{tab:solving_operation_after_benchmark} and~\ref{tab:solving_operation_after_pass_at_k} are a generic continued-training effect unrelated to the diagnosed weak slices.

At the same time, the constraint-tracking case is an instructive limit on this success. Under the standard suite, the constraint-tracking-bearing pairs and triples are exactly the slices that fail to improve, which, read on its own, would look like a failure of the intervention on precisely the operations it was meant to address. Only the Pass@128 re-evaluation reveals that the underlying capability gap on these slices has in fact narrowed substantially. This suggests that the diagnosis-aligned weighting succeeds at increasing the representation of the targeted operation combinations in the training signal, but that a single-shot, modal-output evaluation can still under-detect this kind of improvement when it manifests as broadened trajectory coverage rather than as a sharpened top completion. Consistent with the caution already raised in Section~\ref{subsubsec:mapping_rules_motivation} and Section~\ref{subsubsec:sampling_from_synthetic_instruction_data}, the mapping rules and sampling procedure are best read as heuristics for generating and prioritizing data-intervention hypotheses rather than as guarantees of improvement under any single evaluation protocol; this case study indicates that the evaluation side of the loop needs to be correspondingly multi-protocol, since a heuristic intervention can succeed at the level the mapping rules target without that success being visible under every downstream metric.

This case study is, in turn, an end-to-end demonstration of the data--evaluation closed loop introduced in Section~\ref{sec:analysis_toolkit}. A benchmark-level weakness on mathematical problem solving (Section~\ref{subsubsec:solving_operation_background}) was localized, via the evaluation sample taxonomy, into solving-operation slices, pairs, and triples whose raw error rates, pairwise effects, and non-additive composition effects pointed to compositional derivation rather than isolated arithmetic failure as the dominant failure mode (Section~\ref{subsubsec:solving_operation_diagnosis}). This diagnosis was translated into a concrete data action, a diagnosis-aligned weighted-sampling policy over synthetic instruction data (Section~\ref{subsubsec:sampling_from_synthetic_instruction_data}), and the resulting intervention was validated under a controlled, single-variable experimental design (Section~\ref{subsubsec:solving_operation_controlled_validation}). Finally, the outcome was mapped back onto the evaluation sample taxonomy at the operation level, under two protocols that probe different aspects of the model's output distribution, to determine whether the benchmark-level changes originated from the targeted slices rather than from an unrelated source. Each step left an inspectable, quantitative artifact, the weakness score, the retained feature pool, the per-slice sampling weights, and the per-slice accuracy changes under both protocols, so that the link between the original diagnosis and the final benchmark movement can be audited rather than asserted. The constraint-tracking case shows concretely why this auditability matters: it is the re-diagnosis step, and not the standard-suite score taken alone, that produces the correct attribution. In this sense, the closed loop accomplishes what Section~\ref{sec:analysis_toolkit} sets out for it, converting a noisy, benchmark-level observation into an actionable, auditable, and iterable procedure, while preserving throughout an explicit distinction between what the diagnosis and mapping rules suggest and what the controlled experimental evidence, examined at more than one level and under more than one protocol, actually shows.

\section{Conclusion}
\label{sec:conclusion}

\paragraph{Summary.}
This paper develops a closed-loop methodology for connecting evaluation failures in LLM pre-training to concrete training-data interventions. Its starting premise, developed in Section~\ref{sec:conceptual_framework}, is that a benchmark score is too coarse a unit, and an individual evaluation sample too noisy a unit, to drive data decisions; the appropriate intermediate unit is the \emph{capability slice}, a group of evaluation samples that share structured conditions along background condition, task type, solving operation, and output constraint (Sections~\ref{subsec:understanding_evaluation}--\ref{subsec:understanding_model_capability}). Training data is, in turn, treated as the controllable substrate from which capability slices are formed: we separate it into an instruction-data path that shares the same four-dimensional vocabulary as evaluation samples, and a non-instruction-data path characterized along content world, discourse structure, operation opportunity, and supervision density (Section~\ref{subsec:understanding_training_data}).

Section~\ref{sec:analysis_toolkit} operationalizes this framework into three components. An evaluation sample taxonomy, annotated over sixteen commonly used pre-training benchmarks through iterative human--LLM co-design, localizes weaknesses below the level of the benchmark name. A non-instruction data taxonomy characterizes corpus segments along the four data-side dimensions above. A set of evaluation-to-data mapping rules connects the two: taxonomy-aligned correspondences where the dimensions were designed jointly, and an LLM-assisted mechanism for slices that fall outside them. Together, these components instantiate the five-step workflow introduced in Section~\ref{subsec:evaluation_to_data_mapping_rules},
\[
\text{Weak Capability Slice} \rightarrow \text{Data Affordance Profile} \rightarrow \text{Data Action} \rightarrow \text{Experimental Validation} \rightarrow \text{Result Analysis},
\]
which turns a benchmark-level observation into an auditable, iterable data-intervention loop.

The two case studies in Section~\ref{sec:case_study} exercise this loop under qualitatively different regimes. In the first (Section~\ref{subsec:case_study_output_constraint}), an aggregate $-46.82\%$ relative drop on BBH after continued pre-training initially resembled a regression in multi-step reasoning, the capability BBH is designed to probe. Decomposing the regression along the output-constraint dimension instead traced it to a masked \(\texttt{\textless EOS\textgreater}\) loss in the continued pre-training pipeline: the model reached the correct closed-set answer but did not terminate generation afterward, so an otherwise correct response was rejected by the benchmark's exact-match scorer. Restoring loss supervision on the \(\texttt{\textless EOS\textgreater}\) token raised BBH from 25.14 to 66.44, above the warm-start checkpoint, while leaving the other fifteen benchmarks largely unaffected and, on several of them, modestly improving performance further. Here, the benchmark name actively misled the initial diagnosis: the failure resided in the construction of the training objective rather than in the data mixture or in the model's underlying capability.

The second case study (Section~\ref{subsec:case_study_solving_operation}) addresses a more characteristic pre-training weakness: near-zero Pass@128 on AIME, indicating that increasing the test-time sampling budget alone could not surface a valid reasoning trajectory. Decomposing this weakness by solving operation, together with a pairwise- and composition-effect analysis distinguishing additive from non-additive difficulty, localized the failure to specific operation combinations, among them constraint tracking, boundary-case reasoning, symbolic transformation, counting, and comparison, rather than to a uniform deficit in arithmetic computation or equation formulation. Translating this diagnosis into a diagnosis-aligned, weakness-targeted importance-sampling procedure over a pool of synthetic instruction data, with the training recipe and token budget held fixed, raised AIME2025 Pass@128 from 6.67 to 26.67 and AIME2026 Pass@128 from 0.00 to 26.67, while leaving general-domain performance close to unchanged. A subsequent re-evaluation at the level of the targeted operation combinations confirmed that this improvement was concentrated on the diagnosed slices, with the clearest and most consistent gains appearing under Pass@128, the protocol best suited to detecting reasoning paths that have become reachable but are not yet the model's modal completion.

Across both case studies, the diagnostic decomposition performed the work that an aggregate score could not: distinguishing whether an observed change in benchmark performance reflected the capability the benchmark was designed to measure, or a more specific and more directly addressable property of the training objective, the data mixture, or the evaluation protocol itself.

\paragraph{Limitations.}
Several aspects of this work are best understood as an initial methodology rather than as established conclusions. The evaluation-to-data mapping rules (Section~\ref{subsec:evaluation_to_data_mapping_rules}) are structured heuristics for generating data-intervention hypotheses, not causal claims about the relationship between a corpus segment and a benchmark score. On the non-instruction-data side, they are further restricted to a small set of dimension-level correspondences that were aligned by design during taxonomy construction (domain, discourse form, context scope, and solving operation), supplemented by an LLM-assisted mechanism for slices that fall outside this correspondence. Any data intervention these rules suggest, whether taxonomy-aligned or LLM-assisted, remains a hypothesis until subjected to the kind of controlled experimental validation demonstrated in the two case studies, rather than being accepted at face value.

The empirical scope of this paper is also limited in ways that bear on how broadly its findings should be read. Both taxonomies, and the two case studies that exercise them, were instantiated on a fixed set of sixteen commonly used pre-training benchmarks and on a single continued pre-training pipeline; their behavior on other other training stages, or evaluation suites oriented toward different capabilities remains to be tested.

A further limitation concerns the asymmetry between the two sides of the mapping. The controlled comparison in Section~\ref{subsec:case_study_solving_operation} exercises only the instruction-data branch, constructing a targeted instruction subset from a synthetic pool; in both case studies, the non-instruction data taxonomy is used mainly as a descriptive vocabulary for stating data-side targets rather than as the basis for an executed intervention. A symmetric test, retrieving, reweighting, or filtering pre-training corpus segments by the same data affordance profile and validating the result under a comparably controlled design, has not been performed, and it remains an open question whether the loop is equally effective on this more indirect side of the mapping.

\bibliographystyle{unsrtnat}
\bibliography{references}

\appendix
\section{Additional Figures for Operation-level Diagnosis}
\label{app:operation_level_diagnosis_figures}

\subsection{Figures of Pairwise Composition Effects}
\label{app:pairwise_figures}

\begin{figure}[t]
    \centering
    \includegraphics[width=0.95\linewidth]{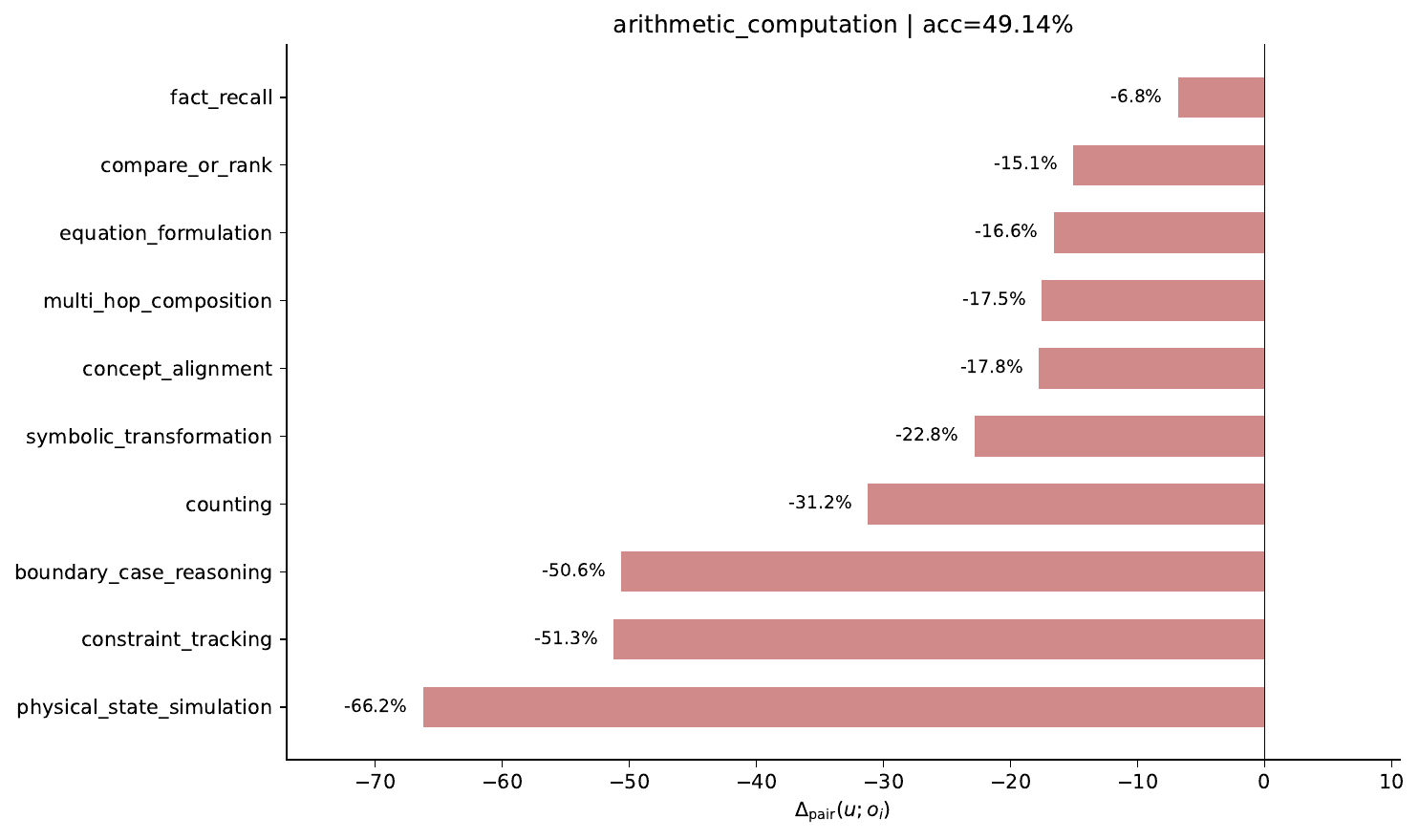}
    \caption{Pairwise Composition Effect of \texttt{arithmetic\_computation}}
    \label{fig:pairwise_arithmetic_computation}
\end{figure}

\begin{figure}[t]
    \centering
    \includegraphics[width=0.95\linewidth]{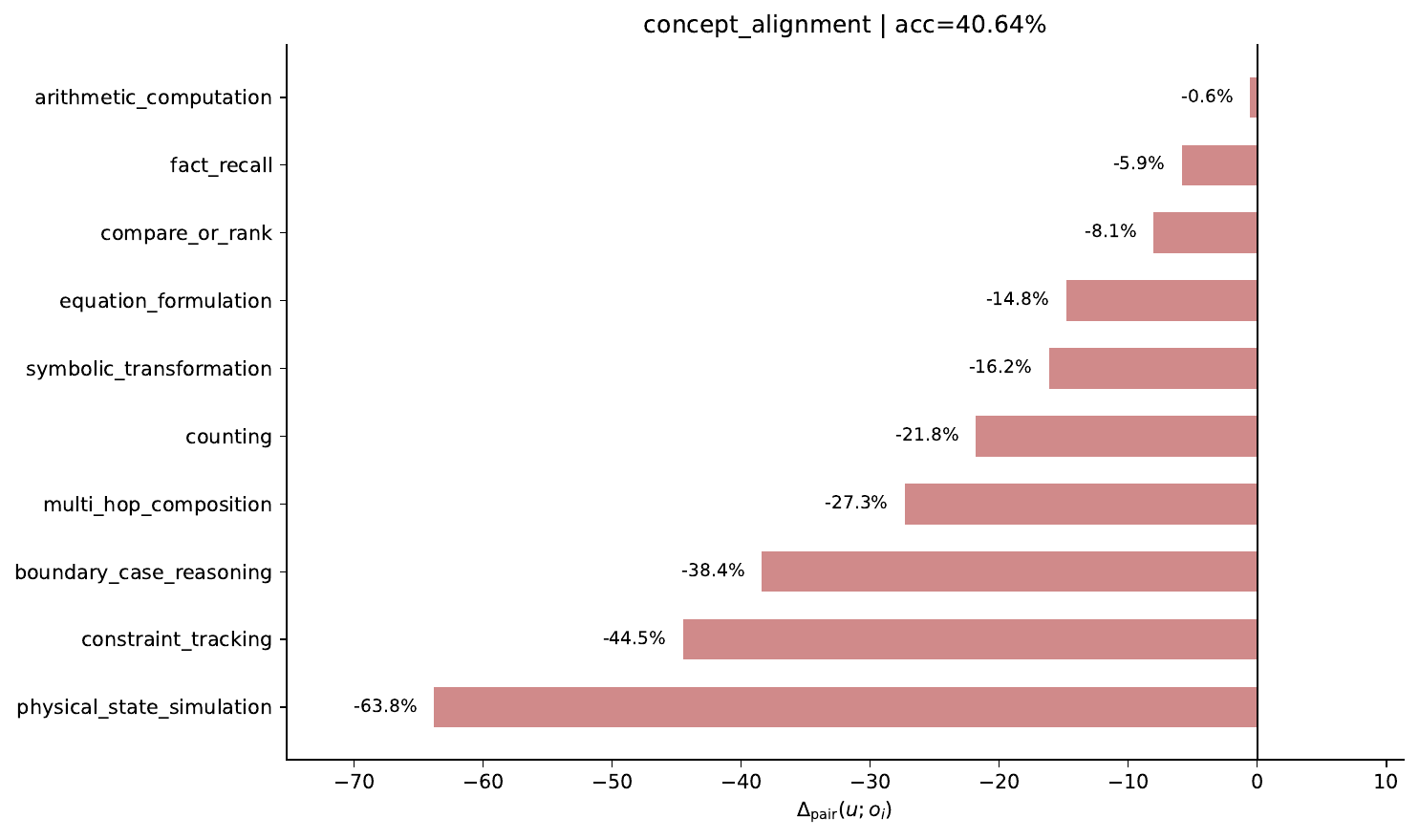}
    \caption{Pairwise Composition Effect of \texttt{concept\_alignment}}
    \label{fig:pairwise_concept_alignment}
\end{figure}

\begin{figure}[t]
    \centering
    \includegraphics[width=0.95\linewidth]{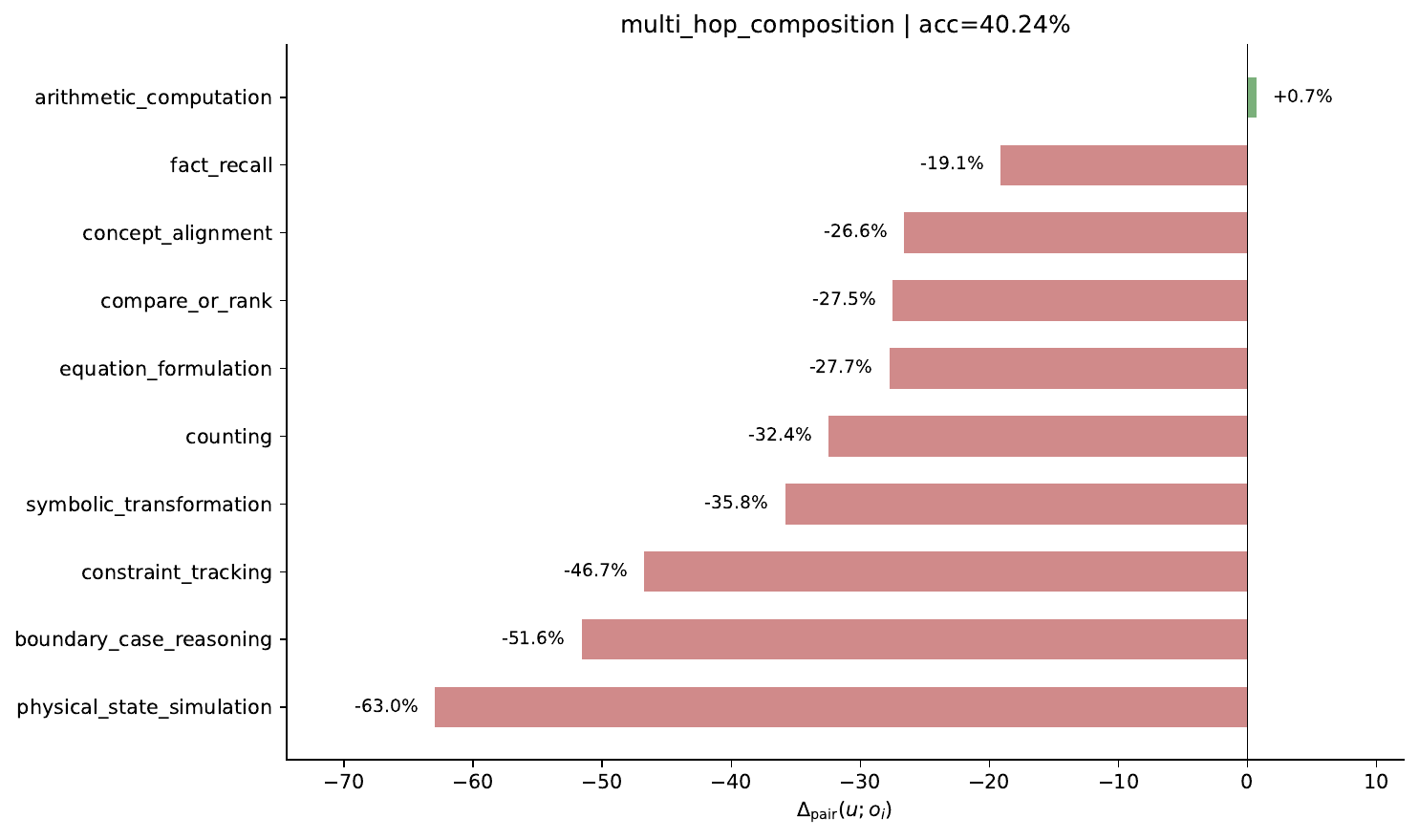}
    \caption{Pairwise Composition Effect of \texttt{multi\_hop\_composition}}
    \label{fig:pairwise_multi_hop_composition}
\end{figure}

\begin{figure}[t]
    \centering
    \includegraphics[width=0.95\linewidth]{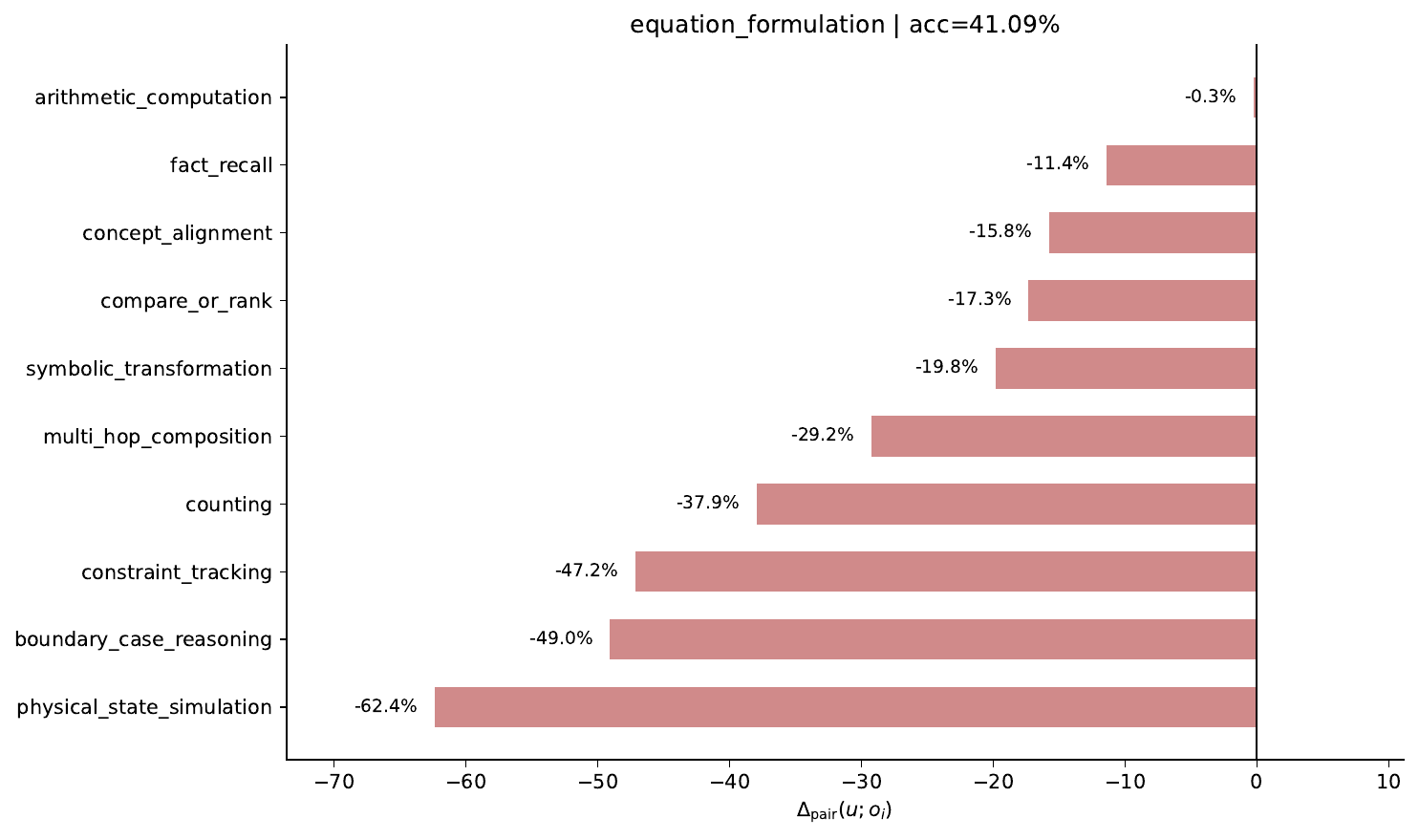}
    \caption{Pairwise Composition Effect of \texttt{equation\_formulation}}
    \label{fig:pairwise_equation_formulation}
\end{figure}

\begin{figure}[t]
    \centering
    \includegraphics[width=0.95\linewidth]{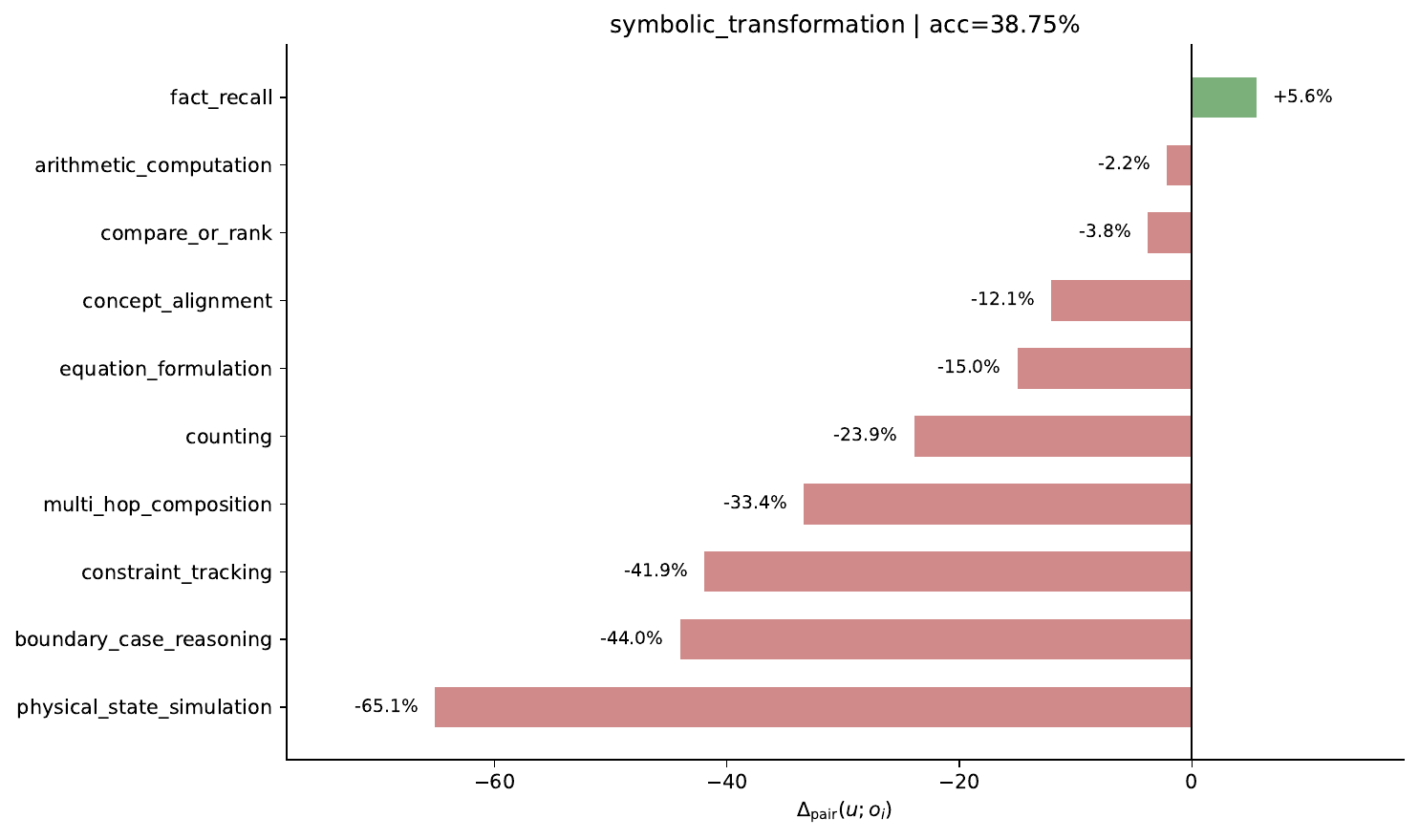}
    \caption{Pairwise Composition Effect of \texttt{symbolic\_transformation}}
    \label{fig:pairwise_symbolic_transformation}
\end{figure}

\begin{figure}[t]
    \centering
    \includegraphics[width=0.95\linewidth]{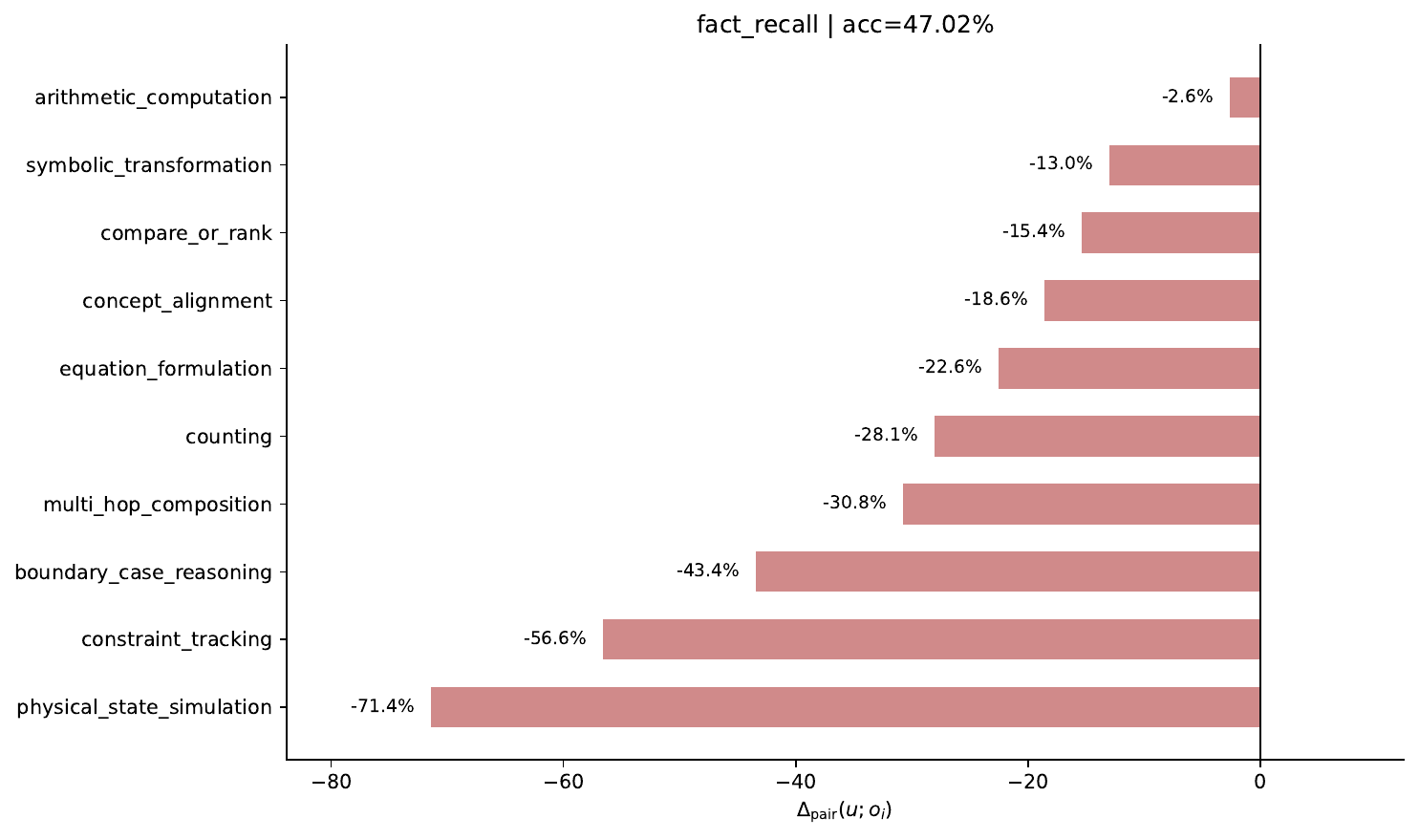}
    \caption{Pairwise Composition Effect of \texttt{fact\_recall}}
    \label{fig:pairwise_fact_recall}
\end{figure}

\begin{figure}[t]
    \centering
    \includegraphics[width=0.95\linewidth]{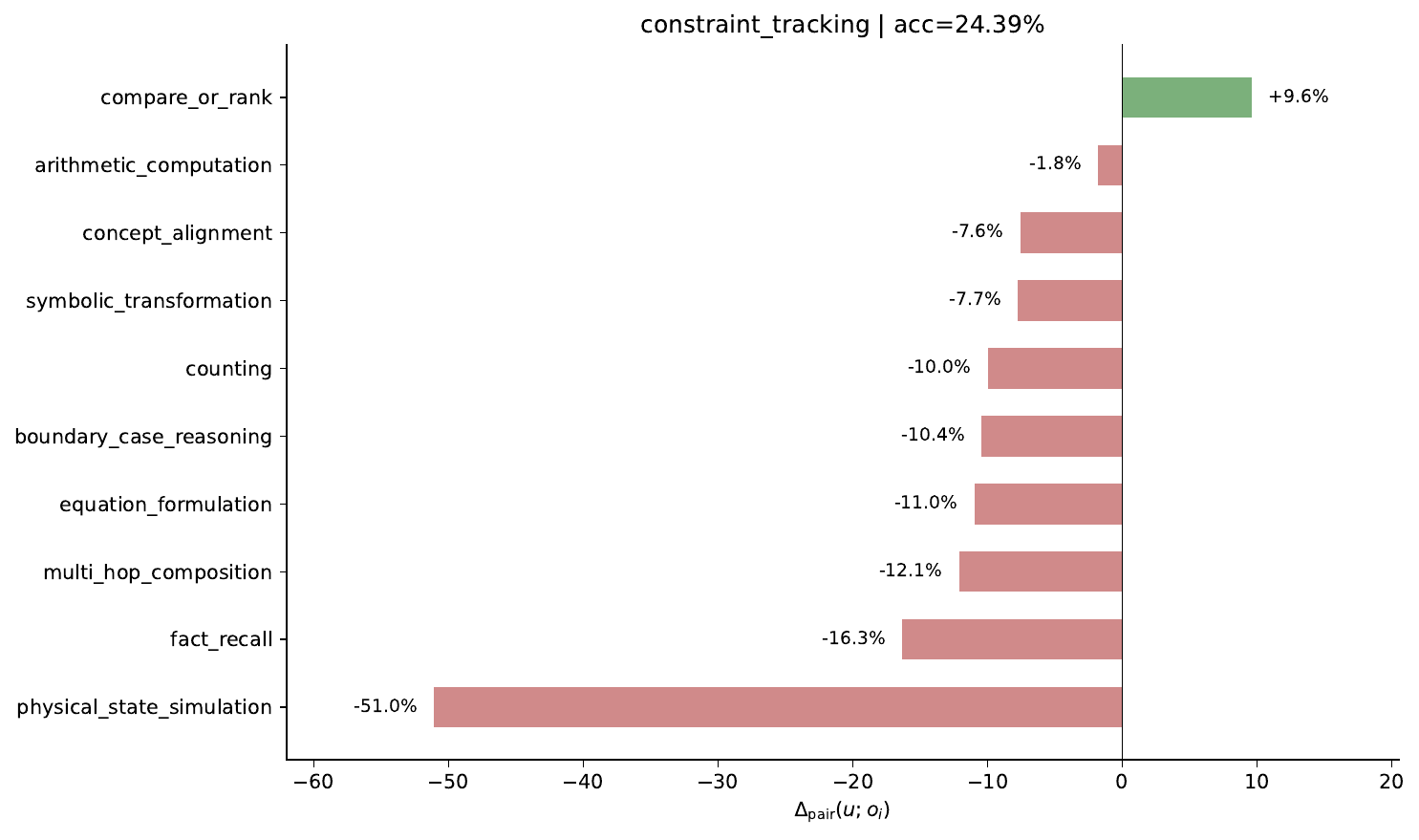}
    \caption{Pairwise Composition Effect of \texttt{constraint\_tracking}}
    \label{fig:pairwise_constraint_tracking}
\end{figure}

\begin{figure}[t]
    \centering
    \includegraphics[width=0.95\linewidth]{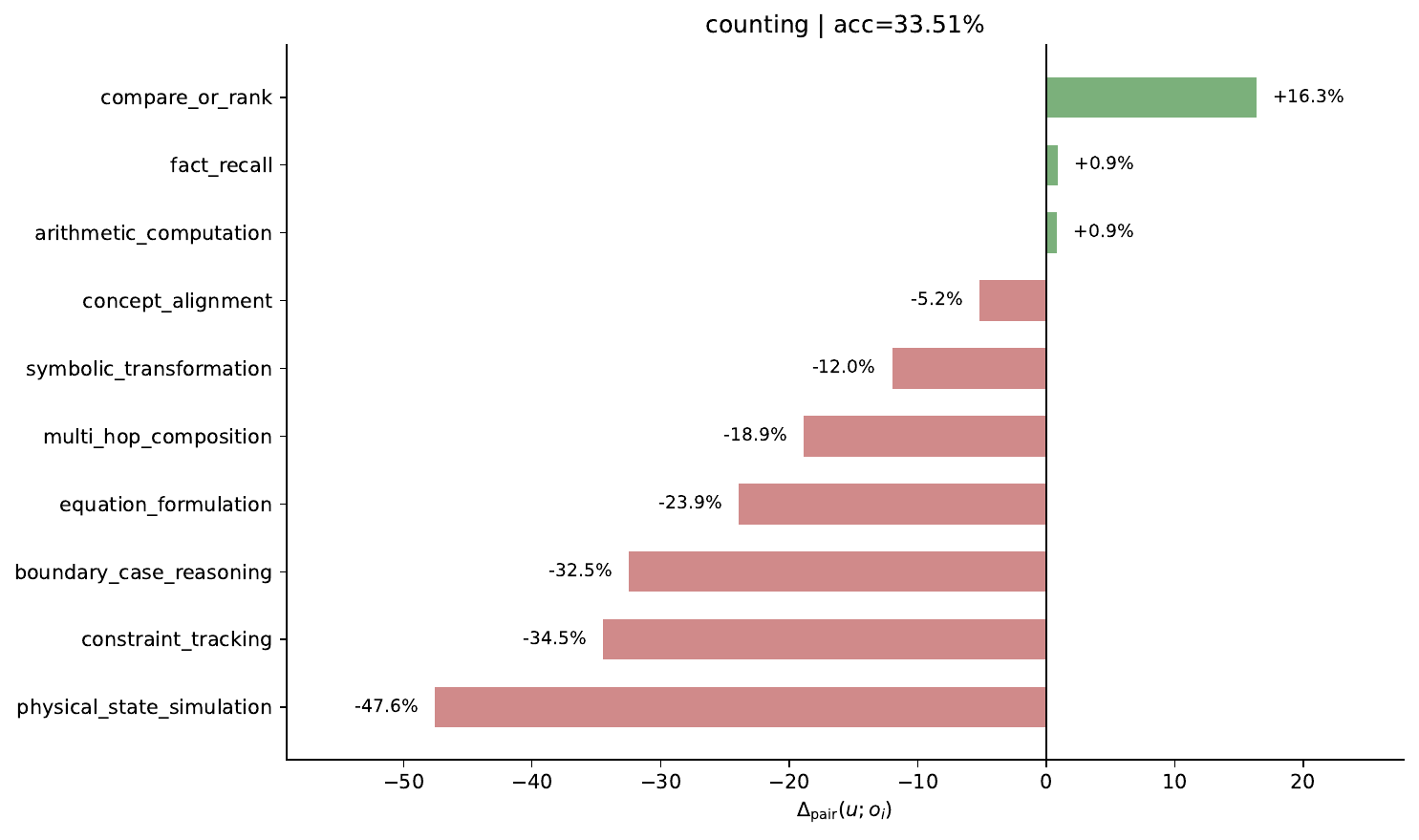}
    \caption{Pairwise Composition Effect of \texttt{counting}}
    \label{fig:pairwise_counting}
\end{figure}

\begin{figure}[t]
    \centering
    \includegraphics[width=0.95\linewidth]{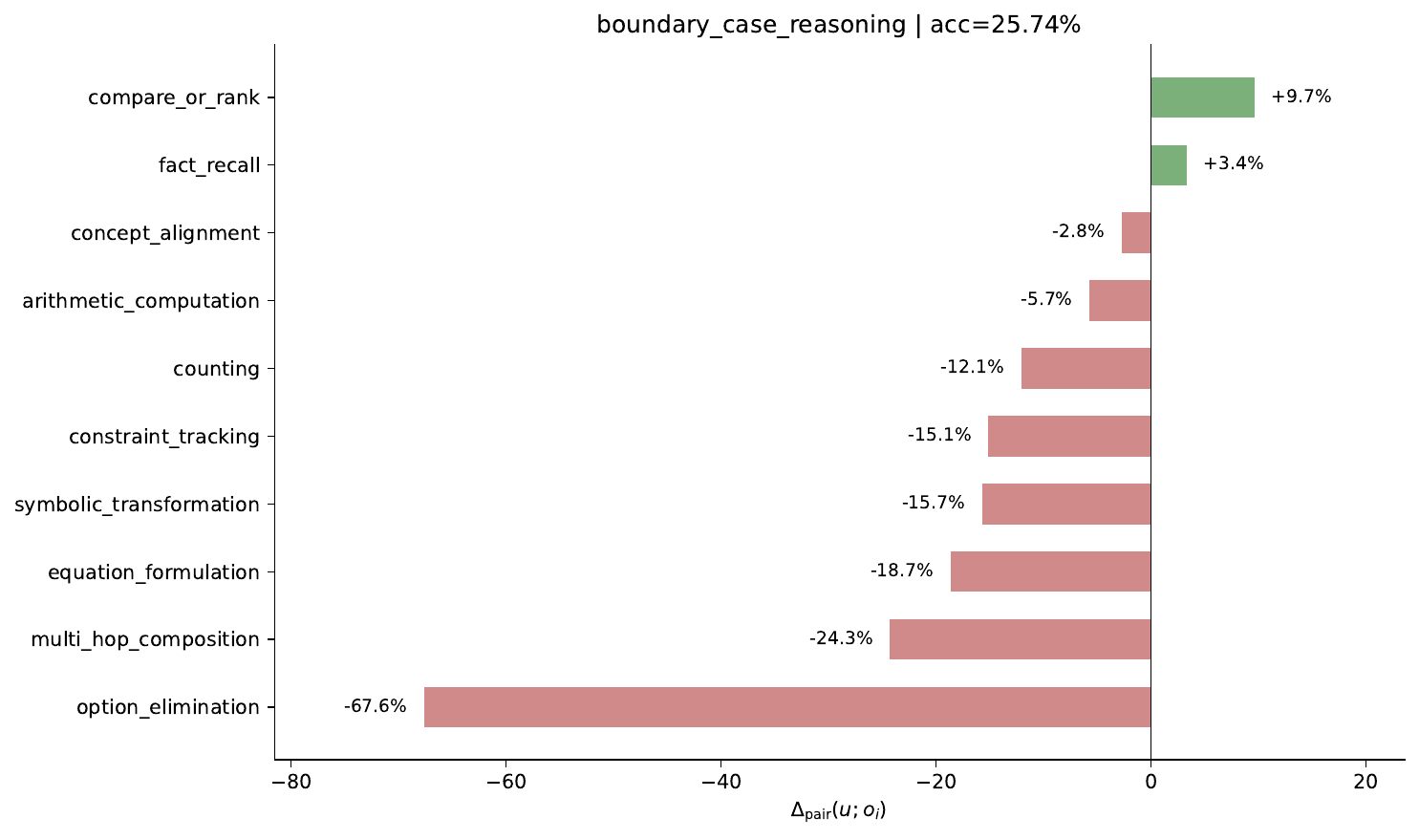}
    \caption{Pairwise Composition Effect of \texttt{boundary\_case\_reasoning}}
    \label{fig:pairwise_boundary_case_reasoning}
\end{figure}

\begin{figure}[t]
    \centering
    \includegraphics[width=0.95\linewidth]{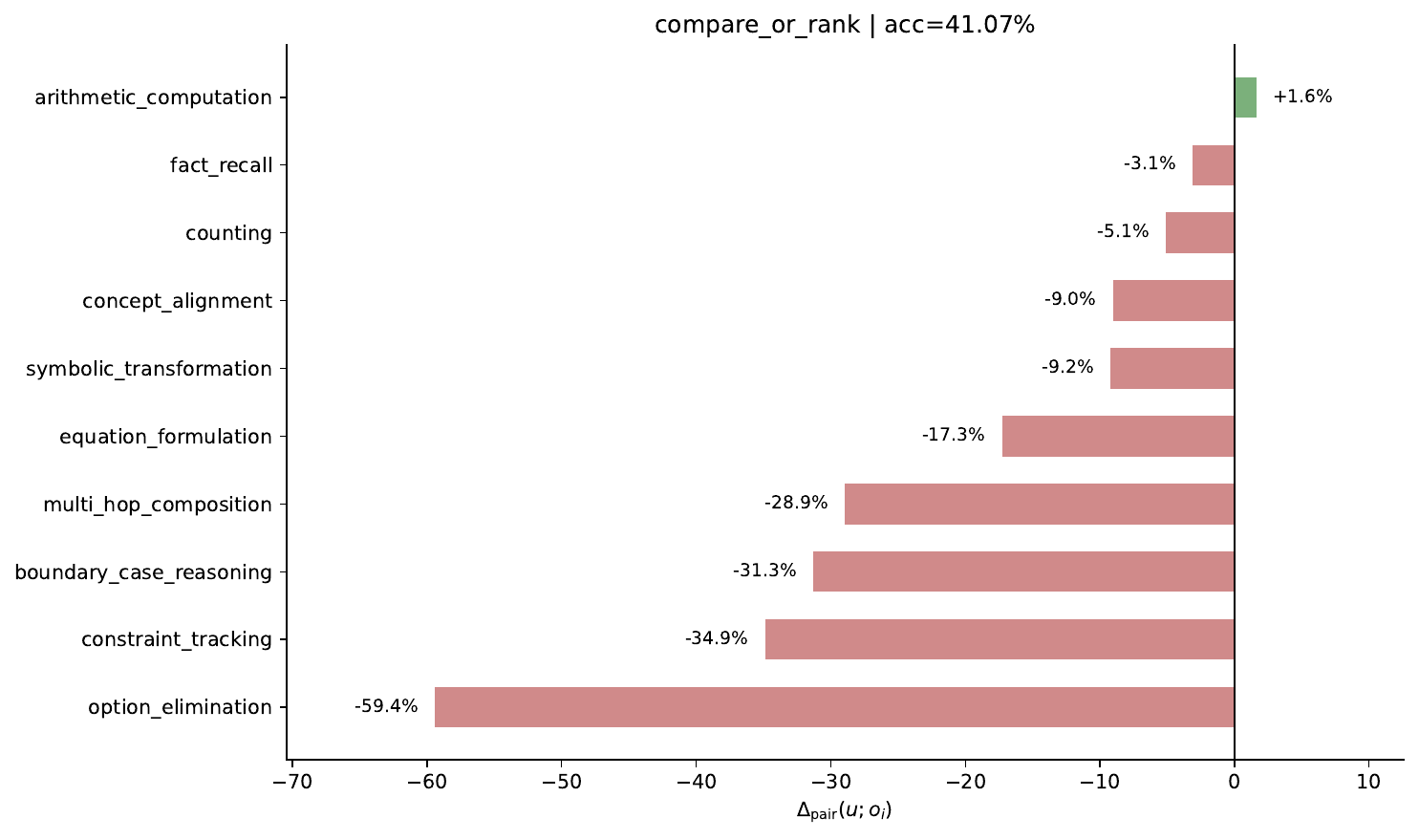}
    \caption{Pairwise Composition Effect of \texttt{compare\_or\_rank}}
    \label{fig:pairwise_compare_or_rank}
\end{figure}

\subsection{Figures of Non-additive Composition Effects}
\label{app:non_additive_figures}

\begin{figure}[t]
    \centering
    \includegraphics[width=0.95\linewidth]{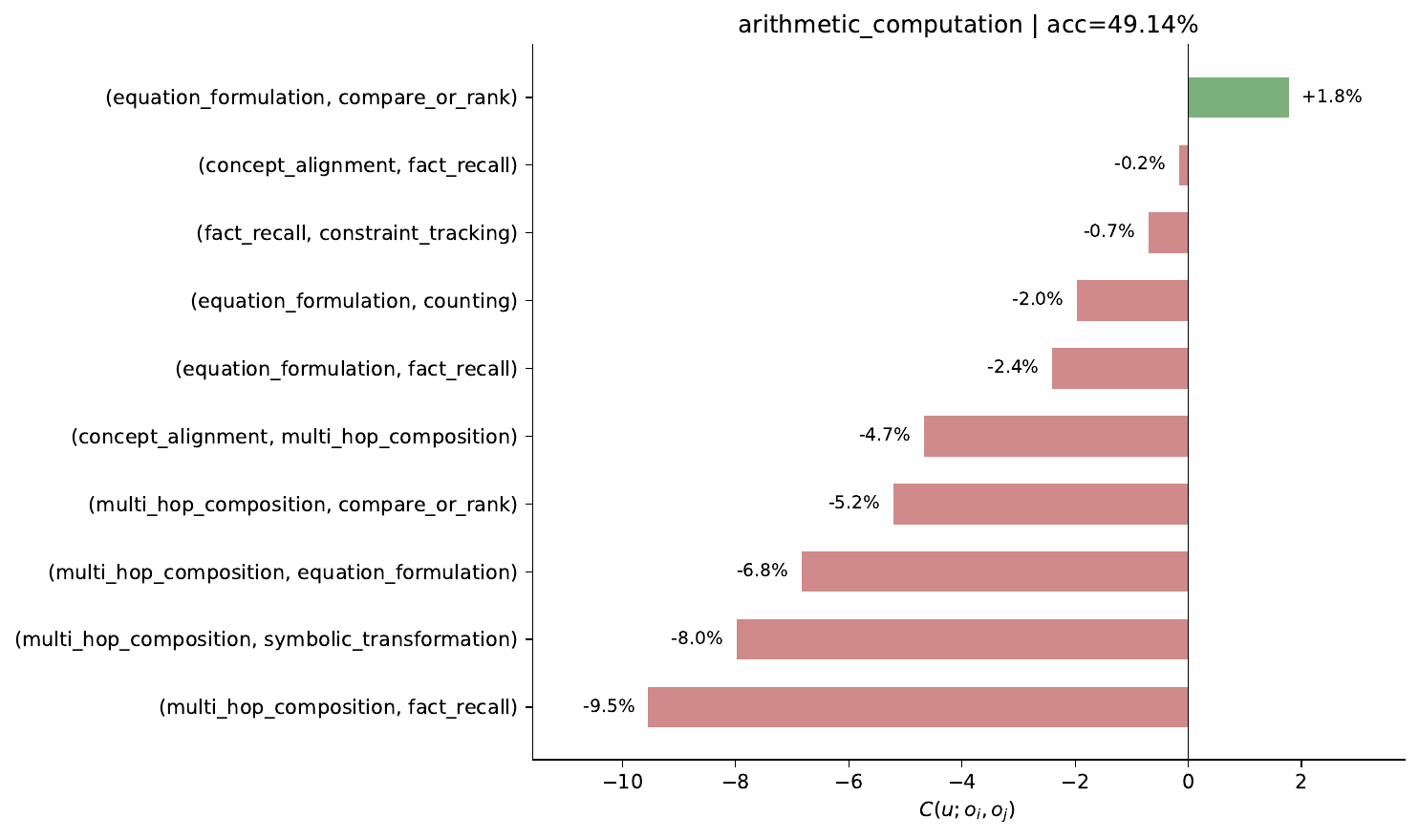}
    \caption{Non-additive Composition Effect of \texttt{arithmetic\_computation}}
    \label{fig:non_additive_arithmetic_computation}
\end{figure}

\begin{figure}[t]
    \centering
    \includegraphics[width=0.95\linewidth]{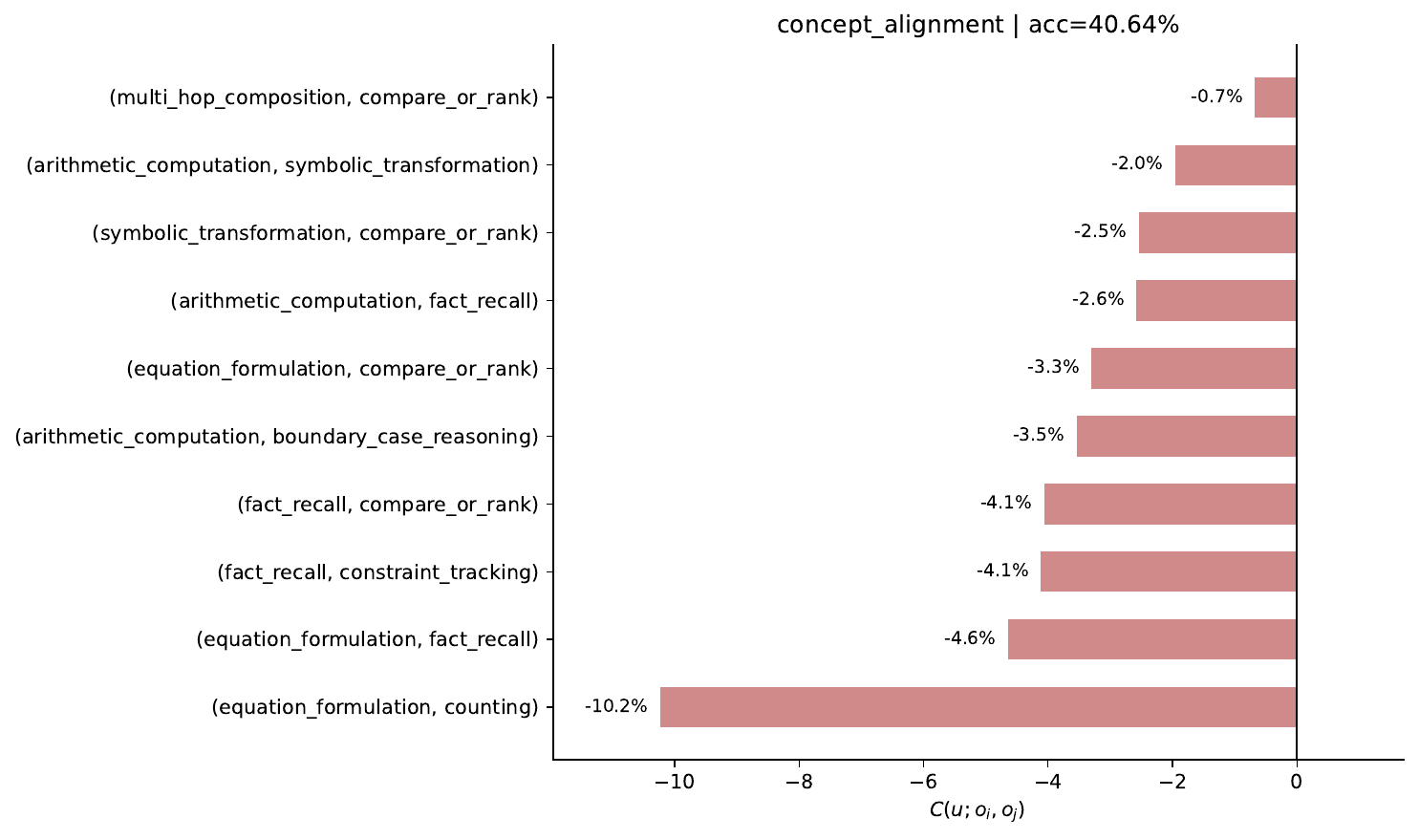}
    \caption{Non-additive Composition Effect of \texttt{concept\_alignment}}
    \label{fig:non_additive_concept_alignment}
\end{figure}

\begin{figure}[t]
    \centering
    \includegraphics[width=0.95\linewidth]{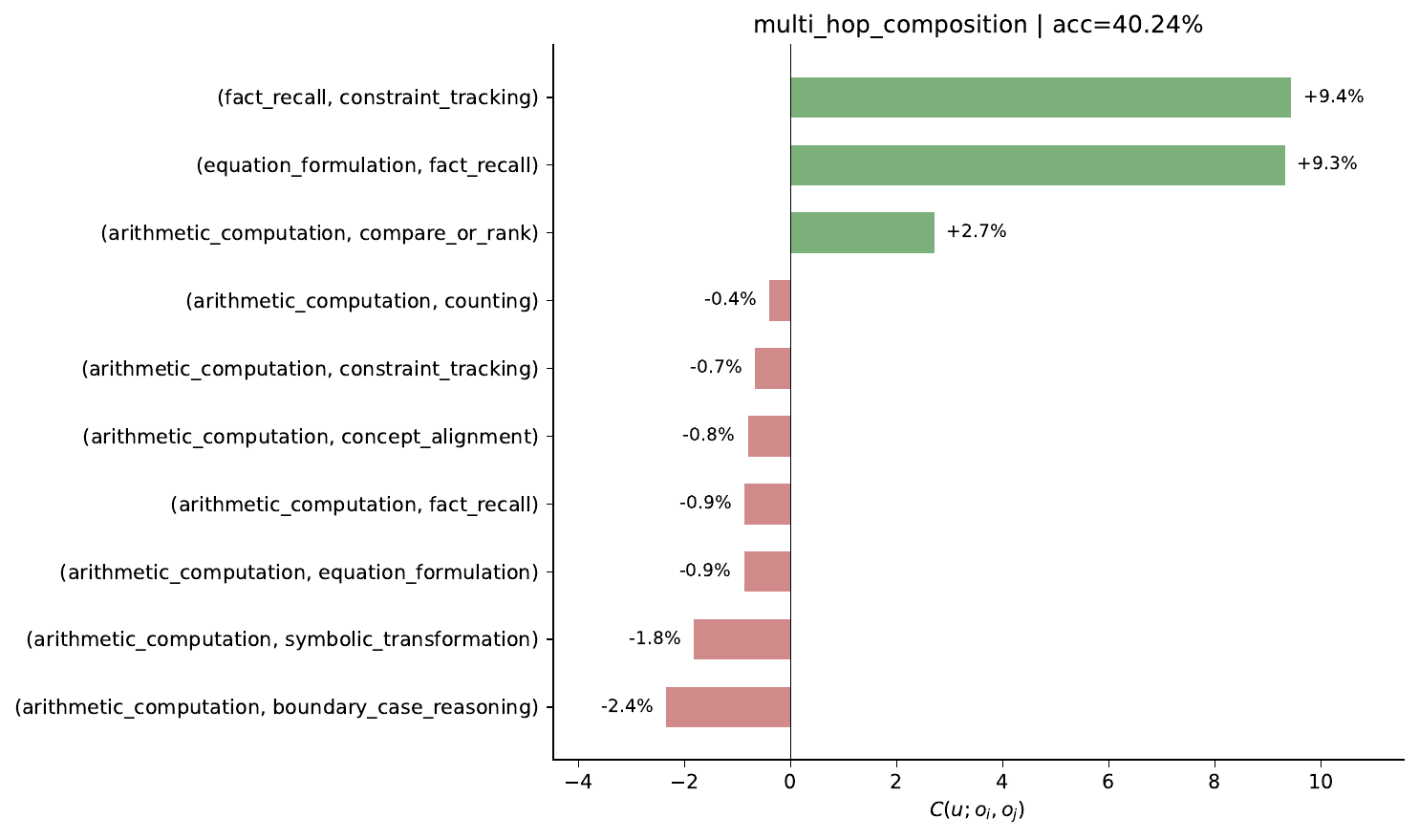}
    \caption{Non-additive Composition Effect of \texttt{multi\_hop\_composition}}
    \label{fig:non_additive_multi_hop_composition}
\end{figure}

\begin{figure}[t]
    \centering
    \includegraphics[width=0.95\linewidth]{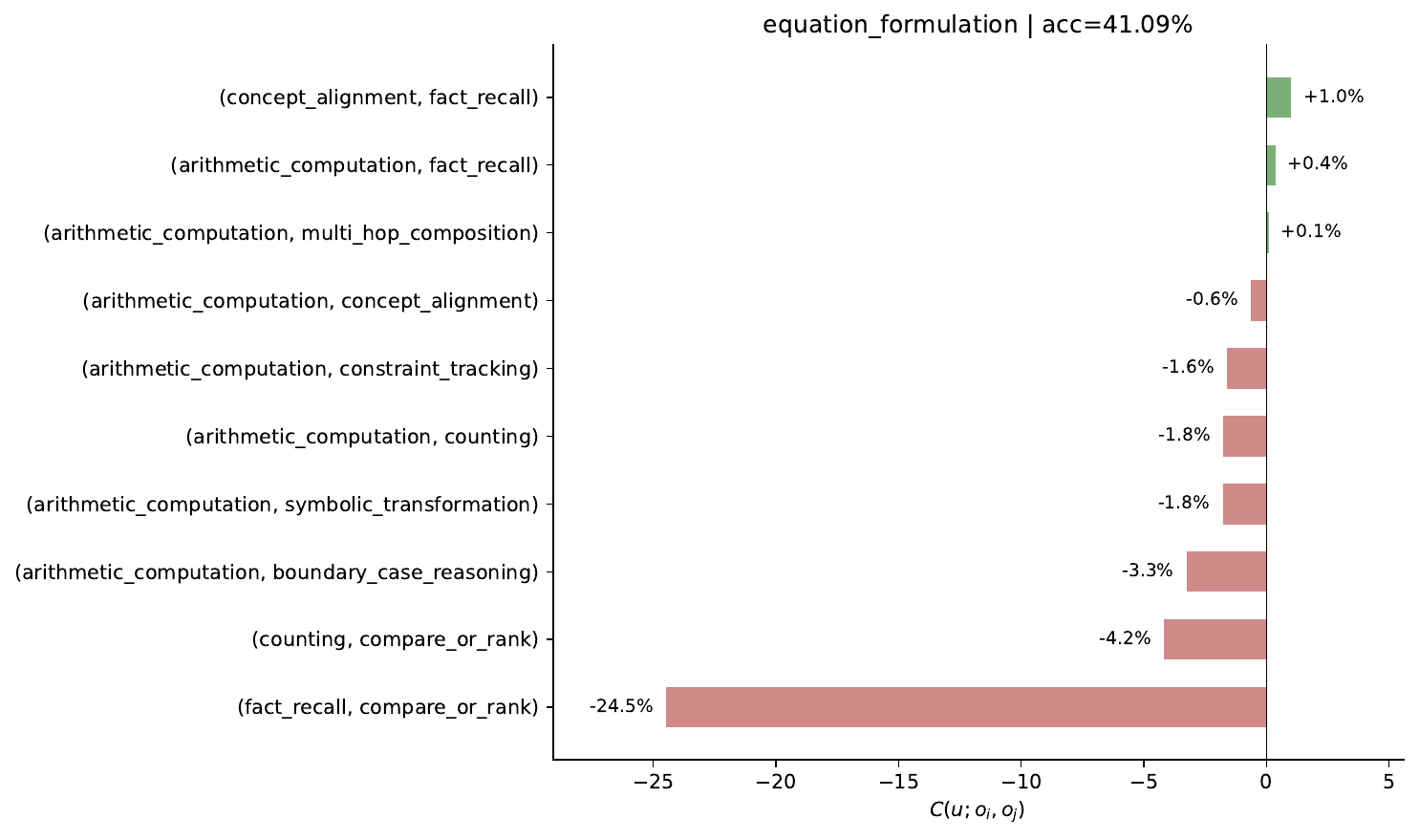}
    \caption{Non-additive Composition Effect of \texttt{equation\_formulation}}
    \label{fig:non_additive_equation_formulation}
\end{figure}

\begin{figure}[t]
    \centering
    \includegraphics[width=0.95\linewidth]{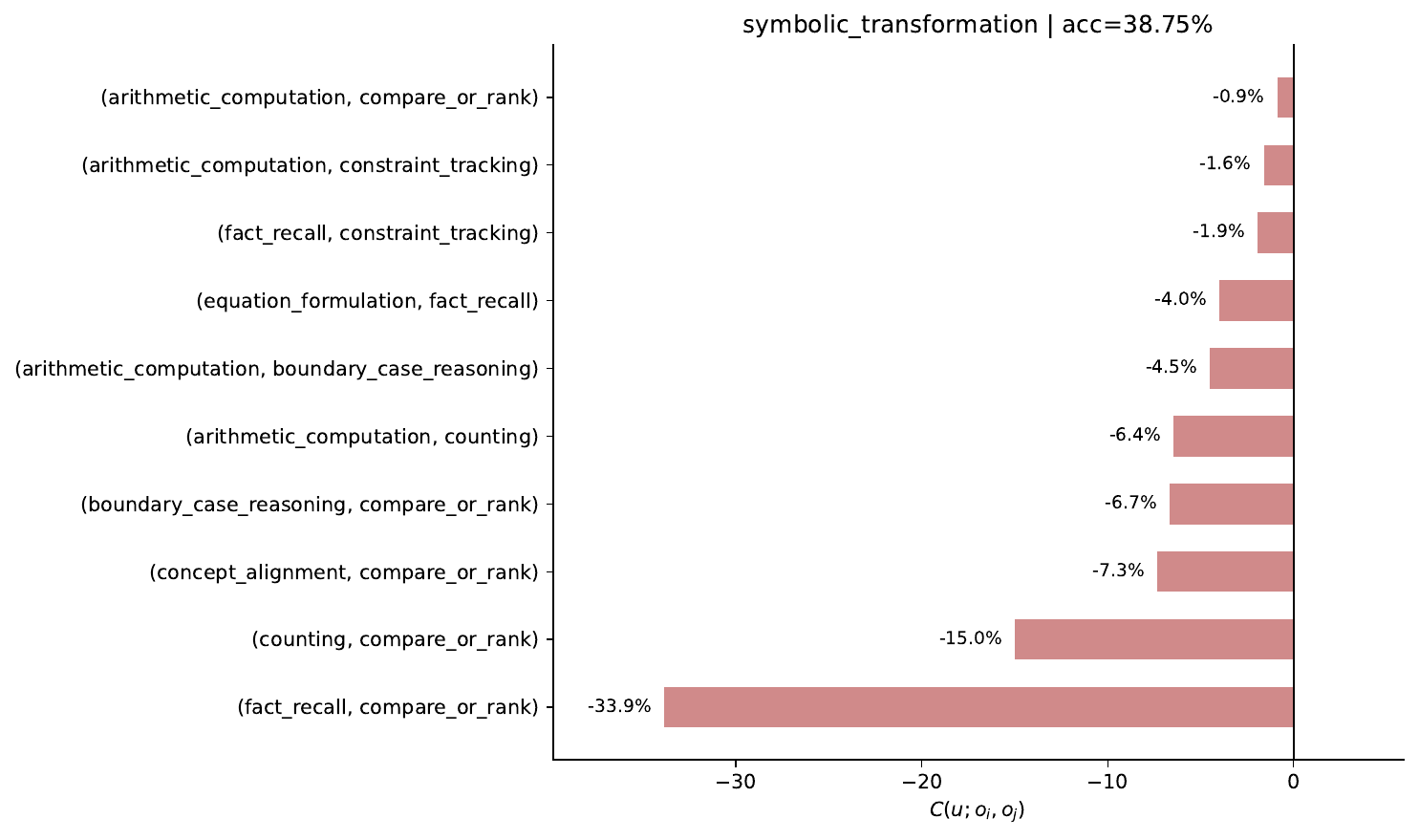}
    \caption{Non-additive Composition Effect of \texttt{symbolic\_transformation}}
    \label{fig:non_additive_symbolic_transformation}
\end{figure}

\begin{figure}[t]
    \centering
    \includegraphics[width=0.95\linewidth]{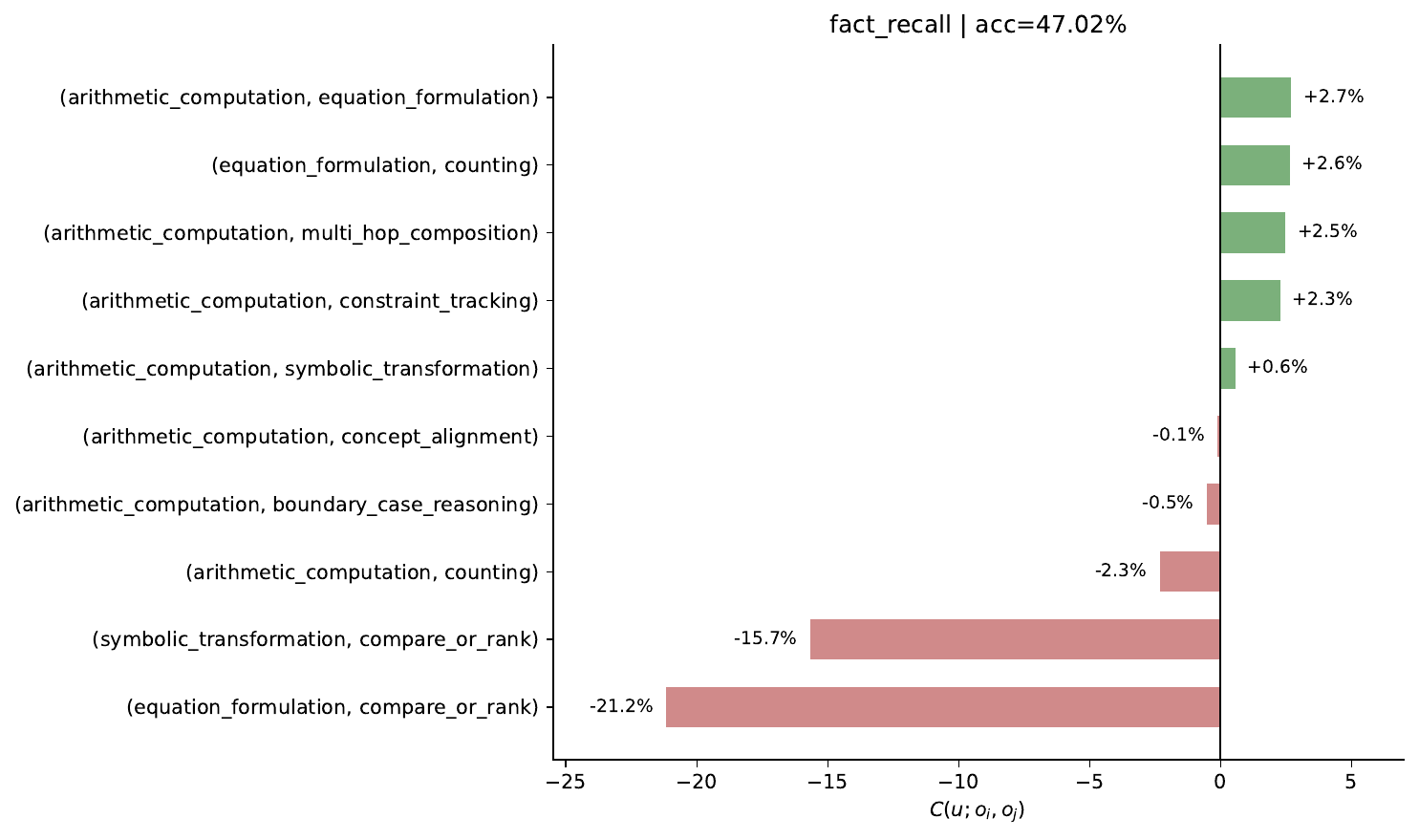}
    \caption{Non-additive Composition Effect of \texttt{fact\_recall}}
    \label{fig:non_additive_fact_recall}
\end{figure}

\begin{figure}[t]
    \centering
    \includegraphics[width=0.95\linewidth]{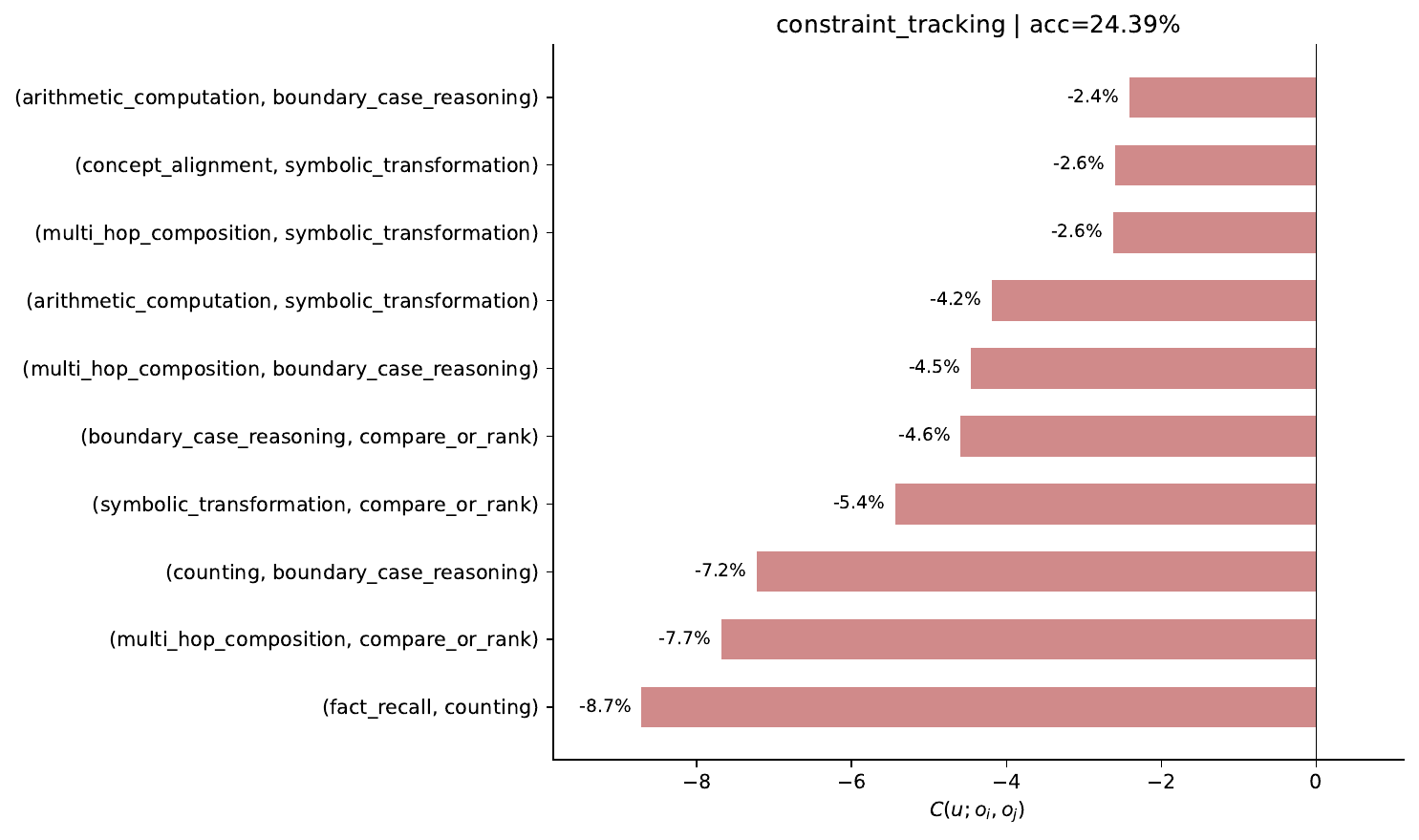}
    \caption{Non-additive Composition Effect of \texttt{constraint\_tracking}}
    \label{fig:non_additive_constraint_tracking}
\end{figure}

\begin{figure}[t]
    \centering
    \includegraphics[width=0.95\linewidth]{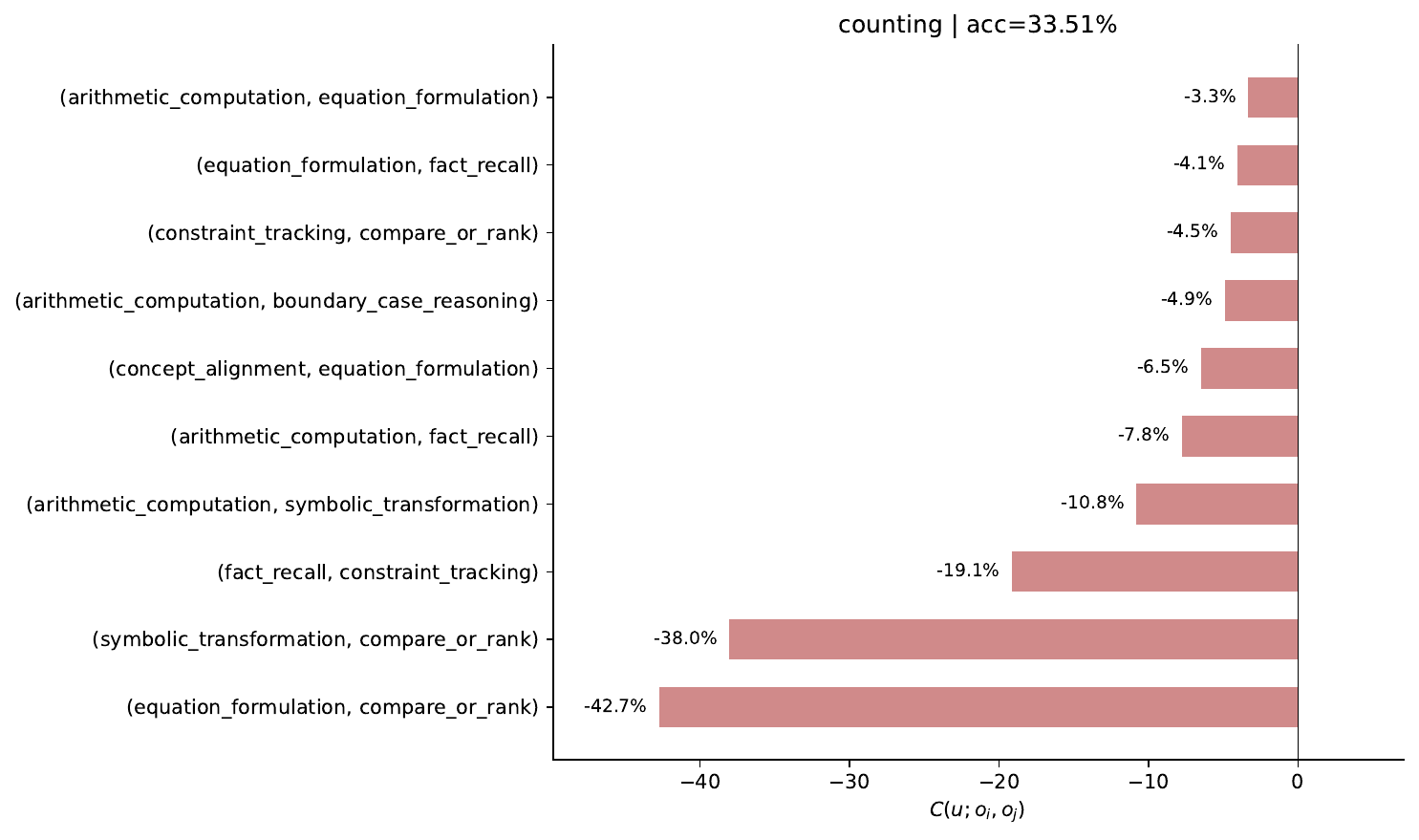}
    \caption{Non-additive Composition Effect of \texttt{counting}}
    \label{fig:non_additive_counting}
\end{figure}

\begin{figure}[t]
    \centering
    \includegraphics[width=0.95\linewidth]{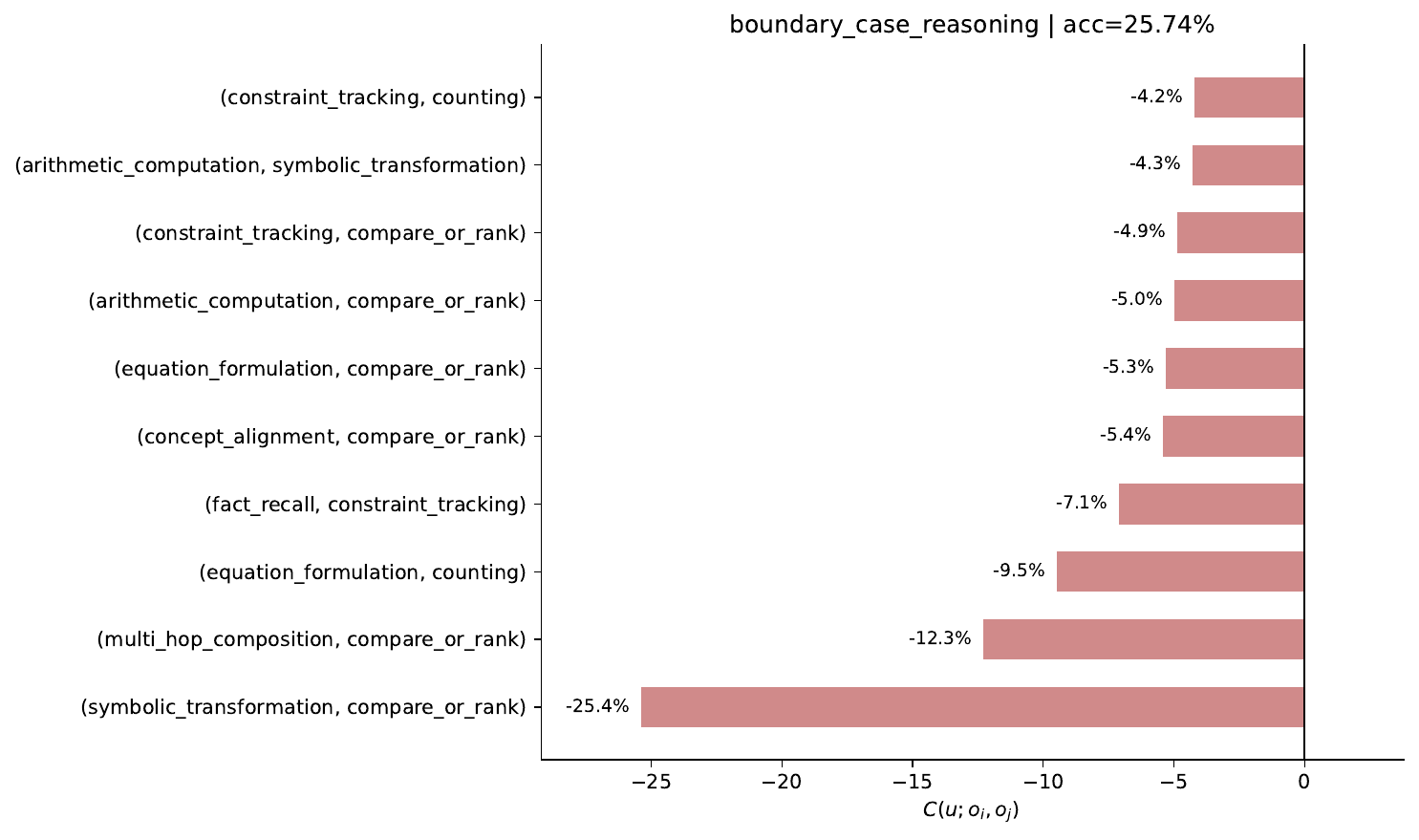}
    \caption{Non-additive Composition Effect of \texttt{boundary\_case\_reasoning}}
    \label{fig:non_additive_boundary_case_reasoning}
\end{figure}

\begin{figure}[t]
    \centering
    \includegraphics[width=0.95\linewidth]{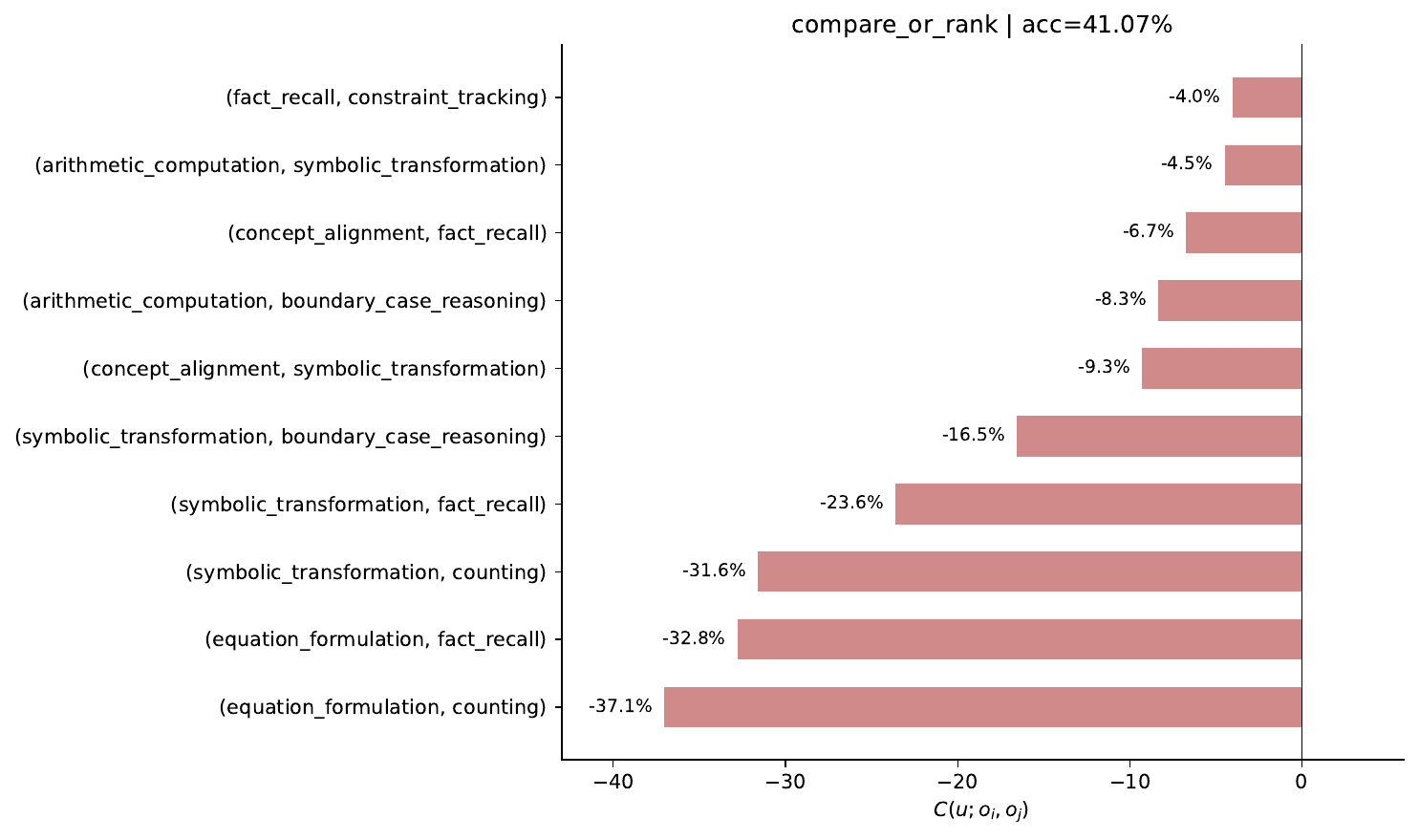}
    \caption{Non-additive Composition Effect of \texttt{compare\_or\_rank}}
    \label{fig:non_additive_compare_or_rank}
\end{figure}

\end{document}